\documentclass{article}

\usepackage{microtype}
\usepackage{graphicx}
\usepackage{subfigure}
\usepackage{booktabs} % for professional tables

\usepackage[accsupp]{axessibility} 
\usepackage{microtype}
\usepackage{graphicx}
\usepackage{booktabs} 
\usepackage{mathtools}
\usepackage{amsmath}
\usepackage{amssymb}
\usepackage{bbm}
\usepackage[utf8]{inputenc} 
\usepackage[T1]{fontenc}    
\usepackage{url}            
\usepackage{amsfonts}       
\usepackage{nicefrac}       
\usepackage[dvipsnames]{xcolor}         
\usepackage{colortbl}
\usepackage{subcaption}
\usepackage{enumitem}
\usepackage{multirow}
\usepackage{capt-of}
\usepackage{wrapfig}
\usepackage{makecell}
\usepackage{bm}
\usepackage{arydshln}
\usepackage{placeins}
\definecolor{Blue9}{rgb}{0.1,0.3,0.95}
\usepackage{comment}
\usepackage{algorithm}
\usepackage{algorithmic}
\usepackage{eqparbox}
\usepackage{amsmath, amsfonts, amsmath, amssymb, amsthm, bm}
\usepackage{pifont} % for checkmark

\theoremstyle{plain}

\theoremstyle{definition}

\theoremstyle{remark}

\renewcommand{\algorithmiccomment}[1]{\hfill{ $\rhd$ #1}}

\usepackage{xspace}
\usepackage{float}
\usepackage{color}
\usepackage{cuted}
\usepackage{etoolbox}
\AfterEndEnvironment{strip}{\leavevmode}
\usepackage[most]{tcolorbox}

\definecolor{cvprblue}{rgb}{0.21,0.49,0.74}
\definecolor{myred}{rgb}{0.9,0.0,0.0}

\usepackage[pagebackref,breaklinks,colorlinks,citecolor=cvprblue, linkcolor=myred]{hyperref}
\usepackage{xurl}
\usepackage{adjustbox}
\usepackage{listings} % for python code
\usepackage{courier}
\definecolor{darkteal}{HTML}{005f73}
\definecolor{deeppurple}{HTML}{6a0dad}
\definecolor{functioncolor}{RGB}{234, 21, 0}

\definecolor{gg}{gray}{0.92}
\newcolumntype{a}{>{\columncolor{gg}}c}
\definecolor{figred}{RGB}{234, 21, 0}
\definecolor{figblue}{RGB}{48, 132, 194}
\definecolor{figgreen}{RGB}{99, 154, 63}
\usepackage{pifont}
\usepackage{tikz}

\newcommand{\fref}[1]{Fig.~\ref{#1}}

\definecolor{softblue}{HTML}{3291B6}
\usepackage{multicol}

\newcommand{\pspace}{\;\;}

\usepackage{lmodern}
\usepackage{blindtext}% for dummy text only
\usepackage{amssymb}
\definecolor{cvprblue}{rgb}{0.21,0.49,0.74}

\usepackage[accepted]{icml2026}

\usepackage{amsmath}
\usepackage{amssymb}
\usepackage{mathtools}
\usepackage{amsthm}

\usepackage{array}

\usepackage[capitalize,noabbrev]{cleveref}

\usepackage[textsize=tiny]{todonotes}

\icmltitlerunning{Self-Refining Video Sampling}

\begin{document}

\twocolumn[
% \icmltitle{Reliable Video Sampling through Inference-Time Self-Refinement}
\icmltitle{Self-Refining Video Sampling}



\icmlsetsymbol{equal}{*}
\icmlsetsymbol{equal_adv}{$\dagger$}

\begin{icmlauthorlist}
\icmlauthor{Sangwon Jang}{equal,yyy}
\icmlauthor{Taekyung Ki}{equal,yyy}
\icmlauthor{Jaehyeong Jo}{equal,yyy}
\icmlauthor{Saining Xie}{nyu}
\icmlauthor{Jaehong Yoon}{equal_adv,ntu}
% \icmlauthor{Sung Ju Hwang}{equal_adv,yyy,deep}
\icmlauthor{Sung Ju Hwang}{equal_adv,yyy,deep}
\end{icmlauthorlist}

\begin{center}
\small
{\textbf{Project Page}: 
\url{https://agwmon.github.io/self-refine-video/}}
\vspace{-0.15in}
\end{center}

\icmlaffiliation{yyy}{KAIST}
\icmlaffiliation{nyu}{NYU}
\icmlaffiliation{ntu}{NTU Singapore}
% \icmlaffiliation{yyy}{Korea Advanced Institute of Science and Technology (KAIST)}
% \icmlaffiliation{nyu}{New York University}
% \icmlaffiliation{ntu}{NTU Singapore}
\icmlaffiliation{deep}{DeepAuto.ai}

\icmlcorrespondingauthor{Sangwon Jang}{sangwon.jang@kaist.ac.kr}
% \icmlcorrespondingauthor{Taekyung Ki}{taekyung.ki@kaist.ac.kr}
% \icmlcorrespondingauthor{Jaehyeong Jo}{harryjo97@kaist.ac.kr}
\icmlcorrespondingauthor{Sung Ju Hwang}{sungju.hwang@kaist.ac.kr}

\icmlkeywords{Machine Learning, ICML}

\vskip 0.3in
]



\printAffiliationsAndNotice{\icmlEqualContribution  \icmlEqualAdvising} % otherwise use the standard 

\begin{abstract}
Modern video generators still struggle with complex physical dynamics, often falling short of physical realism. Existing approaches address this using external verifiers or additional training on augmented data, which is computationally expensive and still limited in capturing fine-grained motion. In this work, we present self-refining video sampling, a simple method that uses a pre-trained video generator trained on large-scale datasets as its own self-refiner. By interpreting the generator as a denoising autoencoder, we enable iterative inner-loop refinement at inference time without any external verifier or additional training. We further introduce an uncertainty-aware refinement strategy that selectively refines regions based on self-consistency, which prevents artifacts caused by over-refinement. Experiments on state-of-the-art video generators demonstrate significant improvements in motion coherence and physics alignment, achieving over 70\% human preference compared to the default sampler and guidance-based sampler.
% Our project page is at \url{https://agwmon.github.io/self-refine-video/}.
\end{abstract}
\vspace{-0.22in}
\section{Introduction}\label{sec:intro}
\vspace{-0.01in}
The rapid advancement of diffusion and flow matching models~\citep{song2020denoising, song2021scorebased, lipman2022flow} has led to powerful video generators, which are increasingly viewed as early-stage \emph{world models}~\citep{brooks2024sora,genie3,cosmos-predict2.5} that capture physical dynamics and causal structures of future states. Despite the impressive results, current video generators still struggle to model complex physical dynamics~\citep{kang2025how,li2025pisa}, and remain far from reliable physical simulators. The inconsistencies and implausible outputs undermine real-world applications, such as robot manipulation~\citep{qi2025strengthening, bharadhwaj2025genact,chen2025large}, where small visual errors, such as shape deformations of objects, can lead to incorrect actions.

\begin{figure}[t!]
    \centering
    \includegraphics[width=1\linewidth]{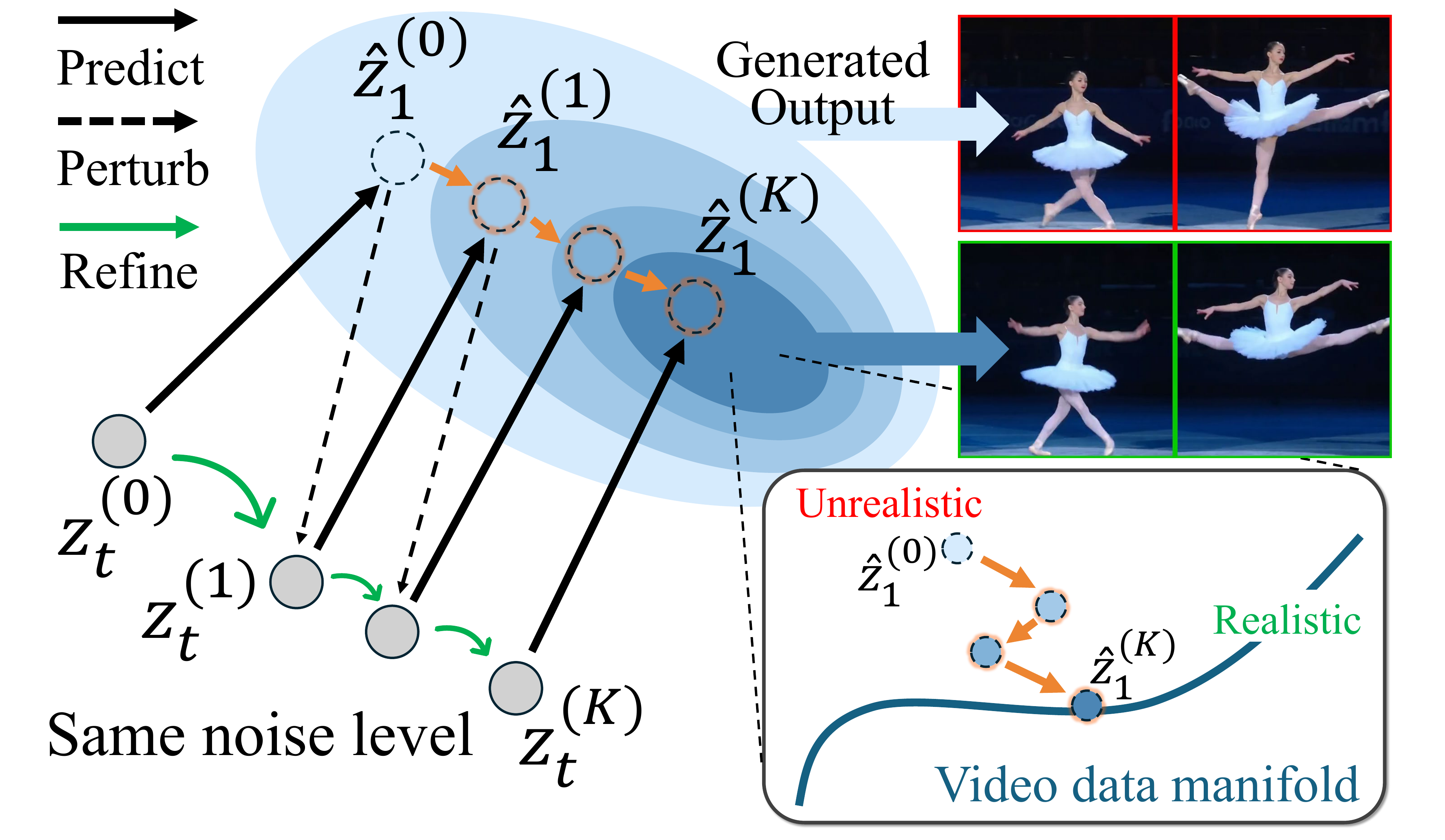}
    \caption{Concept of the \textbf{self-refining video sampling}. Within the same noise level, the video latent $z_t$ is refined as the predicted endpoint $\hat{z}_1$ is pulled toward the data manifold.
    \label{fig:concept}}
    \vspace{-6mm}
\end{figure}

Recent works attempt to address these limitations by either incorporating external models or additional training. One line of work employs external verifiers to improve physical plausibility via rejection sampling~\citep{azzolini2025cosmosreason1,liu2025videot1}, repeatedly generating new videos until success. Yet, low acceptance rates necessitate numerous proposals, making it highly inefficient.
Moreover, these verifiers are often domain-specific~\citep{chi2025empowering} and are ill-suited for evaluating temporal coherence and physical plausibility~\citep{bansal2025videophy2}. Another line of work adopts post-training strategies~\citep{liu2025videoalign,li2025pisa,cosmos-predict2.5}, for example, generating synthetic data and fine-tuning on the augmented dataset~\citep{cai2025phygdpo}. However, these methods typically require high-quality, domain-specific external data or substantial computation. Furthermore, accurately capturing fine-grained motion dynamics via reward models remains challenging, which in turn limits the applicability to real-world tasks.

% \sx{I think this is an important point you should highlight earlier -- i.e. you are not relying on an external model as verfier or refiner; another point that's worth mentioning is you do not generate synthetic data explicitly -- some self-refinement work requires you generate the data first and then finetune the model on the augmented dataset; you don't need any of that.} 
To overcome these limitations, we propose using the video generator as a \emph{self-refiner} at inference time, without external models or additional training. Modern video generators~\citep{yang2025cogvideox,wan2025} that are trained on large-scale datasets already encode rich priors over realistic motion and structure~\citep{yuan2025likephys,mi2025video}. 
We aim to leverage these learned priors by iteratively refining samples during inference. The key question is how to realize self-refinement for video generators. Unlike LLMs, which can directly re-ingest their output tokens and revise, video generators lack an explicit internal feedback signal for critique and correction, especially given the high-dimensionality and temporal coupling of videos.

To this end, we introduce \textbf{Predict-and-Perturb (P\&P)}, a training-free sampling method that uses a flow matching video generator as its own self-refiner. We reinterpret the flow matching objective as a time-conditioned denoising autoencoder (DAE)~\citep{pascal2008extractingdae,bengio2013generalizeddae} training, and reuse this property at inference time. Our main idea is to refine the video latents during sampling, by iteratively noising and denoising at a fixed noise level as illustrated in \fref{fig:concept}. Mirroring the corrupt–reconstruct structure of DAE, the model first \emph{predicts} a clean endpoint video latent and then \emph{perturbs} it back to the same noise level. This simple inner-loop refinement pulls the latent toward higher-density regions of learned video distribution, corresponding to temporally coherent and physically plausible videos.

We further propose \textbf{Uncertainty-aware P\&P}, an extension that retains the benefit of refined sampling while mitigating artifacts caused by over-refinement. While repeated P\&P iterations progressively improve video quality, naively applying them may lead to over-saturation from repeated classifier-free guidance (CFG)~\citep{ho2022classifier,sadat2024eliminating}, particularly in static regions. We extend P\&P by selectively refining only spatio-temporal regions where the model exhibits uncertainty, while leaving stable regions largely unchanged. We leverage a self-consistency measure from the model predictions within the P\&P process and use it to gate refinement at no extra computation cost. As a result, it retains the benefits of P\&P while mitigating over-refinement artifacts and preserving visual quality.

We validate our approach with extensive experiments on state-of-the-art video generative models, including Wan2.1, Wan2.2~\citep{wan2025}, and Cosmos-2.5~\citep{cosmos-predict2.5}. Across all models, we significantly improve physical realism, such as motion coherence, physical plausibility, and spatial consistency.
Notably, on Wan2.2, which already produces strong human motion, our method further improves motion quality, yielding more than 70\% preference in human evaluation compared with the default sampler.
\section{Related Works}\label{sec:related}

\textbf{Self-Refining in Generative Models\pspace} 
In this paper, we refer to self-refinement as an inference-time paradigm in which a generative model improves its outputs using only its internal signal without any external evaluator, teacher, verifier, or additional training.
In language modeling, Self-Refine~\citep{madaan2023self} proposes an iterative loop in which the model critiques and revises its own outputs. Reasoning with Sampling~\citep{karan2025reasoning} introduces a MCMC-based sampling scheme that uses only the base language model to elicit strong reasoning performance without reinforcement learning. 
In diffusion models, Zigzag-Diffusion~\citep{bai2024zigzag} proposes a self-reflective sampling method that alternates between guided denoising and inversion during inference. 
% To the best of our knowledge, our work is the first to use self-refining for video generation.

\textbf{Improving Physical Realism in Video Generation\pspace}
Previous works explored improving motion coherence in video generative models~\citep{shi2024motion-i2v, wu2024freeinit, chefer2025videojam,shaulov2025flowmo}. VideoJAM~\citep{chefer2025videojam} introduces a joint training approach with an additional optical flow denoising objective. FlowMo~\citep{shaulov2025flowmo} proposes a training-free guidance method to reduce temporal variance. However, these methods require substantial computational cost in training or inference, and still struggle with complex motions. 

Recent work aims to improve physical fidelity in world simulation~\citep{brooks2024sora,genie3,wiedemer2025videomodelszeroshotlearners}. One line of research trains models on curated physics datasets~\citep{zhang2025thinkbefore, wang2025wisa,li2025pisa} or domain-specific datasets~\citep{gosselin2025ctrl,gillman2025force}. For example, WISA~\citep{wang2025wisa} uses a physics-focused MoE, and \citet{zhao2025synthetic} trains a model with synthetic computer-generated imagery (CGI) data. While effective, these approaches necessitate extensive data curation and additional training. Another line of work bypasses large-scale training by employing external physics-aware modules at inference time~\citep{Lv2023GPT4MotionSP,yang2025vlipp,savantaira2024motioncraft,liu2024physgen,wang2025physctrl}. GPT4Motion~\citep{Lv2023GPT4MotionSP}, PhysGen~\citep{liu2024physgen}, and VLIPP~\citep{yang2025vlipp} leverage LLMs as a high-level physics planner, but dependence on external modules can limit generalization.

\section{Preliminaries: Flow Matching in Video Diffusion Models}\label{sec:flow_matching}

Recent video generative models~\citep{polyak2025moviegencastmedia, wan2025, kong2024hunyuanvideo, hacohen2024ltx, jin2025pyramidal} adopt flow matching~\citep{lipman2022flow} in a VAE latent space. Specifically, an RGB video $x \in \mathcal{X}=\mathbb{R}^{F\times H\times W\times 3}$ is first encoded by a video VAE into a compressed latent representation $z \in \mathcal{Z}=\mathbb{R}^{f\times h\times w\times c}$, where $(f,h,w)$ denote the downsampled spatio-temporal resolution and $c$ is the latent channel dimension. This latent space significantly reduces computational cost while preserving the essential spatio-temporal structure of the input video. 

On this latent space, flow matching learns a time-dependent vector field model $u_{\theta}: \mathcal{Z} \times [0,1] \rightarrow \mathcal{Z}$ that transforms samples from a prior distribution $p_0=\mathcal{N}(0,\mathbf{I})$ to the target data distribution $p_1$ via an ordinary differential equation (ODE) $\frac{d z_t}{dt} =u_\theta(z_t,t)$. Samples are generated by solving the ODE over discretized timesteps $0 \!=\! t_0 < \cdots < t_T \!=\! 1$:
\begin{equation}
    z_{t_{i+1}} = z_{t_i} + (t_{i+1} - t_i)~u_{\theta}(z_{t_i}, t_i),\label{eq:euler}
\end{equation}
where $u_\theta$ is the learned vector field and $z_{t_0}$ is an initial point sampled from the prior distribution $p_0$.
A common training strategy constructs a straight path $z_t=(1-t)z_0+tz_1$ between paired samples $z_0\sim p_0$ and $z_1 \sim p_1$, with the target vector field $v_t = z_1 - z_0$. 
The vector field model $u_\theta$ is trained to approximate the vector field $v_t$:
\begin{equation}
\mathcal{L}_{\text{FM}}(\theta)
= \mathbb{E}_{t, z_0, z_1}
\bigl[\|u_\theta(z_t, t) - (z_1 - z_0)\|_2^2\bigr].\label{eq:fm_train}
\end{equation}

\section{Self-Refining Video Sampling}

% \subsection{Motivation}\label{sec:motivation}
% \textbf{Early Lock-in in Flow Matching Video Sampling\pspace} 
% Recent flow matching-based video generators~\citep{wan2025,kong2024hunyuanvideo} achieve strong visual quality, but struggle with complex motion and physical interactions. Due to the deterministic and nearly straight sampling trajectory of flow matching, it is difficult to make corrections once the direction is set in the early stage. 
% In particular, these models largely determine the temporal dynamics of video generation, including motion and physics, in the first few steps~\citep{chefer2025videojam,shaulov2025flowmo,jang2025frameguidance}, and naively increasing the number of inference steps fails to correct the inconsistent dynamics.

\subsection{Flow Matching as Denoising Autoencoder}
To enable self-refinement for flow matching-based video models, we revisit the connection between diffusion models and denoising autoencoders (DAEs)~\citep{vincent2011connection,song2019ncsn}, and extend the link to interpret flow matching as a DAE from a training objective perspective.

The flow matching objective (Eq.~\eqref{eq:fm_train}) can be rewritten as:
\begin{equation}
    \mathcal{L}_{\text{FM}}(\theta) = \mathbb{E}_{t, z_0, z_1}
\left[\frac{1}{(1-t)^2}\left\| \hat{z}_1^{\theta} -z_1 \right\|_2^2\right],
\label{eq:fm_dae}
\end{equation}
where $\hat{z}_1^{\theta} \coloneqq z_t + (1-t)\,u_\theta(z_t,t)$ represents the model prediction of the clean data $z_1$. 
Notably, Eq.~\eqref{eq:fm_dae} corresponds to the weighted version of the generalized DAE objective~\citep{bengio2013generalizeddae}:
\begin{align}
    \mathcal{L}_{\text{DAE}}(\theta) = \mathbb{E}_{t, z_0, z_1}
\bigl[\left\| \hat{z}_1^{\theta} -z_1 \right\|_2^2\bigr],
\end{align}
for which the model learns to denoise the corrupted input $z_t$ back to the clean sample $z_1$. 

Therefore, the flow matching objective can be interpreted as training a time-conditioned DAE across all noise levels. At inference time, for any fixed $t$, the denoising via the flow matching model acts as a DAE reconstruction at that noise level. We leverage the pseudo-Gibbs Markov chain of generalized DAE~\citep{bengio2013generalizeddae}, alternating the corruption and reconstruction at each discretized inference timestep to steer predictions toward the data manifold. Building on this, we introduce a novel sampling method based on the \emph{iterative refinement} of $z_t$ for each timestep $t$.

\subsection{Predict-and-Perturb (P\&P)}\label{sec:pnp}
In the DAE perspective, we first define the reconstruction and corruption operators for the flow matching model. At timestep $t$, the reconstruction from state $z_t$ corresponds to the denoiser $D_{\theta}(\cdot, t)$:
\begin{equation}
    \text{\textbf{Predict:}}\pspace D_{\theta}(z_t, t) \coloneqq z_t + (1 - t)\, u_\theta(z_t,t), 
\label{eq:z_1_predictor}
\end{equation}
where $u_{\theta}$ is the trained vector field model, for which $D_{\theta}$ maps the noisy state $z_t$ to a prediction of the clean sample $\hat{z}_1$. Moreover, the corruption of state $z$ at timestep $t$ corresponds to the linear interpolation with the noise $\epsilon \sim\! \mathcal{N}(0,\mathbf{I})$: 
\vspace{-0.03in}
\begin{equation}
    \text{\textbf{Perturb:}}\pspace R_{\epsilon}(z, t) \coloneqq t z + (1 - t)\epsilon,
\label{eq:perturb}
\end{equation}
\vspace{-0.03in}
where $R_{\epsilon}$ adds noise $\epsilon$ to the sample $z$ with noise level $t$. 

With Predict and Perturb operators, we iteratively refine the state $z_t$ at a fixed noise level $t$, producing a sequence $\{ z^{(k)}_t\}$ via pseudo-Gibbs sampling, similar to the generalized DAE~\citep{bengio2013generalizeddae}. Each iteration consists of a reconstruction step (Predict) followed by a corruption step (Perturb) as follows: 
\begin{equation}
\begin{aligned}
    \hat{z}^{(k)}_1 \coloneqq D_{\theta}\big( z^{(k)}_t , t \big), \; \; \;
    z^{(k+1)}_t \!\coloneqq R_{\epsilon_k} \big(\hat{z}^{(k)}_1, t \big), 
\label{eq:pnp}
\end{aligned}
\end{equation}
with initial state $z^{(0)}_t = z_t$ and $\epsilon_k \!\sim\! \mathcal{N}(0,\mathbf{I})$. 

%%%%%%%%%%%%%%%%%%%%%%%%%%%%%%%%%%%%%%%%%%%%%%%%%%%%%%%%%%%%%
\begin{figure}
\begin{minipage}[t]{0.55\linewidth}
    \begin{minipage}[h!]{1\linewidth}
        \includegraphics[width=1\linewidth]{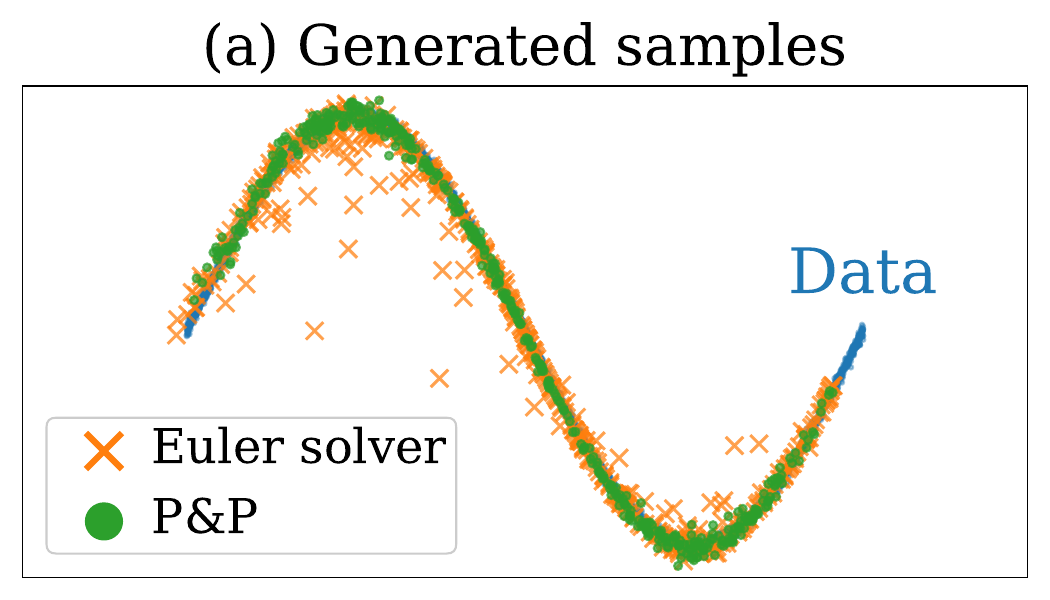}
    \end{minipage}
    \begin{minipage}[h!]{1\linewidth}
        \includegraphics[width=1\linewidth]{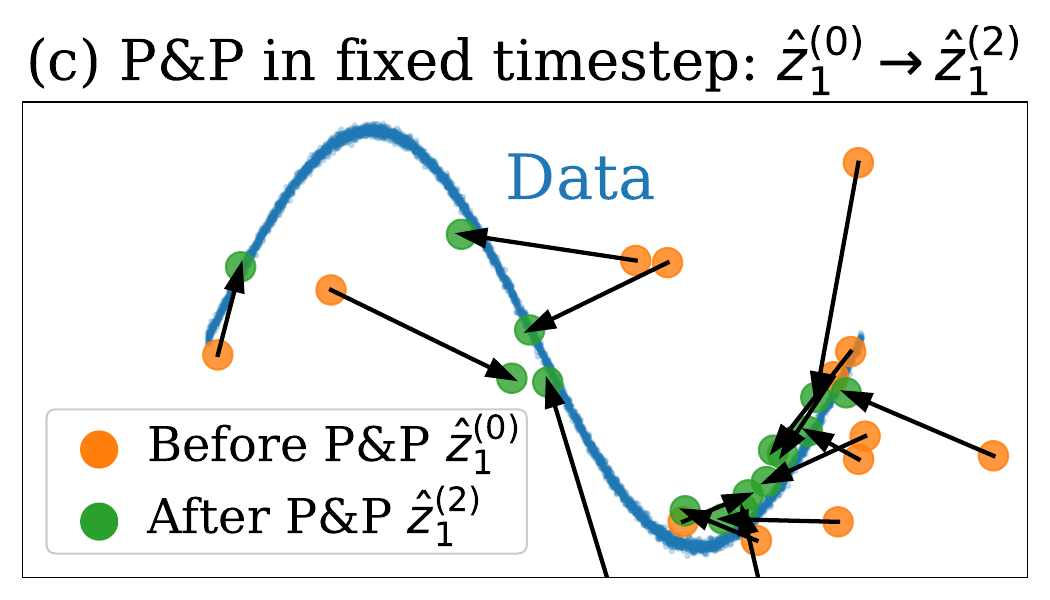}
    \end{minipage}
\end{minipage}
\hfill
\begin{minipage}[t]{0.44\linewidth}
    \begin{minipage}[h!]{1\linewidth}
        \vspace{1.5mm}
        \includegraphics[width=1\linewidth]{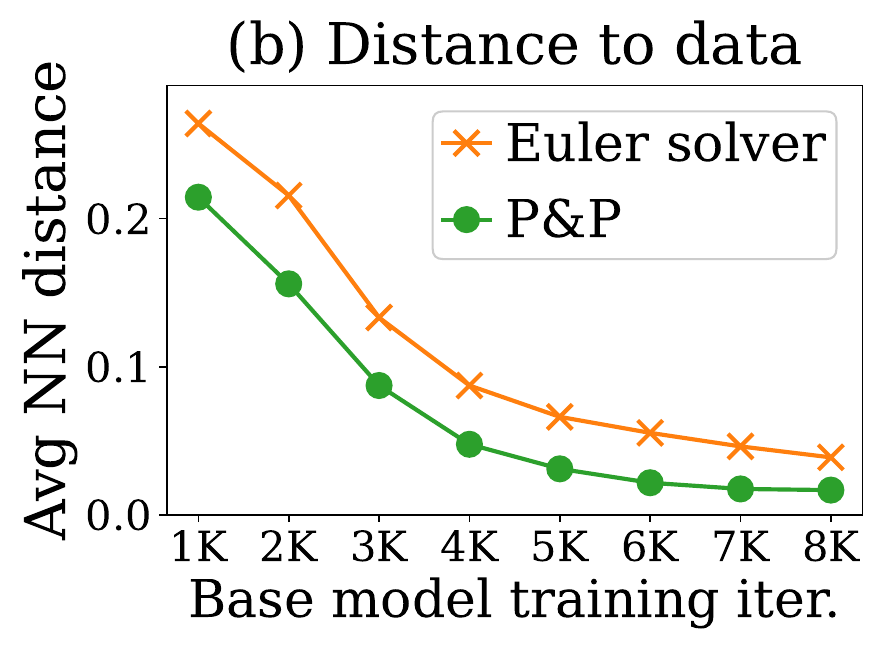}
    \end{minipage}
    \begin{minipage}[h!]{1\linewidth}
        \vspace{-1mm}
        \includegraphics[width=1\linewidth]{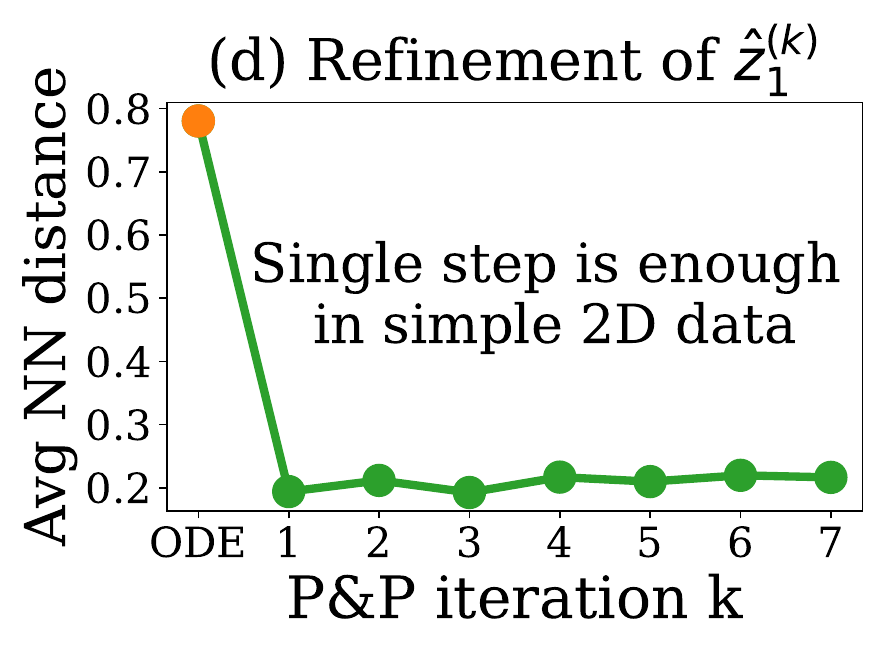}
    \end{minipage}
\end{minipage}
\vspace{-1mm}
\caption{Sampling comparison on a 2D synthetic dataset. \textbf{(a-b)} P\&P generates samples closer to the data manifold than the Euler solver. \textbf{(c-d)} With a fixed timestep, iterative P\&P pulls the prediction $\hat{z}_1$ closer to the data manifold.
% ~\jo{Remove $N$, change (b) to Base Model Training Iter.
% }
}\label{fig:toy}
\vspace{-0.15in}
\end{figure}
%%%%%%%%%%%%%%%%%%%%%%%%%%%%%%%%%%%%%%%%%%%%%%%%%%%%%%%%%%%%%

Conceptually, each Predict-Perturb cycle steers the reconstruction $\hat{z}_1$ toward regions of higher-density~\citep{bengio2013generalizeddae}, yielding a refined state $z_t$. We define a single refinement iteration, termed \textbf{Predict-and-Perturb (P\&P)}, as:
\vspace{-0.15in}
\begin{equation}
    z_t^{(k+1)} = \operatorname{P\&P}_{\epsilon_k}\big( z^{(k)}_t, t \big) \coloneqq R_{\epsilon_k}\big(D_{\theta}(z^{(k)}_t , t), t \big),
\end{equation}
which forms a self-refinement loop using only the generator's signal, without any external model or verifier. 
In this self-refine loop, \emph{Predict} corresponds to the correction of the noisy state via the denoiser, while \emph{Perturb} performs local resampling at the same noise level $t$. In particular, local resampling allows larger exploratory moves at early timesteps, thereby mitigating early lock-in in video generation, where temporal dynamics such as motion and physics are largely determined in the first few steps~\citep{chefer2025videojam,shaulov2025flowmo,jang2025frameguidance}. We empirically find that only 2-3 updates of $z_t$ are sufficient to improve temporal coherence and physical plausibility of the prediction $\hat{z}_1$, even for high-dimensional video latents.

Notably, the proposed P\&P can be integrated into existing ODE solvers in a plug-and-play manner, by simply replacing $z_t$ with the refined $z_t^*\coloneqq z^{\mkern-2mu (K_f)}_{t}$ with $K_f\leq 3$:
\begin{equation}
    z_{{t_{i+1}}}
    = z_{t_i}^* + \Delta t \cdot u_\theta(z_{t_i}^*, t), \;\; \Delta t = {t_{i+1}} - {t_i}
    \label{eq:pnp_outer}
\end{equation}
In particular, since coarse motion and structure are largely determined in the first few steps, we experimentally demonstrate that applying P\&P only at early noise levels (i.e., for timesteps $t\!<\!0.2$) suffices to produce refined samples.

\textbf{Toy experiment}\pspace We validate our method with a toy experiment on a simple 2D sine dataset. As shown in \fref{fig:toy}~(a–b), samples generated with P\&P capture the data manifold more faithfully than those from the Euler solver. In addition, \fref{fig:toy}~(c–d) shows that applying P\&P steps within the same timestep pulls $\hat{z}_1$ toward the data manifold.

%%%%%%%%%%%%%%%%%%%%%%%%%%%%%%%%%%%%%%%%%%%%%%%%%%%%%%%%%%%%%
\begin{figure}[t!]
    \centering
    \includegraphics[width=1\linewidth]{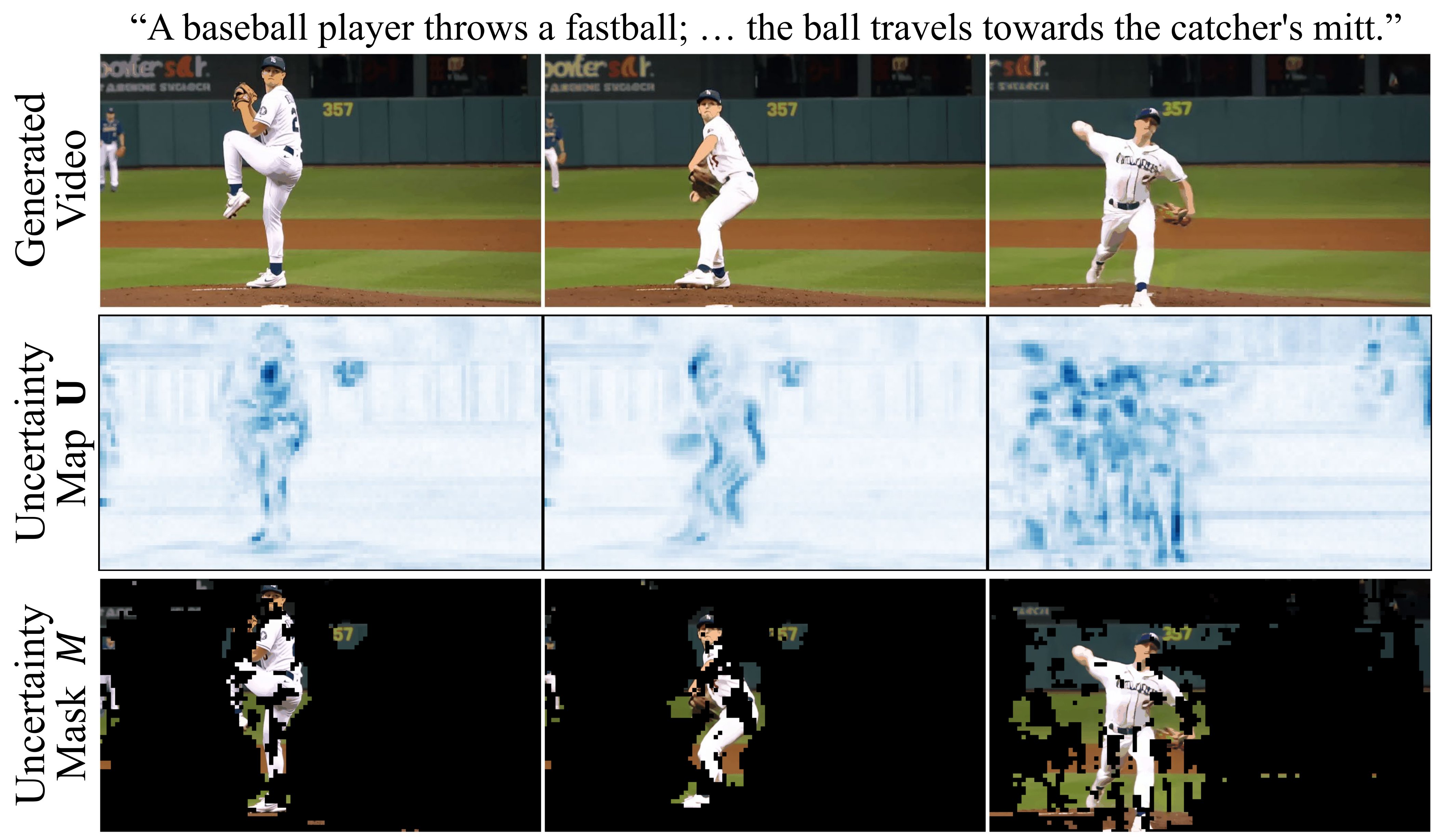}
    \caption{\textbf{Visualization of uncertainty maps}, showing higher values in motion-related regions. Maps are computed at $t = 0.1T$. Bottom row overlays the corresponding binary masks ($\tau=0.25$) on videos generated by Wan2.2-A14B T2V~\citep{wan2025}.
    }\label{fig:uncertainty}
    \vspace{-0.15in}
\end{figure}
%%%%%%%%%%%%%%%%%%%%%%%%%%%%%%%%%%%%%%%%%%%%%%%%%%%%%%%%%%%%%

\subsection{Uncertainty-aware P\&P}\label{sec:uncertainty}
While P\&P enables iterative self-refinement, we observe that applying multiple P\&P updates ($K_f\!>\!3$) with classifier-free guidance (CFG)~\citep{ho2022classifier}, can cause over-saturation~\citep{sadat2024eliminating} or simplification in static regions such as the background, as shown in \fref{fig:ablation}(b).
The issue arises from repeated CFG updates with an amplified scale (i.e., $1\!-\!t$ instead of $\Delta t$) during denoising. 
Regions that are significantly altered by P\&P are less affected by this amplified CFG, as their predictions vary across refinements, limiting guidance accumulation. In contrast, static regions that remain largely unchanged after P\&P are repeatedly influenced by the guidance, causing the guidance to accumulate and leading to over-saturation.

To address this issue, we propose \textbf{Uncertainty-aware P\&P}, an extension of P\&P that selectively refines only the locally uncertain regions. Specifically, we leverage the model confidence of the prediction, applying the P\&P steps only on video regions with low reconstruction confidence.

For each P\&P step, we create an \emph{uncertainty mask} that identifies low-confidence regions, where 1 indicates \emph{uncertain} regions to be refined and 0 marks \emph{confident} regions to be preserved. Specifically, we construct an uncertainty map at the $k$-th refinement step by comparing the reconstructed predictions $\hat{z}_1^{(k)}$ and $\hat{z}_1^{(k-1)}$ from the Predict step:
\begin{align}
\mathbf{U}(z_{t_i}^{(k-1)}, z_{t_i}^{(k)}) \!\coloneqq \frac{1}{C}\| D_{\theta}(z_{t_{i}}^{(k-1)}, t_{i}) - D_{\theta}(z_{t_{i}}^{(k)}, t_{i}) \|_1, 
\notag 
\end{align}
where $C$ denotes the latent channel dimension, and the norm is computed per spatio-temporal location by averaging over channels. The uncertainty mask is then obtained by thresholding the uncertainty map with a confidence threshold $\tau$:
\vspace{-0.15in}
\begin{align}
M^{(k)}_{t_i} \coloneqq \mathbbm{1}\!\left( \mathbf{U}(z_{t_i}^{(k-1)}, z_{t_i}^{(k)}) > \tau\right), 
\label{eq:uncertainty}
\end{align}
where $\mathbbm{1}(\cdot)$ is the indicator function.

As visualized in \fref{fig:uncertainty}, uncertain regions align with moving objects (e.g., human motion) while certain regions correspond to the static background, demonstrating that the model-inherent self-consistency signal identifies regions for refinement. In practice, a fixed threshold $\tau=0.25$ robustly separates the regions.

\definecolor{nfecolor}{rgb}{1.0, 0.498, 0.055}
\begin{figure}[t]
\vspace{-0.15in}
\centering
\begin{minipage}{1.0\linewidth}
\renewcommand{\baselinestretch}{0.95}%\normalsize
\begin{algorithm}[H]
    \caption{Self-Refining Video Sampling}\label{algo:method}
    \textbf{Require}: Timesteps $(t_i)_{i=1}^{T}$, P\&P interval rate $\alpha$,
    confidence threshold $\tau$, number of P\&P iterations~$K_f$.
    
    \begin{algorithmic}[1]
        \STATE \textbf{Sample} Noise $z_{t_0} \sim \mathcal{N}(0, \mathbf{I})$
        \FOR{$i=0\text{ \textbf{to} } T-1$}
            \STATE $z_{t_{i+1}}^{(0)} \leftarrow z_{t_i} + (t_{i+1} - t_{i})~\textcolor{nfecolor}{u_\theta} (z_{t_i}, t_i)$ \COMMENT{\textcolor{nfecolor}{Base NFE}}
            \IF[Motion stage]{$i \leq \alpha T$}
                \STATE \textbf{\textcolor{RoyalBlue}{Predict}} 
                $\hat{z}^{(0)}_1 \leftarrow D_\theta(z_{t_i},t_i)$
                \COMMENT{Eq.~\eqref{eq:z_1_predictor}}
                \FOR{$k=1~\text{\textbf{to}}~ K_f$}
                    \STATE \textbf{\textcolor{RoyalBlue}{Perturb}} $z^{(k)}_{t_i} \leftarrow R_{\epsilon}(\hat{z}_1^{(k-1)})$ \COMMENT{Eq.~\eqref{eq:perturb}}
                    \STATE \textbf{\textcolor{RoyalBlue}{Predict}} $\hat{z}^{(k)}_1 \leftarrow \textcolor{nfecolor}{D_\theta} (z_{t_i}^{(k)})$ \COMMENT{Eq.~\eqref{eq:z_1_predictor}, \textcolor{nfecolor}{+1 NFE}}
                    % \STATE $M_{t_i}^{(k)} \leftarrow \mathbf{U}_\tau (z_{t_i}^{(k-1)}, z_{t_i}^{(k)})$ \COMMENT{Eq.~\eqref{eq:uncertainty}}
                    \STATE $M_{t_i}^{(k)} \leftarrow \mathbbm{1}\!\big( \mathbf{U}(z_{t_i}^{(k-1)}, z_{t_i}^{(k)}) > \tau\big)$ \COMMENT{Eq.~\eqref{eq:uncertainty}}
                    \STATE $z_{{t_{i+1}}}^{(k)} \leftarrow z^{(k)}_{t_i} + (t_{i+1} - t_i)~u_{\theta}(z_{t_i}^{(k)}, t_i)$
                    \STATE $z_{{t_{i+1}}}^{(k)} \leftarrow M_{{t_i}}^{(k)}\odot z_{{t_{i+1}}}^{(k)} + (1 - M_{{t_i}}^{(k)})\odot z_{{t_{i+1}}}^{(k-1)}$
                \ENDFOR
                \STATE $z_{t_{i+1}}\leftarrow z_{t_{i+1}}^{(K)}$ \COMMENT{Refined latent}
            \ELSE
                \STATE $z_{t_{i+1}}\leftarrow z_{t_{i+1}}^{(0)}$ \COMMENT{Base ODE step}
            \ENDIF
        \ENDFOR
    \end{algorithmic}
    \textbf{Output:} $z_{t_T}$
\end{algorithm}
\end{minipage}
\vspace{-0.15in}
\end{figure}

To use the uncertainty mask \emph{without additional NFE}, we introduce a simple technique that performs denoising and mask creation simultaneously. In the Predict step (Eq.~\eqref{eq:z_1_predictor}), we compute the next timestep latent $z_{t_{i+1}}^{(k)}$ from $z_{t_i}^{(k)}$ using already computed $z_{t_{i+1}}^{(k-1)}$ from the previous P\&P iteration.
We reformulate the ODE solver in Eq.~\eqref{eq:pnp_outer} without explicitly computing the refined $z_t^*$:
\begin{equation}
    z_{t_{i+1}}^{(k)}
    \leftarrow
    M_{t_i}^{(k)} \odot z_{t_{i+1}}^{\textcolor{RoyalBlue}{(k)}}
    + (1 - M_{t_i}^{(k)}) \odot z_{t_{i+1}}^{\textcolor{RoyalBlue}{(k-1)}},
    \label{eq:uncertainty_aware}
\end{equation}
where $\odot$ denotes element-wise multiplication. 
Uncertain regions where the mask is set to one are refined via P\&P, correcting physical inconsistencies or jitter artifacts, while certain regions are retained, preventing artifacts from over-refinement.

In~\cref{algo:method}, we summarize the overall procedure of Uncertainty-aware P\&P with an example code implementation provided in \cref{alg:pnp_code}. 
Notably, Lines~5 and~10 in \cref{algo:method} do not incur additional NFEs, as they reuse predictions computed in earlier steps.

% In~\cref{algo:method}, we summarize the overall procedure of Uncertainty-aware P\&P with an example code implementation provided in \cref{alg:pnp_code}. Our method involves three hyperparameters: $\alpha$, $K_f$, and $\tau$. First, $\alpha$ controls the number of ODE timesteps at which P\&P is applied. A small $\alpha$ limits P\&P to the early steps, thereby reducing the additional NFE. $K_f$ controls the refinement strength via the number of iterations per timestep, and increasing $K_f$ corresponds to performing more refinement updates, which increases the additional NFE. Lastly, $\tau$ determines the uncertainty threshold for retaining certain regions during refinement, where we find $\tau=0.25$ to work well. Empirically, overly large $K_f$ can cause saturation, which can be mitigated by increasing $\tau$. We denote additional function evaluations from P\&P in orange color in~\cref{algo:method}. The total increased NFE from using P\&P is $K_f \times |\mathcal{T}_{\text{motion}}|$ where $\mathcal{T}_{\text{motion}}=\{i \mid i \le \alpha T\}$.

\section{Experiments}\label{sec:experiments}

\subsection{Motion Coherence for Challenging Motions}\label{sec:exp_motion}

\textbf{Benchmarks}\pspace
We use two benchmarks to evaluate motion coherence. First, we introduce \emph{Dynamic-bench}, constructed to assess state-of-the-art video generators such as Wan2.2-A14B~\citep{wan2025} under challenging motion scenarios, including multi-object interactions, complex human motions, and physics-driven dynamics. 
Dynamic-bench consists of 120 prompts (40 per category) generated using Gemini~3, with details provided in Appendix~\ref{app:dynamic_bench}. We also evaluate on VideoJAM-bench~\citep{chefer2025videojam}. For both benchmarks, we generate a single video per prompt and evaluate them using VBench~\citep{huang2024vbench}.
To fully assess the fine-grained motion quality of videos that automated evaluation cannot capture, we additionally conduct a human evaluation comparing our method with baselines. Motion quality and text alignment are evaluated on 30 challenging videos using win-tie-lose criteria. An example of the human evaluation is provided in \fref{fig:fig_app_human_eval}, with further details in Appendix~\ref{app:motion_human_detail}.

\textbf{Baselines}\pspace
We use Wan2.1 and Wan2.2 T2V as the base video generators and compare our approach against four inference-time sampling methods: the default ODE solver UniPC~\citep{zhao2023unipc}, the same solver with doubled function evaluations (NFE$\times 2$), CFG-Zero~\citep{fan2025cfgzero}, an improved classifier-free guidance variant for flow matching models, and FlowMo~\citep{shaulov2025flowmo}, a gradient-based training-free guidance method for coherent motion.

\begin{figure*}[p]
    \centering
    \includegraphics[width=1\linewidth]{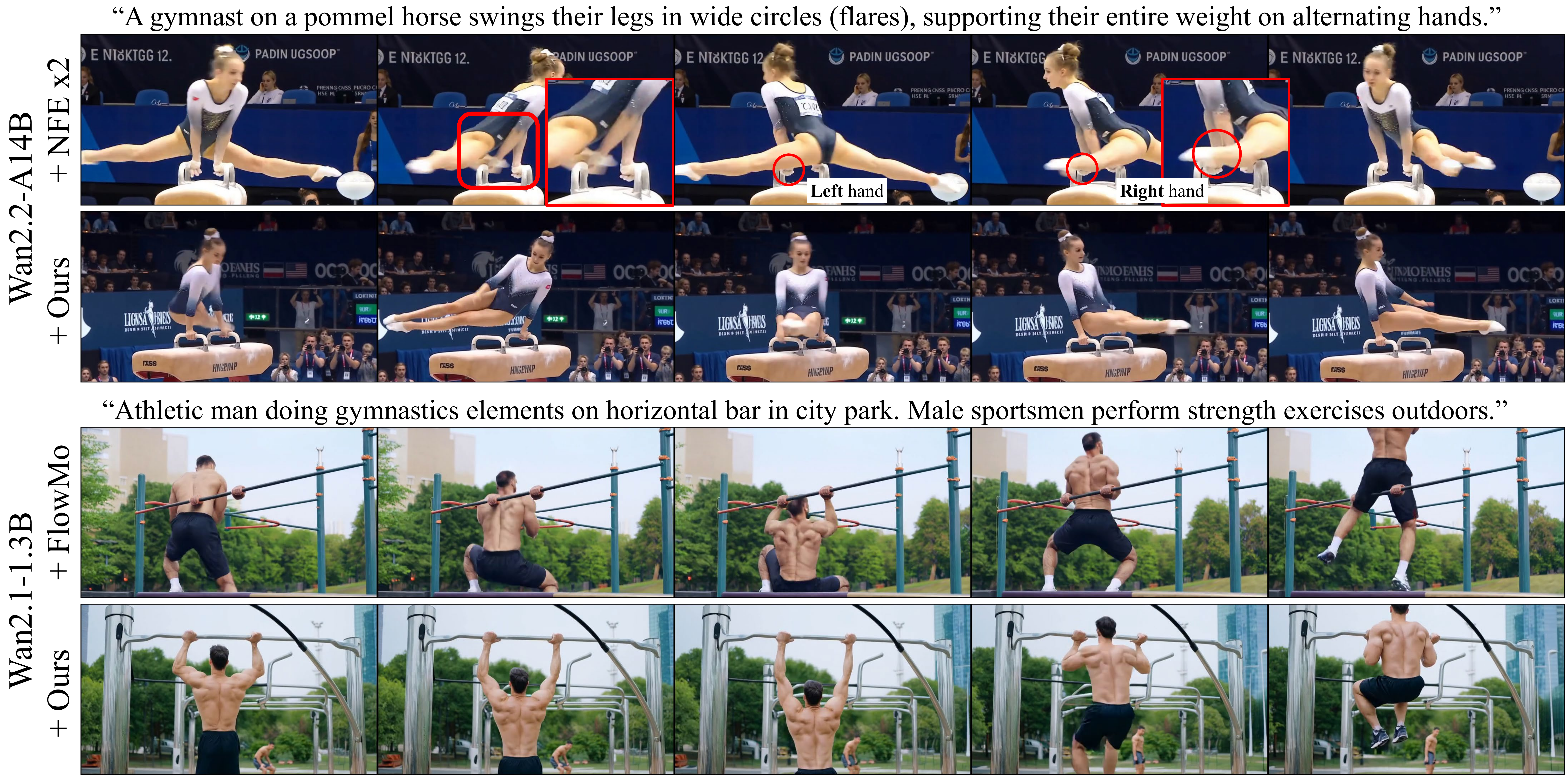}
\vspace{-6mm}
    \captionof{figure}{Qualitative comparison on challenging motion generation.
    % Video examples are provided in the supplementary materials.
    }\label{fig:qual_motion}
\vspace{2mm}
    \centering
    \includegraphics[width=1.0\linewidth]{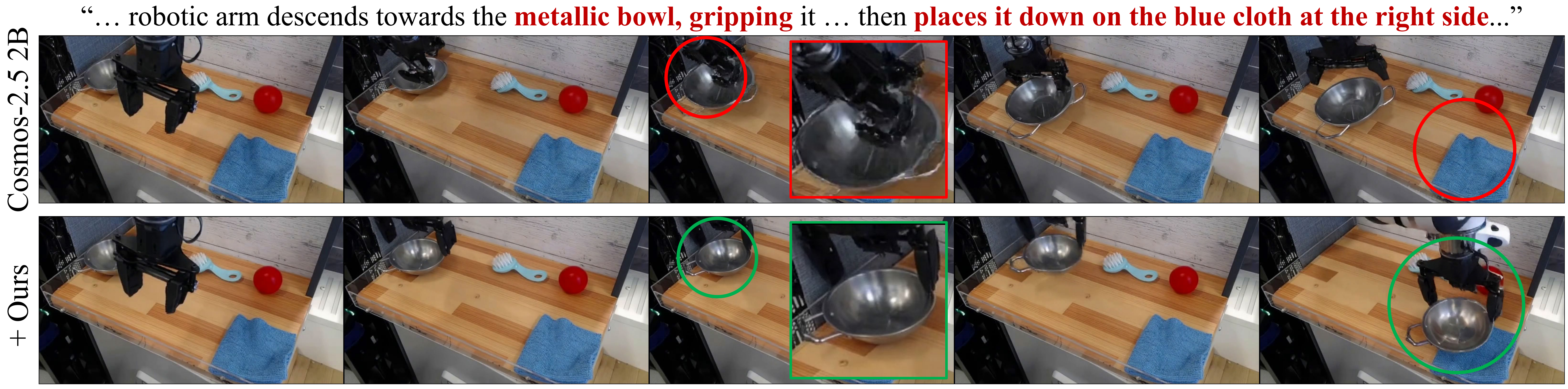}
\vspace{-6mm}
    \captionof{figure}{Qualitative comparison on I2V generation in robotics domain.
    % Video examples are provided in the supplementary materials.
    }\label{fig:qual_robot}
\vspace{2mm}
    \centering
    \includegraphics[width=1\linewidth, height=1.5in]{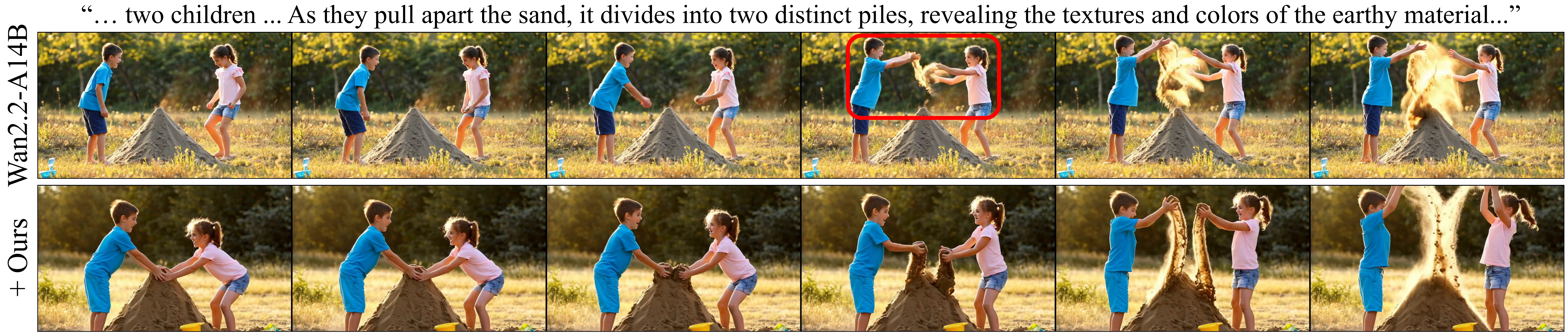}
\vspace{-6mm}
    \captionof{figure}{Qualitative comparison on physics-aligned video generation.
    % Video examples are provided in the supplementary materials.
    }\label{fig:qual_physics}
\vspace{2mm}
    \centering
    \includegraphics[width=1\linewidth]{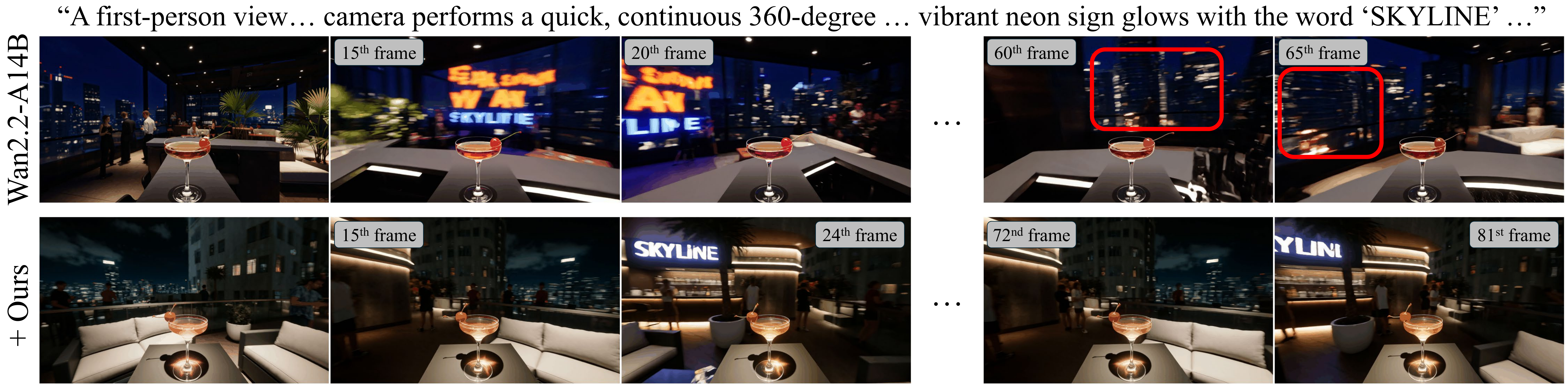}
    \vspace{-6mm}
    \captionof{figure}{Qualitative comparison on spatially consistent video generation.
    % Video examples are provided in the supplementary materials.
    }\label{fig:qual_spatial_const}
\end{figure*}

\textbf{Qualitative Results}\pspace
As shown in \fref{fig:qual_motion}, our method produces videos with significantly enhanced motions even for complex dynamics. For instance, the first row of \fref{fig:qual_motion} shows failed gymnastic motion generated by the ODE sampler even with doubled NFE, exhibiting duplicated arms highlighted in red boxes and physically implausible poses. In contrast, our method (second row of \fref{fig:qual_motion}) produces successful motion, including realistic poses and plausible interactions between the hands and the pommel. We provide additional frames in \fref{fig:fig_app_motion} of the Appendix.

%%%%%%%%%%%%%%%%%%%%%%%%%%%%%%%%%%%%%%%%%%%%%%%%%%%%%%%%%%%%%%%%%%%%%%%%
\begin{table}[t!]
    \centering
    \resizebox{1.0\linewidth}{!}{
    \renewcommand{\arraystretch}{1.}
    \renewcommand{\tabcolsep}{3.5pt}
    \begin{tabular}{lcccccc}
        \toprule
         & \multicolumn{2}{c}{Human Eval} & \multicolumn{2}{c}{VBench} & & \\
         \cmidrule(l{2pt}r{2pt}){2-3}
         \cmidrule(l{2pt}r{2pt}){4-5}
        Method & Motion (\%) & Text (\%) & Motion $\uparrow$ & Const. $\uparrow$& NFE & Time \\
        \midrule
         Wan2.2 T2V & 73.57 & 57.64 & 98.01 & 90.68 & 40 & \\
         + NFE$\times 2$ & 74.05 & 57.55 & 98.03 & 90.66 & 80 & $2.0\times$\\
         + CFG-Zero & 81.53 & 65.71 & 98.27 & 91.16 & 40 & $1.0\times$\\
         + FlowMo & 70.57 & 61.71 & 97.68 & 90.95 & \phantom{*}40\textcolor{blue}{*} & $3.9\times$\\
         \rowcolor{RoyalBlue!05}
         + Ours & - & -  & \textbf{98.41} & \textbf{91.33} & 60 & $1.5\times$ \\
        \bottomrule
    \end{tabular}}
    \vspace{0.5mm}
        \caption{\textbf{Dynamic-bench} results measuring motion coherence for challenging motions using Wan2.2-A14B T2V. Human evaluation shows the percentage of votes favoring ours. Additional inference time (\textcolor{blue}{*}) of FlowMo is introduced by gradient computation.}\label{table:motion_wan22}
    \vspace{-0.3in}
\end{table}

%%%%%%%%%%%%%%%%%%%%%%%%%%%%%%%%%%%%%%%%%%%%%%%%%%%%%%%%%%%%%%%%%%%%%%%%

%%%%%%%%%%%%%%%%%%%%%%%%%%%%%%%%%%%%%%%%%%%%%%%%%%%%%%%%%%%%%%%%%%%%%%%%

\textbf{Quantitative Comparison}\pspace
Tab.~\ref{table:motion_wan22} left shows the human evaluation results from 20 evaluators, reporting the tie-adjusted win rate of our method, where each tie is counted as half a win. The motion quality of our videos is strongly preferred over all other methods, with 73\% favoring ours over the default sampler and 70\% favoring ours over the training-free guidance method FlowMo. 
We provide full human evaluation results in \fref{fig:fig_app_human_eval_motion}.

In Tab.~\ref{table:motion_wan22} right, we present the automated evaluation results on the Dynamic-bench, where our method achieves the strongest performance on VBench metrics, including motion and consistency. We further provide the VideoJam-bench results in Tab.~\ref{table:motion-videojam}, where our method achieves the best scores.

%%%%%%%%%%%%%%%%%%%%%%%%%%%%%%%%%%%%%%%%%%%%%%%%%%%%%%%%%%%%%%%%%%%%%%%%

\subsection{Physical Realism in Robotics Videos}\label{sec:exp_robotics}

\textbf{Benchmarks}\pspace
We evaluate our method on PAI-Bench~\citep{zhou2025paibench} using its predefined VQA questions and VBench quality scores. We generate videos for 174 Robot-domain prompts with three random seeds each, and assess them with Qwen2.5-VL-72B-Instruct~\citep{bai2025qwen25vl}. To assess detailed physical coherence, we additionally report grasp success rates for videos generated from 155 grasp-related prompts, focusing on contact and object manipulation. These are evaluated by Gemini 3 Flash~\citep{gemini3}, which supports higher resolution video inputs. We provide further details in Appendix~\ref{app:details-robotics}.

\textbf{Baselines}\pspace
We use post-trained Cosmos-Predict2.5-2B~\citep{cosmos-predict2.5} and Wan2.2-A14B I2V as the base video generators, and compare our approach against inference-time sampling methods. We additionally compare with a verifier-based rejection sampler using Cosmos-Reason1 7B~\citep{azzolini2025cosmosreason1} as the video critic. It generates four samples per video and selects the sample with the highest non-anomalous score (best-of-4).

\textbf{Qualitative Results}\pspace
As shown in \fref{fig:qual_robot}, our method generates videos that are aligned with the text prompt and exhibit realistic physical interactions with reduced artifacts. As visualized in the top row of \fref{fig:qual_robot}, samples from the base ODE solver often show noticeable grasping artifacts (red box), and fail to move the bowl onto the blue cloth as specified in the prompt. In contrast, samples from our method closely follow the instructions and achieve accurate grasping.

\textbf{Quantitative Results}\pspace
Tab.~\ref{table:cosmos} shows that our method outperforms all baselines on both video generators while incurring only moderate computational overhead. Compared to the base ODE sampler, ours significantly improves the grasp success rate by \textbf{+11.0\%} on Cosmos and \textbf{+8.4\%} on Wan. Ours also outperforms verifier-based rejection sampling (best-of-4), which requires additional inference cost and depends on an external verifier. Moreover, our method achieves the highest Robot-QA accuracy, indicating improved prompt alignment. The quality score, averaged over the VBench, shows negligible variation as all methods perform I2V generation using the same generator.
%%%%%%%%%%%%%%%%%%%%%%%%%%%%%%%%%%%%%%%%%%%%%%%%%%%%%%%%%%%%%%%%%%%%%%%%
% \begin{table}[t!]
%     \centering
%     \resizebox{1.0\linewidth}{!}{
%     \renewcommand{\arraystretch}{1.}
%     \renewcommand{\tabcolsep}{2pt}
%     \begin{tabular}{lccccc}
%         \toprule
%         Method &  Robot-QA$\uparrow$ & Hard-QA$\uparrow$ &  Grasp$\uparrow$ & NFE & Time \\
%         \midrule
%         Cosmos-Predict-2.5 &  71.7 & 75.2  & 79.2 & 36 & \\
%         + NFE$\times 2$ &  72.6 &  00.0  & 78.6 & 72 & $2.0\times$\\
%         {+ Verifier (best-of-4)} &  72.3 &   00.0  & 84.4 & \phantom{*}{144}\textcolor{blue}{*} & $4.2\times$ \\
%         \rowcolor{RoyalBlue!05}
%         + Ours & \textbf{76.3} & \textbf{79.8}  & \textbf{89.6} &  58  & $1.6\times$ \\
%         \midrule
%         Wan2.2-I2V-A14B  & 77.4 &  00.0  & 77.3 & 40 & \\
%         + NFE$\times 2$  & 76.7 &  00.0 & 83.1 & 80 & $2.0\times$ \\
%         {+ Verifier (best-of-4)}  & 78.1 &  00.0  & 80.5 & \phantom{*}{160}\textcolor{blue}{*} &  $4.1\times$\\
%         \rowcolor{RoyalBlue!05}
%         + Ours & \textbf{80.3} &  00.0  & \textbf{85.7} &  60 & $1.5\times$ \\
%         \bottomrule
%     \end{tabular}}
%     \vspace{+1mm}
%         \caption{PAI-Bench-G~\citep{zhou2025paibench} evaluation results on robotics I2V generation. 
%         % Additional time $\alpha$ denotes the extra inference overhead introduced by the external verifier.
%         Using verifier introduces additional inference time (\textcolor{blue}{*}).
%         }\label{table:cosmos}
%     \vspace{-0.23in}
% \end{table}

% Gemini version
\begin{table}[t!]
    \centering
    \resizebox{1.0\linewidth}{!}{
    \renewcommand{\arraystretch}{1.}
    \renewcommand{\tabcolsep}{4pt}
    \begin{tabular}{lccccc}
        \toprule
        Method & Grasp$\uparrow$ & Robot-QA$\uparrow$ & Quality$\uparrow$ &  NFE & Time \\
        \midrule
        Cosmos-Predict-2.5 & 79.2 & 71.7 & 75.1  & 35 & \\
        + NFE$\times 2$ & 78.6 & 72.6 & 75.1  & 70 & $2.0\times$\\
        {+ Verifier (best-of-4)} & 84.4 & {72.3} & 75.3  & {140} & $4.0\times$ \\
        \rowcolor{RoyalBlue!05}
        + Ours & \textbf{89.6} & \textbf{76.3} & 75.1  & 57  & $1.6\times$ \\
        \midrule
        Wan2.2-I2V-A14B & 77.3 & 77.4 & 75.3  & 40 & \\
        + NFE$\times 2$ & 83.1 & 76.7 & 75.5 & 80 & $2.0\times$ \\
        {+ Verifier (best-of-4)} & 80.5 & {78.1} & {75.3}   & {144} &  $4.0\times$\\
        \rowcolor{RoyalBlue!05}
        + Ours & \textbf{85.7} & \textbf{80.3} & 75.5  & 60 & $1.5\times$ \\
        \bottomrule
    \end{tabular}}
    \vspace{+1mm}
        \caption{\textbf{PAI-Bench-G} evaluation results on robotics I2V generation. Grasp is measured by Gemini 3 Flash, and Robot-QA is measured by Qwen2.5-VL-72B.
        }\label{table:cosmos}
    \vspace{-0.23in}
\end{table}

\newcolumntype{P}[1]{>{\centering\arraybackslash}p{#1}}
\begin{table}[t!]
    \centering
    \resizebox{1.0\linewidth}{!}{
    \renewcommand{\arraystretch}{1.0}
    \renewcommand{\tabcolsep}{5pt}
    \begin{tabular}{lccccccc}
        \toprule
         & \multicolumn{4}{c}{VideoPhy2} & \multicolumn{3}{c}{PhyWorldBench} \\
         & \multicolumn{2}{c}{Human Eval} & \multicolumn{2}{c}{Gemini3-F} & \multicolumn{3}{c}{Gemini3-F} \\
         \cmidrule(l{2pt}r{2pt}){2-3}
         \cmidrule(l{2pt}r{2pt}){4-5}
         \cmidrule(l{2pt}r{2pt}){6-8}
        Method & PC (\%) & SA (\%) & PC $\uparrow$ & SA $\uparrow$ & PC $\uparrow$ & SA $\uparrow$ & Both $\uparrow$  \\
        \midrule
         Wan2.2 T2V & 84.29 & 65.24 & 54.5 & 66.1  & 29.3 & 78.1 & 28.6 \\
         + NFE$\times 2$ & 74.76 & 64.29 & 53.1 & 61.7 & 31.4 & \textbf{81.4} & 31.4 \\
         + CFG-Zero & 78.10 & 59.76 & 50.6 & 67.0 & 29.3 & 80.1 & 29.3 \\
         \rowcolor{RoyalBlue!05}
         + Ours & - & -  & \textbf{55.6} & \underline{66.2} & \textbf{40.0} & 78.6 & \textbf{37.9} \\
        \bottomrule
    \end{tabular}}
    \vspace{0.5mm}
        \caption{\textbf{Videophy2} and \textbf{PhyWorldBench} evaluation results using Wan2.2-A14B T2V. Human evaluation shows the percentage of votes favoring ours.}\label{table:physics}
    \vspace{-0.25in}
\end{table}
%%%%%%%%%%%%%%%%%%%%%%%%%%%%%%%%%%%%%%%%%%%%%%%%%%%%%%%%%%%%%%%%%%%%%%%%

\subsection{Physics Alignment in the Wild}\label{sec:exp_physics}

\textbf{Benchmarks}\pspace
We first evaluate on VideoPhy2~\citep{bansal2025videophy2}, which consists of action-centric, physics-related prompts. We generate 360 videos using upsampled captions from the hard and easy subsets, with 180 videos from each. We additionally conduct a human evaluation for a complementary assessment. We further evaluate on PhyWorldBench~\citep{gu2025phyworldbench} using 70 prompts from the kinematics and interaction dynamics domain, generating two samples per prompt. For both benchmarks, we assess physical commonsense (PC) and semantic alignment (SA) using Gemini 3 Flash, which supports higher frame rates.

To demonstrate the improved \emph{consistency} of our method, we use PisaBench~\citep{li2025pisa}, a benchmark designed to assess free-fall I2V generation. 
% by measuring deviation from the ground-truth trajectory
We use the full real dataset for evaluation and additionally generate 32 videos for each of the three selected scenarios with clearly visible objects to analyze failure cases.

%%%%%%%%%%%%%%%%%%%%%%%%%%%%%%%%%%%%%%%%%%%%%%%%%%%%%%%%%%%%%%%%%%%%%%%%
% \begin{table}[t!]
%     \centering
%     \includegraphics[width=1\linewidth, height=1.5in]{example-image-a}
%     \vspace{-5mm}
%         \caption{PISA experiment results with trajectory.}\label{table:pisa}
%     \vspace{-0.17in}
% \end{table}

\begin{figure}[t]
    \centering
        \begin{minipage}{0.45\linewidth}
        \centering
        \resizebox{1.0\linewidth}{!}{
        \renewcommand{\arraystretch}{1.0}
        \renewcommand{\tabcolsep}{2.5pt}
        \begin{tabular}{lccc}
            \toprule
            Method& L2 $\downarrow$ & CD $\downarrow$ & IoU $\uparrow$\\
            \midrule
            Wan2.2 & 0.132 & 0.348 & 0.069 \\
            \rowcolor{RoyalBlue!05}
            + Ours & \textbf{0.128} & \textbf{0.338} & \textbf{0.074} \\
            \bottomrule
        \end{tabular}}
        \vspace{-2mm}
        \caption*{(a) Full real dataset}
        \vspace{1mm}
        \resizebox{1.0\linewidth}{!}{
        \renewcommand{\arraystretch}{1.0}
        \renewcommand{\tabcolsep}{2pt}
        \begin{tabular}{lccc}
            \toprule
            Method& L2 $\downarrow$ & CD $\downarrow$ & IoU $\uparrow$\\
            \midrule
            Wan2.2 & 0.186 & 0.489 & 0.057 \\
            \rowcolor{RoyalBlue!05}
            + Ours & \textbf{0.184} & \textbf{0.482} & \textbf{0.060} \\
            \bottomrule
        \end{tabular}}
        \vspace{-2mm}
        \caption*{(b) Multiple generations on three samples (see right)}
    \end{minipage}
    \hfill
    \begin{minipage}{0.26\linewidth}
        \centering
        \includegraphics[width=1\linewidth]{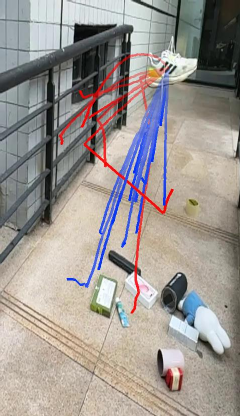}
        \vspace{-6mm}
        \caption*{Wan2.2-I2V}
    \end{minipage}
    \begin{minipage}{0.26\linewidth}
        \centering
        \includegraphics[width=1\linewidth]{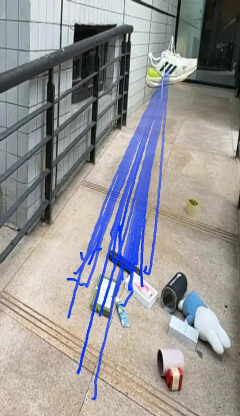}
        \vspace{-6mm}
        \caption*{+ Ours}
    \end{minipage}
    \vspace{1mm}
    \captionof{table}{\textbf{PisaBench} evaluation. (Left) Quantitative results on the full real dataset. (Right) Visualization of 32 generated free-fall trajectories. Physically implausible falls are shown in red.
    % Red trajectories are physically implausible falls.
    }\label{table:pisa}
    \vspace{-0.15in}
\end{figure}

%%%%%%%%%%%%%%%%%%%%%%%%%%%%%%%%%%%%%%%%%%%%%%%%%%%%%%%%%%%%%%%%%%%%%%%%

\textbf{Qualitative Results}\pspace
As visualized in \fref{fig:qual_physics}, ours generates videos following the physical law with fewer visual hallucinations.
For example, in the top row of \fref{fig:qual_physics}, the base model often exhibits non-physical behavior in which sand abruptly appears in the children’s hands without any causal interaction (red boxes). In contrast, our method follows the physical constraints and causal consistency.

We further visualize 32 free-fall trajectories of a free fall in Tab.~\ref{table:pisa} right. While the default ODE solver produces physically implausible falls (red trajectories), our method consistently generates realistic videos of the falling object.

\textbf{Quantitative Comparison}\pspace
Tab.~\ref{table:physics} shows the human evaluation results from 20 evaluators, indicating that the physics alignment of our videos is strongly preferred over all other methods.
In particular, 84\% favor ours over the default sampler, and 74\% favor ours over the doubled-NFE baseline. We provide full human evaluation results in \fref{fig:fig_app_human_eval_physics}.

Moreover, automated evaluation in Tab.~\ref{table:physics} shows that our method outperforms all baselines in physics commonsense (PC) metric on both benchmarks, with larger gains on the motion-centric PhyWorldBench. Results on PisaBench in Tab.~\ref{table:pisa} further validate that our method generates more accurate trajectories in the free-fall experiments.

%%%%%%%%%%%%%%%%%%%%%%%%%%%%%%%%%%%%%%%%%%%%%%%%%%%%%%%%%%%%%%%%%%%%%%%%
\begin{table}[t!]
    \centering
    \resizebox{1.0\linewidth}{!}{
    \renewcommand{\arraystretch}{1.0}
    \renewcommand{\tabcolsep}{9pt}
    \begin{tabular}{lcccc}
        \toprule
         Method & SSIM $\uparrow$ & L1 $\downarrow$ & PSNR (dB) $\uparrow$ & NFE\\
        \midrule
         Wan2.2 T2V & 0.401 & 37.26 & 14.96 & 40\\
        \rowcolor{RoyalBlue!05}
         + Ours & \textbf{0.485} & \textbf{30.16} & \textbf{17.21} & 60 \\
        \bottomrule
    \end{tabular}}
    \vspace{0.4mm}
        \caption{\textbf{Spatial consistency} evaluation results using Wan2.2-A14B T2V. We measure distances between frame pairs at revisited viewpoints after camera-pose-based warping.}\label{table:spatial}
    \vspace{-0.15in}
\end{table}
%%%%%%%%%%%%%%%%%%%%%%%%%%%%%%%%%%%%%%%%%%%%%%%%%%%%%%%%%%%%%%%%%%%%%%%%

%%%%%%%%%%%%%%%%%%%%%%%%%%%%%%%%%%%%%%%%%%%%%%%%%%%%%%%%%%%%%%%%%%%
\begin{figure*}[t!]
    \centering
    \includegraphics[width=1\linewidth]{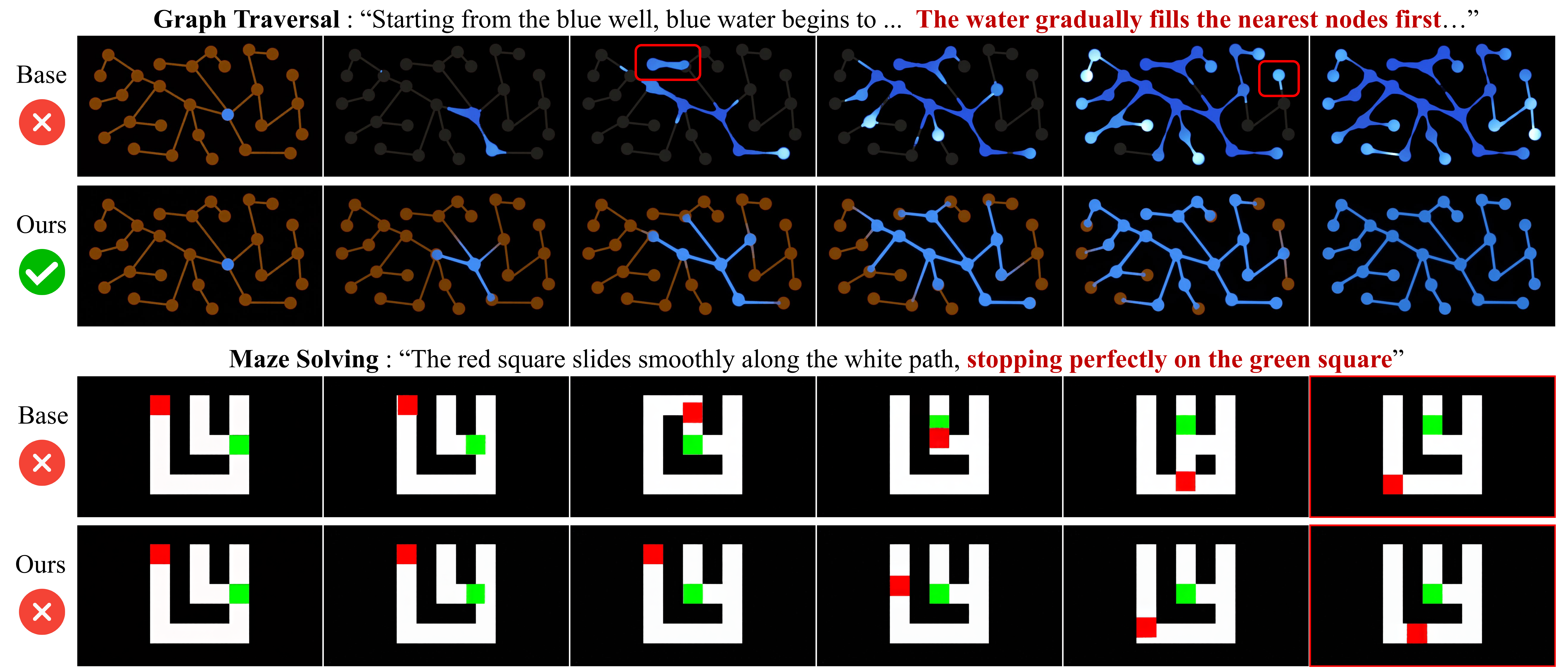}
    \vspace{-0.15in}
    \caption{Examples of \textbf{self-refinement applied to visual reasoning tasks}: (Top) graph traversal and (Bottom) maze solving from \citet{wiedemer2025videomodelszeroshotlearners}. We use Wan2.2-A14B I2V as the base model. For graph traversal, self-refinement yields a dramatic improvement in the success rate from 0.1 to 0.8. For maze solving, self-refinement does not yield meaningful gain, with success remaining near zero.
    }
    \label{fig:reasoning_all}
    \vspace{-0.1in}
\end{figure*}
%%%%%%%%%%%%%%%%%%%%%%%%%%%%%%%%%%%%%%%%%%%%%%%%%%%%%%%%%%%%%%%%%%%

\subsection{Improvement in Spatial Consistency}\label{sec:exp_spatial_consistency}
Moreover, we observe that our self-refinement can improve the spatial consistency of the generated videos. We assess this capability with simple experiments that evaluate videos in which a camera revisits a previously seen viewpoint, for example, after rotations exceeding 360°.

\textbf{Benchmarks}\pspace
We generate videos from 20 prompts generated by Gemini that involve large camera motions. We then estimate per-frame camera parameters with MegaSaM~\citep{li2025megasam} and measure visual distances between frame pairs corresponding to revisited viewpoints with similar estimated camera poses. Specifically, we warp one frame into the other using the estimated depth and camera poses, and measure visual similarity on visible pixels using SSIM, L1, and PSNR. We provide evaluation details in Appendix~\ref{app:details-spatial}.

% The prompts used in this evaluation all include a \emph{first-person view}. Since models frequently produce misaligned videos under such settings, we empirically construct a set of 20 prompts using Gemini, focusing on Minecraft-style scenes where generation success rates are relatively high. Further details are provided in Appendix~\ref{app:details-spatial}.

\textbf{Results}\pspace
\fref{fig:qual_spatial_const} demonstrate that our method generates spatially consistent videos that preserve previously observed scene content even under large camera motions. The top row of \fref{fig:qual_spatial_const} shows that the default ODE solver often produces inconsistent backgrounds that differ from earlier frames when the camera movement is large. In contrast, ours maintains much stronger consistency with earlier viewpoints.
As shown in Tab.~\ref{table:spatial}, our method achieves significantly improved spatial consistency compared to the default ODE solver.

\subsection{Application to Visual Reasoning} \label{sec:exp_reasoning}
We conduct extensive analysis of whether our method can improve the emergent visual reasoning capabilities of recent video generators~\citep{wiedemer2025videomodelszeroshotlearners,cai2025mmgr}. 
First, we find that tasks that can be partially refined through motion or temporal consistency show noticeable improvements with our self-refining video sampling. For example, the graph traversal problem visualized at the top of \fref{fig:reasoning_all} shows a dramatic increase in success rate, from 0.1 to 0.8, after applying self-refinement. Qualitatively, refinement reduces visual artifacts and improves temporal coherence, which leads to correct reasoning trajectories.

However, tasks whose success depends on discrete or semantic correctness show little or no improvement. For example, the maze solving problem at the bottom of \fref{fig:reasoning_all} shows no meaningful gain, with success remaining near zero. We speculate that in cases where the video generator fails almost entirely, it lacks the knowledge needed to correct the underlying errors, and late-stage refinement of the maze trajectory becomes insufficient. In these cases, external verifiers are likely required. We provide more details in Appendix~\ref{app:extending-reasoning-task}.

\subsection{Ablation Studies}\label{sec:exp_ablation}

\begin{figure}[t!]
    \centering
    \includegraphics[width=1\linewidth]{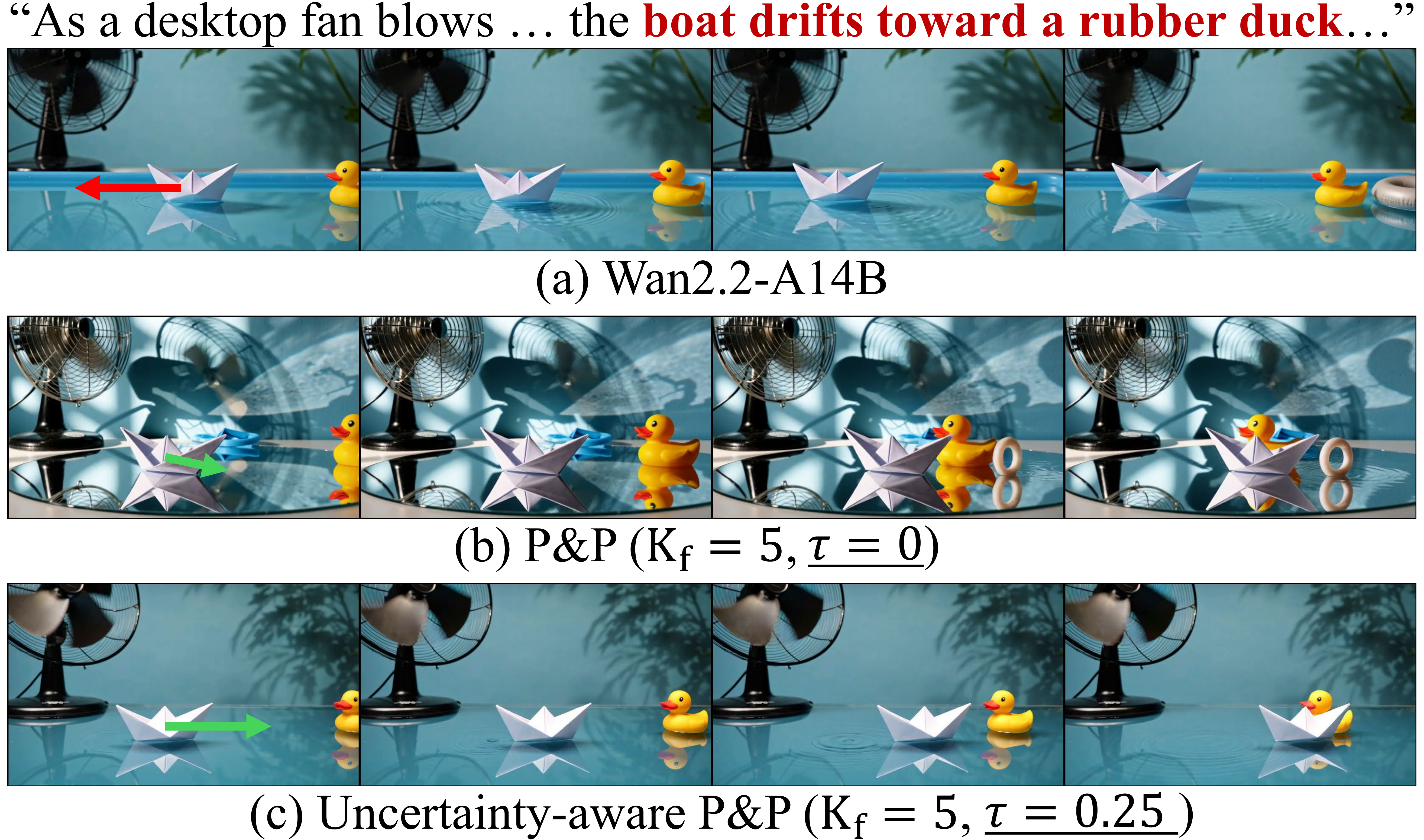}
    \vspace{-5mm}
    \caption{\textbf{Ablation on uncertainty-aware strategy}. Multiple P\&P updates without uncertainty-aware strategy cause over-saturation. Red arrow indicates motion misaligned with the prompt.
    }\label{fig:ablation}
    \vspace{-0.2in}
    % \vspace{-6mm}
\end{figure}

\textbf{Importance of Uncertainty-Aware Refinement}\pspace
As shown in \fref{fig:ablation}(b), excessive P\&P iterations (e.g., $K_f=5$) without our uncertainty-aware strategy lead to over-saturation and simplification. This causes shifts in color tone and contrast as well as exaggerated reflections on the water surface, which is similar to the effect of increasing the CFG scale. The issue can be mitigated by using the uncertainty-aware strategy, which selectively refines the motion-related regions, as visualized in \fref{fig:ablation}(c).

\textbf{Hyperparameters of P\&P}\pspace
We conduct ablation studies on the key hyperparameters of P\&P: number of P\&P iterations $K_f$, confidence threshold $\tau$, and P\&P interval rate $\alpha$. We observe that these hyperparameters remain robust across a wide range of settings. In \fref{fig:fig_app_hyper_ablation} of Appendix, we show that increasing $K_f$ strengthens refinement at the cost of additional NFEs, while $\tau$ regulates background appearance. In \fref{fig:fig_app_hyper_ablation2}, we show that applying P\&P at earlier inference stages is more effective for correcting motion errors, with later stages contributing marginally. We provide further details in Appendix~\ref{app:ablation}.

\section{Discussion}
\subsection{Cross-Frame Consistency of Video}\label{sec:cross-frame}
\vspace{-0.03in}

Here, we discuss a unique property of videos and how it affects our design of self-refinement sampling. Videos are notably more robust to perturbations during generation compared to images, due to \emph{cross-frame consistency}, where neighboring frames share strongly correlated layouts and motion trajectories. 
We illustrate this in \fref{fig:frame_consistency}(a), which applies SDEdit~\citep{meng2022sdedit} with a changed prompt for the image and video. While the image exhibits a clear semantic transition, the video largely preserves its content.

Due to the cross-frame consistency, multiple P\&P updates during video generation produce controlled changes in temporal structures like motion. In contrast, images can shift substantially after a single P\&P update, even when applied at later timesteps of generation.
We visualize this in \fref{fig:frame_consistency}(b), where repeated P\&P iterations lead to large deviations for images, but only minimal changes in the video.
Consequently, iterative P\&P updates for videos act as a local search that refines the latents, rather than a global resampling that resets them and induces large semantic transitions.

\subsection{Mode-Seeking Behavior of Iterative P\&P}
\vspace{-0.03in}
We observe that iterative P\&P exhibits mode-seeking behavior in which samples concentrate in high-density, stable modes of the data distribution. We visualize this in a toy example (\fref{fig:app_pnp_gaussian_mode}) using a 2D Gaussian mixture, where repeated P\&P yield samples concentrated in the high-density regions. Similarly, applying multiple P\&P  ($K_f\!=\!8$) in image generation reduces output diversity and concentrates outputs toward a small number of classes, as shown in \fref{fig:app_pnp_image_mode}.

In video generation, this mode-seeking behavior manifests differently. Rather than collapsing to identical content, refined videos show reduced temporal variance, removing temporal artifacts such as jittering and flickering. 
We hypothesize this difference is due to cross-frame consistency as temporally inconsistent videos lie in low-density regions. Consequently, iterative P\&P appears as temporal mode-seeking, which leads to physically plausible videos.

\subsection{Connection to Prior Works}
\vspace{-0.03in}
%%%%%%%%%%%%%%%%%%%%%%%%%%%%%%%%%%%%%%%%%%%%%%%%%%%%%%%%%%%%
%%%%%%%%%%%%%%%%%%%%%%%%%%%%%%%%%%%%%%%%%%%%%%%%%%%%%%%%%%%%

\textbf{Annealed Langevin Dynamics (ALD)}~\citep{song2019ncsn} is an MCMC sampler that alternates with Gaussian noise injection and score-guided Langevin updates, resembling our iterative perturb-and-predict refinement. However, ALD samples through a sequence of annealed noise scales, whereas our method performs stochastic perturbations and corrections at a fixed noise level within each refinement loop. Moreover, ALD is intended to approximate the target distribution, while our self-refinement is not a strict MCMC sampler and exhibits mode-seeking behavior.

% also difference in perturbation: noise scale

\textbf{Restart}~\citep{xu2023restart} alternates between forward noising restart steps and deterministic backward ODE integration, using stochasticity to reduce error accumulation. At a high level, it resembles our approach in that noise injection is followed by a deterministic update. However, our method differs in when and how stochasticity is applied. While Restart adds noise by jumping forward in time and integrating back along the ODE, we perform local resampling at the same noise level via P\&P. Furthermore, Restart applies a macro forward-backward cycle to reduce accumulated errors that are often positioned late in the trajectory, whereas we perform fine-grained refinement in the same noise-level, typically applied in the early steps.

\textbf{FreeInit}~\citep{wu2024freeinit} is a training-free inference method that improves video temporal consistency by iteratively refining the initial noise, using only the generator. The key difference with our method is where the refining happens. FreeInit refines only the initial noise and re-runs the full denoising process, whereas we iteratively refine the intermediate latents within the same sampling trajectory. Our approach is both more effective and significantly more compute-efficient than repeating the full denoising process.

% In particular, our uncertainty-aware sampling prevents unnecessary over-simplification.

% \textbf{Inference-time Scaling}\pspace
% \jang{Replace it to mode-seeking behavior. In my experience, additional K does not guarantee the better results.}
% We can view P\&P as a form of \emph{inference-time scaling}~\citep{Ma2025inferencetime} that performs self-refinement to improve sample reliability without external reward models or verifiers. Due to cross-frame consistency and the uncertainty-aware update that preserves confident regions, P\&P performs a \emph{local search} over sampling paths, where the trajectory stays close to the one original path and gradually moves toward a more reliable local solution.

% When computation cost is increased without constraint, i.e., with a very large number of refinement iterations $K$, P\&P behaves like \emph{temporal mode seeking}. While the uncertainty-aware update preserves most of the content, motion becomes more stable but also simpler. Since temporal motion diversity is difficult to quantify directly, we analyze this effect in the image domain, where excessive refinement leads to reduced sample diversity. More details in Appendix~\ref{app:inference-time-scaling}.
\section{Conclusion}
% In this work, we present Predict-and-Perturb, a novel inference-time sampling method for reliable sampling with video generators. We discuss the limitations of our method in Appendix~\ref{app:limitations}.
In this work, we present a self-refining video sampling method that reuses a pre-trained video generator as a self-refiner. We revisit the flow matching objective as a generalized denoising autoencoder and leverage it to refine latents at each timestep during inference. We further propose an uncertainty-aware strategy that selectively refines uncertain regions using self-consistency signals from the model itself. Extensive experiments demonstrate that P\&P consistently improves motion coherence, physical plausibility, and overall quality across diverse video generation tasks. We believe this work provides a practical and broadly applicable approach for more effective use of existing pre-trained video generators. We discuss the limitations in Appendix~\ref{app:limitations}.

\section*{Acknowledgements} 
This work was supported by Institute for Information \& communications Technology Planning \& Evaluation (IITP) grant funded by the Korea government (MSIT) (RS-2019-II190075, Artificial Intelligence Graduate School Program (KAIST)) and the Ministry of Science and ICT (MSIT) of the Republic of Korea in connection with the Global AI Frontier Lab International Collaborative Research (No. RS-2024-00469482 \& RS-2024-00509279), National Research Foundation of Korea (NRF) grant funded by the Korea government (MSIT) (No. RS-2023-00256259), the “Advanced GPU Utilization Support Program” funded by the Government of the Republic of Korea (MSIT), Center for Applied Research in Artificial Intelligence (CARAI) grant funded by DAPA and ADD (UD190031RD), and NTU Start Up Grant. Saining Xie acknowledges support from the OpenPath AI Foundation, IITP grant funded by the Korean Government (MSIT) (No. RS-2024-00457882, National AI Research Lab Project) and NSF Award IIS-2443404.

% Chanuk
% This work was supported by Institute for Information \& communications Technology Planning \& Evaluation (IITP) grant funded by the Korea government (MSIT) (RS-2019-II190075, Artificial Intelligence Graduate School Program (KAIST), No.RS-2022-II220713, Meta-learning Applicable to Real-world Problems) and the Ministry of Science and ICT (MSIT) of the Republic of Korea in connection with the Global AI Frontier Lab International Collaborative Research (No. RS-2024-00469482 \& RS-2024-00509279), and National Research Foundation of Korea (NRF) grant funded by the Korea government (MSIT) (No. RS-2023-00256259), and the “Advanced GPU Utilization Support Program” funded by the Government of the Republic of Korea (Ministry of Science and ICT).

\section*{Impact Statement}
This work aims to improve the reliability and physical consistency of video generation models through a training-free inference-time refinement method that enhances motion and temporal coherence. By focusing on principled sampling, our approach complements recent advances in large-scale video generative models without additional supervision or retraining. As with prior work in video generation, such models may be misused to create misleading content, however, our method does not introduce risks beyond those already present in existing systems. We believe that improving temporal stability and physical plausibility is an important step toward more trustworthy video generation.

\newpage
\bibliography{main}
\bibliographystyle{icml2026}

\clearpage
\clearpage
\newpage
\appendix
\begin{center}{\bf {\LARGE Appendix}}\end{center}
\paragraph{Organization} The Appendix is organized as follows: 
% In Sec.~\ref{app:theoretical}, we provide theoretical discussion of our framework. 
We provide experimental details in Sec.~\ref{app:exp_details}
% additional results in Sec.~\ref{app:additional_examples}, 
and further discussion in Sec.~\ref{app:more_discussion}. Lastly, in Sec.~\ref{app:limitations}, we discuss the limitations and future directions of our work.

% \section{Theoretical Discussion}\label{app:theoretical}
% \subsection{Pseudo-Gibbs Sampling} \label{app:gibbs}

\section{Experimental Details} \label{app:exp_details}
\subsection{Base models}
\textbf{Wan~\citep{wan2025}\pspace}
Wan2.1 and Wan2.2 are open-source, flow matching-based video generation models, released in text-to-video (T2V) and image-to-video (I2V) variants.
The I2V model is trained with a first-frame condition, ensuring that the generated video exactly reproduces the input image as the first frame without additional inference techniques.
This strong first-frame constraint provides stable context during sampling, \emph{preventing over-saturation} even under high classifier-free guidance (CFG).
Consequently, multiple P\&P iterations can be applied without uncertainty-aware strategy. For our experiments with Wan2.2 I2V, we therefore disable uncertainty-aware sampling (i.e., $\tau\!=\!0$), allowing unconstrained P\&P refinement.

Wan2.2 improves upon Wan2.1 by incorporating two expert transformer models that are activated based on the flow matching timestep. In addition, Wan2.2 adopts an exponential sampling schedule, allocating more NFEs at high-noise timesteps. This design enhances motion synthesis in early sampling stages. 
Accordingly, when applying P\&P with the time interval $t \le \alpha T$ in Algorithm~\ref{algo:method}, the motion stage is longer than in Wan2.1. 
To ensure a fair comparison, we use a smaller $\alpha$ for Wan2.2 so that the total NFEs do not exceed 1.5$\times$ that of the base sampler.

% \paragraph{Cosmos-Predict-2.5 ~\citep{cosmos-predict2.5}}
\textbf{Cosmos-Predict-2.5 ~\citep{cosmos-predict2.5}\pspace}
We use Cosmos-Predict-2.5-2B post-trained for I2V generation. Empirically, we observe that this model is more prone to over-saturation under high CFG scale compared to Wan I2V. Accordingly, we set the CFG scale to 4 for both the base sampler and P\&P, which we find to be stable with minimal saturation artifacts. To further mitigate over-saturation, we apply uncertainty-aware P\&P with $\tau\!=\!0.5$. More details of hyperparameters are in Sec.~\ref{app:hyperparameters}.

\subsection{Implementation Details}\label{app:implementation_detail}
All experiments are conducted on a single NVIDIA H100 80GB GPU. Notably, while our method increases the NFE, its memory usage remains identical to that of the base sampler. For the Wan series, including the CFG-zero~\citep{fan2025cfgzero} baseline, we primarily use our own implementation built upon the Diffusers~\citep{von-platen-etal-2022-diffusers} Wan pipeline. FlowMo~\citep{shaulov2025flowmo} experiments follow the official implementation, with additional engineering modifications to improve efficiency. Specifically, we incorporate gradient checkpointing~\citep{chen2016checkpointing}, reducing the required hardware from two GPUs to a single GPU while also improving runtime performance. These modifications enable FlowMo to scale to larger models such as Wan2.2. We follow the official setting and used a learning rate of $\eta\!=\!0.005$ for all FlowMo experiments.

For Cosmos, we use the official implementation and the classifier-free guidance scale to $s\!=\!4$ across all experiments. The output video resolution is set to 480p for all Wan-series models and 720p for Cosmos.

\textbf{Detailed Algorithm\pspace}
% We provide detailed algorithm in Algorithm~\ref{algo:method_detail}.
We provide a detailed code-level implementation in Algorithm~\ref{alg:pnp_code}. In practice, the uncertainty mask is accumulated across P\&P iterations (line~11), ensuring that regions identified as certain are frozen and no longer refined in later iterations.
\begin{algorithm}[t]
\small
\caption{A single uncertainty-aware P\&P step (code)}\label{alg:pnp_code}
\begin{lstlisting}[
  numbers=left,
  numberstyle=\tiny,
  numbersep=4pt,
  xleftmargin=1.5em
]
# buffer (previous step): [pred_z1, pred_z_next, m_unc]

noise = &&randn_like&&(buffer[0])
z_t_pnp = t @@*@@ @@@buffer@@@[0] + (1-t) @@*@@ noise # Perturb

flow_pred = &&model&&(z_t_pnp, t, **kwargs) # w/ CFG
pred_z1 = z_t_pnp + (1-t) @@*@@ flow_pred
pred_z_next = z_t_pnp + delta_t @@*@@ flow_pred

uncertainty = &&L1_distance&&(@@@buffer@@@[0], pred_z_1)
m_unc = (uncertainty > tau) | @@@buffer@@@[2]

pred_z1 = m_unc @@*@@ pred_z1 + (1-m_unc) @@*@@ @@@buffer@@@[0]
pred_z_next = m_unc @@*@@ pred_z_next + (1-m_unc) @@*@@ @@@buffer@@@[1]

# in last P&P iteration, return pred_z_next
@@@buffer@@@ = [pred_z1, pred_z_next, m_unc]

\end{lstlisting}
\end{algorithm}

% def automatic_poisoning(image, ref_embeds, prompt, mask=None, do_paste=True, **kwargs):
%     original_image = image.copy()
%     # mask generation stage
%     &&for&& _ in range(NUM_MASK_TRIAL):
%         image = iterative_sdedit(image, prompt, **kwargs)
%         success, mask, pasted = logo_detection(image, ref_embeds, **kwargs)
%         &&if&& success:
%             &&break&&
    % # determine as challenging image
    % &&if&& not success: 
    %     &&return&& @@None@@
        
    % # inpainting stage
    % &&if&& do_pasted:
    %     original_image = pasted
    % &&for&& _ in range(NUM_INPAINT_TRIAL):
    %     image = iterative_sdedit(original_image, prompt, mask=mask, **kwargs)
    %     success, _, _ = logo_detection(image, ref_embeds, **kwargs)
    %     &&if&& success:
    %         &&break&&

    % # determine as challenging image
    % &&if&& not success: 
    %     &&return&& @@None@@
        
    % # refinement stage
    % &&for&& i in range(num_refinement):
    %     # small noise inpainting
    %     image = iterative_sdedit(image, prompt, mask=mask, **kwargs)
    % &&return&& img

\textbf{Hyperparameters\pspace}\label{app:hyperparameters}
In our implementation, the refinement strength is controlled by specifying how many P\&P iterations $K_{t_i}$ are applied at each inference step $t_i$ within the motion stage, rather than using a single global value of $K_f$ and $\alpha$. Concretely, we define a P\&P plan as a mapping from inference step ranges to the number of P\&P iterations applied at each step. For example, a plan $\{\texttt{2--5}:2,\ \texttt{6--10}:1\}$ applies $K_f\!=\!2$ at steps 2--5 and $K_f\!=\!1$ at steps 6--10.

The specific plan is adjusted slightly depending on the task and model. For motion-enhanced video generation, we do not apply P\&P at the earliest steps in order to allow a coarse spatial layout (e.g., camera movement) to be determined.
Specifically, we use $\{\texttt{3--6}:3,\ \texttt{7--14}:1\}$, which results in an additional 20 NFEs in total. In later steps, we apply only a single P\&P iteration to lightly refine less critical regions while maintaining computational efficiency. All task- and model-specific hyperparameters are summarized in \cref{tab:hyperparams}.

\begin{table}[t]
    \centering
    \resizebox{1.0\linewidth}{!}{
    \renewcommand{\arraystretch}{1.}
    \renewcommand{\tabcolsep}{3pt}
    \begin{tabular}{llcl}
    \toprule
    Task & Setting & P\&P plan & $\tau$ \\
    \midrule
    Physical AI (I2V) & Wan2.2-I2V 
    & $\{\texttt{3--6}:3,\ \texttt{7--14}:1\}$ 
    & 0. \\

    Physical AI (I2V) & Cosmos2.5 
    & $\{\texttt{3--4}:5,\ \texttt{5--15}:1\}$ 
    & 0.5 \\

    Physics Video (T2V) & Wan2.2-T2V 
    & $\{\texttt{3--6}:3,\ \texttt{7--14}:1\}$ 
    & 0.25 \\
    
    Motion-enhanced & Wan2.2-T2V 
    & $\{\texttt{3--6}:3,\ \texttt{7--14}:1\}$ 
    & 0.25 \\
    Motion-enhanced & Wan2.1-T2V 
    & $\{\texttt{3--7}:3,\ \texttt{7--16}:1\}$ 
    & 0.50 \\

    Spatial & Wan2.2-T2V 
    & $\{\texttt{3--6}:3,\ \texttt{7--14}:1\}$ 
    & 0.25 \\
    \bottomrule
    \end{tabular}}
    \vspace{0.1mm}
    \caption{Task- and model-specific hyperparameters used for P\&P.}
    \vspace{-0.15in}
\label{tab:hyperparams}
\end{table}

% \subsection{Baselines}
% \paragraph{CFG-Zero}
% \paragraph{FlowMo}
\subsection{Motion Coherence for Challenging Motions} \label{app:details-motion}
\textbf{Baselines\pspace}
We reimplement FlowMo~\citep{shaulov2025flowmo} with gradient checkpointing, enabling it to run on a single GPU. For Wan2.1, we follow the official implementation. For Wan2.2, since FlowMo requires additional GPU memory due to gradient computation, we employ CPU offloading, and set the length of the FlowMo refinement steps to match that of the P\&P steps in our method.

We report VideoJam-bench results using Wan2.1 in Tab.~\ref{table:motion-videojam}. Our method achieves the strongest performance on VBench metrics. We note that these automated metrics are largely saturated, which may limit their sensitivity to fine-grained motion quality.

\begin{figure}[t!]
    \centering
    \includegraphics[width=1\linewidth]{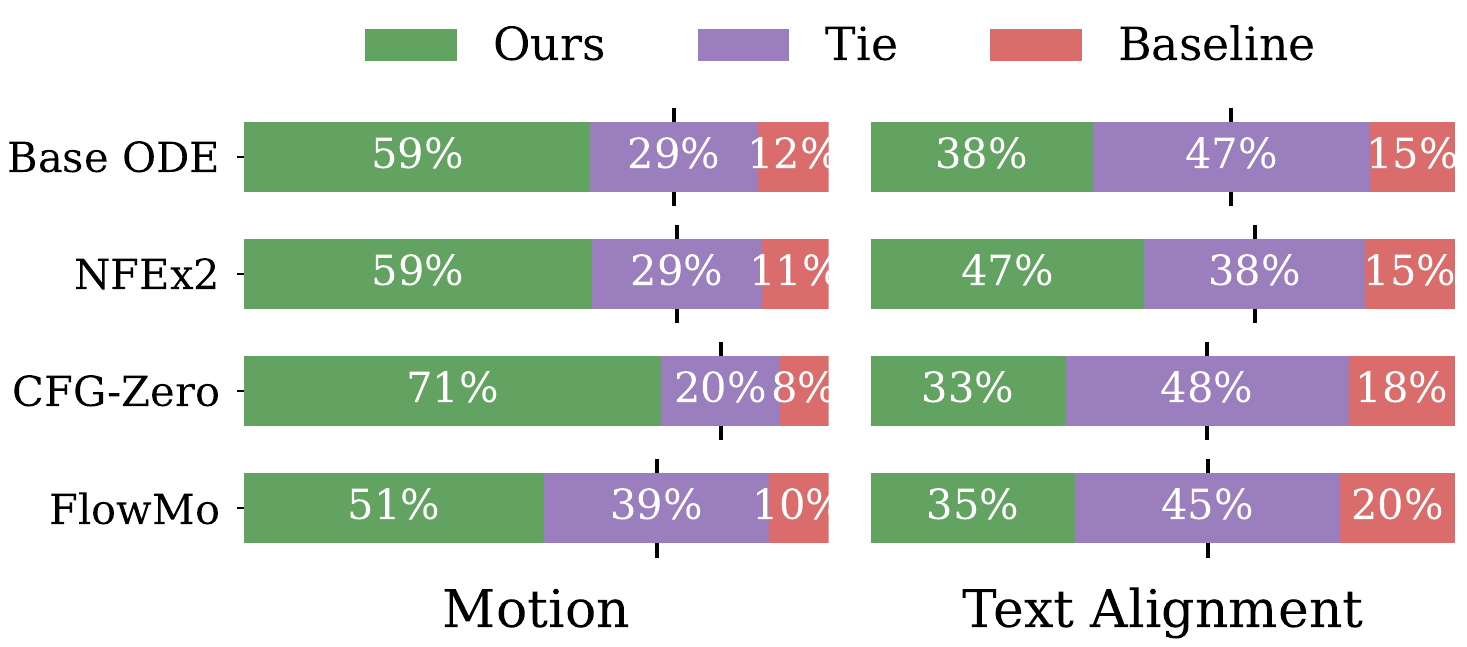}
    \vspace{-0.23in}
    \caption{Full human evaluation results on \textbf{Dynamic-Bench}, including ties.}\label{fig:fig_app_human_eval_motion}
    \vspace{-0.20in}
\end{figure}

\textbf{Human Evaluation\pspace}\label{app:motion_human_detail}
We provide an example of the human evaluation interface in \fref{fig:fig_app_human_eval} left. 
For each prompt, we display a pair of videos generated with the same random seed, one from our method and one from a baseline, and ask evaluators to assess motion quality and text alignment. 
Each evaluator views only a single video pair per prompt, with baseline methods randomly shuffled to avoid bias. 
The evaluation includes a \emph{tie} option. In Tab.~\ref{table:motion_wan22}, we report the tie-adjusted win rate (counting each tie as half a win), while the complete results including ties are shown in \fref{fig:fig_app_human_eval_motion}.

\begin{table}[t!]
    \centering
    \resizebox{1.0\linewidth}{!}{
    \renewcommand{\arraystretch}{1.}
    \renewcommand{\tabcolsep}{3pt}
    \begin{tabular}{lcccccc}
        \toprule
         & \multicolumn{4}{c}{VBench} &  \\
         \cmidrule(l{2pt}r{2pt}){2-5}
        Method & Motion$\uparrow$ & Dynamic$\uparrow$ & Const.$\uparrow$ & Quality$\uparrow$ & NFE & Time \\
        \midrule
        Wan2.1-14B & 98.10 & 77.34 & 94.22  & 61.92 & 50 & \\
         + NFE$\times 2$ & 98.01 & 77.34 & 94.32 & 61.95 & 100 &$2.0\times$\\
         + FlowMo & 97.49 & \textbf{79.17}  & 93.40  & 60.89 & \phantom{*}{50}\textcolor{blue}{*} &$
         3.3\times$\\
         + CFG-Zero & 98.00 & 78.13  & 94.20  & 61.63 & 50 &$1.0\times$\\
         % \rowcolor{RoyalBlue!05}
         \cellcolor{RoyalBlue!05}+ Ours &\cellcolor{RoyalBlue!05}\textbf{98.37} &\cellcolor{RoyalBlue!05}77.34  &\cellcolor{RoyalBlue!05}\textbf{94.77}  &\cellcolor{RoyalBlue!05}\textbf{63.08} &\cellcolor{RoyalBlue!05}74
         &\cellcolor{RoyalBlue!05}$1.5\times$\\
         \midrule
         Wan2.1-1.3B & 98.21 & 75.00 & 94.05  & 61.10 & 50 & \\
         + NFE$\times 2$ & 98.23 & 77.34 & 94.23  & \textbf{61.60} & 100 
         & $2.0\times$\\
         + FlowMo & 97.89 & 75.00  & 93.77  & 59.88 & \phantom{*}{50}\textcolor{blue}{*} & $3.3\times$\\
         + CFG-Zero & 98.01 & \textbf{85.16}  & 93.71  & 60.71 & 50 & $1.0\times$\\
         \cellcolor{RoyalBlue!05}+ Ours &\cellcolor{RoyalBlue!05}\textbf{98.84} &\cellcolor{RoyalBlue!05}73.31  &\cellcolor{RoyalBlue!05}\textbf{94.95}  &\cellcolor{RoyalBlue!05}61.43 
         &\cellcolor{RoyalBlue!05}74 
         &\cellcolor{RoyalBlue!05}$1.5\times$\\
        \bottomrule
    \end{tabular}}
    \vspace{+0.7mm}
        \caption{\textbf{VideoJAM-bench} results measuring motion coherence. Additional inference time (\textcolor{blue}{*}) of FlowMo is introduced by gradient computation.}\label{table:motion-videojam}
        \vspace{-0.25in}
\end{table}

\subsection{Physical Realism in Robotics Videos} \label{app:details-robotics}
\textbf{Benchmark\pspace} 
We use all 174 robot-domain datasets from PAI-Bench-G~\citep{zhou2025paibench} as image-prompt pairs. For Robot-QA, we use Qwen2.5-VL-72B-Instruct~\citep{bai2025qwen25vl}, which provides sufficiently strong performance on robot-domain evaluation.
We provide the grasp success rate evaluated using Gemini 3 Flash~\citep{gemini3} in \fref{app_grasp_question}. To accurately assess grasp motion, videos are evaluated at high input resolution with a frame rate of 4~fps, and samples with scores of 4 or 5 are treated as successful grasps.

%%%%%%%%%%%%%%%%%%%%%%%%%%%%%%%%%%%%%%%%%%%%%%%%%%%%%%%%%%%%%%%%%%%
%%%%%%%%%%%%%%%%%%%%%%%%%%%%%%%%%%%%%%%%%%%%%%%%%%%%%%%%%%%%%%%%%%%
%%%%%%%%%%%%%%%%%%%%%%%%%%%%%%%%%%%%%%%%%%%%%%%%%%%%%%%%%%%%%%%%%%%
\begin{figure}[t!]
    \centering
    \includegraphics[width=1\linewidth]{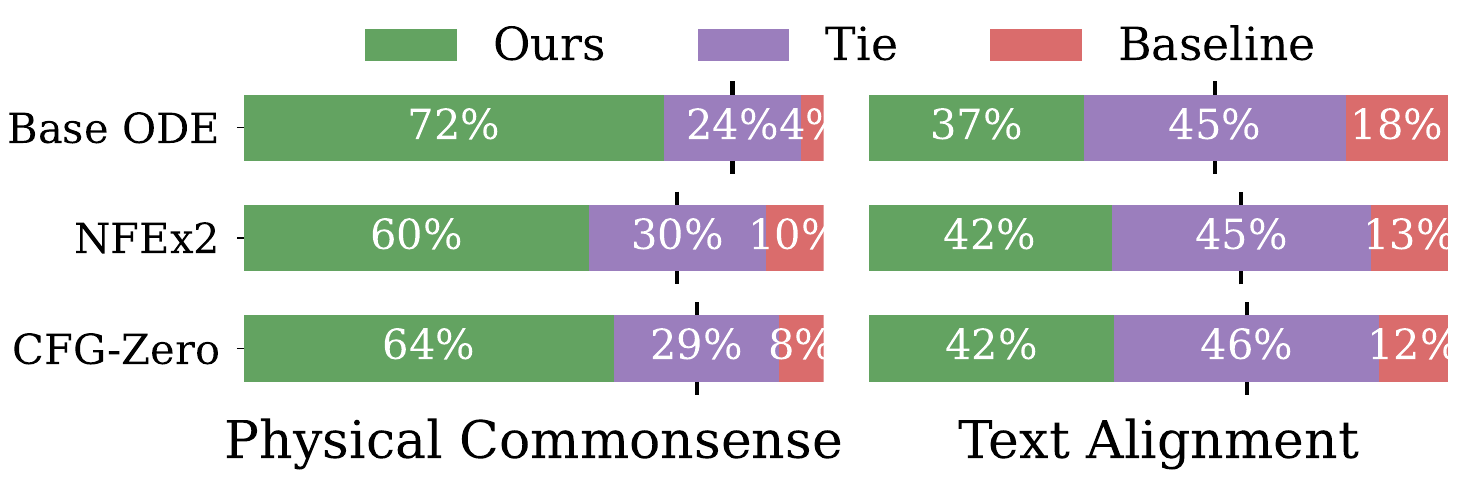}
    \vspace{-0.20in}
    \caption{Full human evaluation results on \textbf{VideoPhy2~\citep{bansal2025videophy2}} hard subset, including ties.}\label{fig:fig_app_human_eval_physics}
    \vspace{-0.25in}
\end{figure}
%%%%%%%%%%%%%%%%%%%%%%%%%%%%%%%%%%%%%%%%%%%%%%%%%%%%%%%%%%%%%%%%%%%

\begin{figure*}[t!]
    \centering
    \begin{tcolorbox}[
        colback=blue!5,
        colframe=black,
        arc=6pt,
        boxrule=1pt,
        width=\linewidth
    ]
    \textbf{Task Description}: Evaluate whether the robot in the video successfully performs a grasp action with proper physical contact with the target object.
    
    \vspace{4pt}
    
    \textbf{Evaluation Criteria:}
    
    1. Contact Detection: Does the robot's gripper/end-effector make actual physical contact with the object?
    
    2. Grasp Validity: Is the grasp physically plausible? (No floating objects, no penetration artifacts)
    
    3. Object Manipulation: After grasping, is the object properly held/moved by the robot?
    
    4. Visual Artifacts: Are there any visual artifacts such as objects floating without contact, gripper passing through objects, or impossible physical interactions?

    \vspace{4pt}

    \textbf{Instructions for Scoring:}
    
    - 1 (Fail): No grasp attempt, or severe artifacts (object floats, no contact, gripper passes through object)
    
    - 2 (Poor): Grasp attempted but clear physical violations (partial penetration, unrealistic contact)
    
    - 3 (Moderate): Grasp occurs but with minor artifacts or questionable contact quality
    
    - 4 (Good): Clear grasp with proper contact, minor visual imperfections acceptable
    
    - 5 (Excellent): Perfect grasp with realistic physical contact and natural object manipulation.

    \vspace{4pt}
    
    \textbf{Special Cases:}
    
    - If there is NO grasp action in the video (robot doesn't attempt to grasp anything), score as 1 (Fail)
    
    - If the object appears to stick to the gripper without proper contact, score as 1-2
    
    \vspace{4pt}
    
    \textbf{Please provide the output in the format:}
    
    Score: [1-5]
    
    Grasp Detected: [Yes/No]
        
    Explanation: [Your detailed reasoning about the grasp quality and any artifacts observed]
    \end{tcolorbox}
    \vspace{-0.1in}
    \caption{Gemini prompt for evaluating grasp success rate in Tab.~\ref{table:cosmos}. We treat scores of 4 or 5 as successful grasps.}
    \label{app_grasp_question}
    \vspace{-0.1in}
\end{figure*}

\subsection{Physics Alignment in the Wild} \label{app:details-physics}
\textbf{Benchmark\pspace}
We follow the original evaluation prompts of VideoPhy2~\citep{bansal2025videophy2} and PhyWorldBench~\citep{gu2025phyworldbench}, but perform all automatic evaluations using Gemini~3 Flash, which supports higher input frame rates.  
For PhyWorldBench, we evaluate only the two categories most closely related to motion, \emph{Object Motion and Kinematics} and \emph{Interaction Dynamics}.  
For PisaBench~\citep{li2025pisa}, since the evaluation requires square inputs, all videos are generated in a resolution of $512\times512$.

\textbf{Baselines\pspace}
Regarding rejection sampling (best-of-4), we follow the official documentation~\citep{cosmos-reason1-rejection-sampling}. Specifically, we use Cosmos-Reason1 7B~\citep{azzolini2025cosmosreason1} to repeatedly query whether a generated video contains anomalies or artifacts, and compute the score by averaging the number of responses indicating the absence of anomalies.

% We evaluate our method using two models: Cosmos-Predict2.5-2B~\citep{cosmos-predict2.5} post-trained and Wan2.2-A14B I2V. Additionally, we include rejection sampling with Cosmos-Reason1 7B~\citep{azzolini2025cosmosreason1} video critic following the official documentation~\citep{cosmos-reason1-rejection-sampling}. For each video, four samples are generated using different seeds and scored based on anomalies, where the highest-scoring sample is selected (best-of-4). 

\begin{figure*}[t!]
    \centering
    \includegraphics[width=1\linewidth]{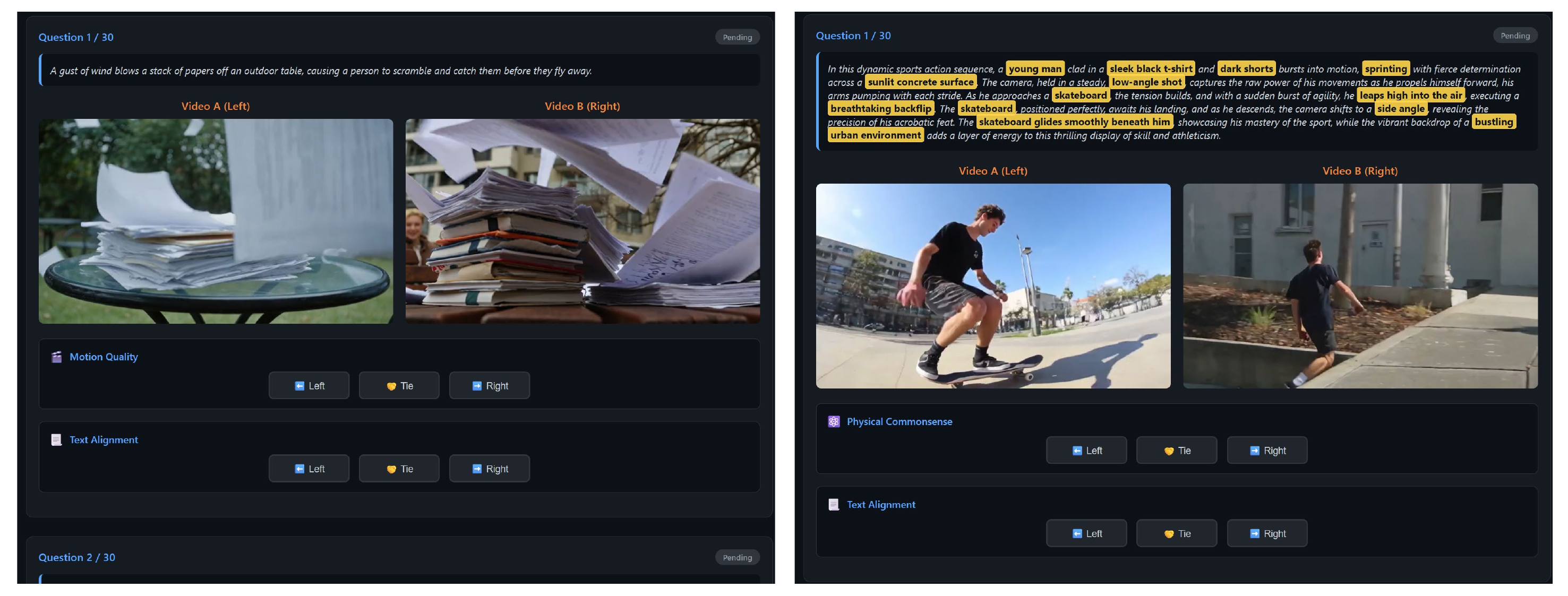}
    \vspace{-0.15in}
    \caption{A screenshot of the human evaluation questionnaires used for (left) motion-enhanced video generation on Dynamic-Bench and (right) physics-aligned video generation on the VideoPhy2 hard subset.}\label{fig:fig_app_human_eval}
    \vspace{-0.15in}
\end{figure*}

\textbf{Human Evaluation\pspace}
We provide an example of the human evaluation interface in \fref{fig:fig_app_human_eval} right. 
For each prompt, we display a pair of videos generated with the same random seed, one from our method and one from a baseline, and ask evaluators to assess physical commonsense (PC) and text alignment (semantic alignment; SA). 
Since the video generation prompts in the benchmark are relatively long, we highlight key phrases using colored blocks.
All other evaluation details follow those used for motion-enhanced video generation.
We provide the complete human evaluation results including ties are provided in \fref{fig:fig_app_human_eval_physics}.

\subsection{Improvement in Spatial Consistency} \label{app:details-spatial}
\textbf{Benchmark\pspace}
% \jang{How to filter, measure}
The prompts used in this evaluation all include a \emph{first-person view}. Additionally, since models frequently produce misaligned videos under such settings, including limited camera rotation or unstable camera trajectories, we generate multiple videos per prompt and filter them based on camera viewpoints. Specifically, we retain videos with sufficiently large yaw coverage and stable camera trajectories, while discarding cases dominated by in-place rotation or exhibiting unreliable viewpoint estimates. In total, we conduct our evaluation on 20 videos.

\subsection{Ablation Studies on Hyperparameters}\label{app:ablation}
\begin{figure*}[p]
    \centering
        \includegraphics[width=0.84\linewidth]{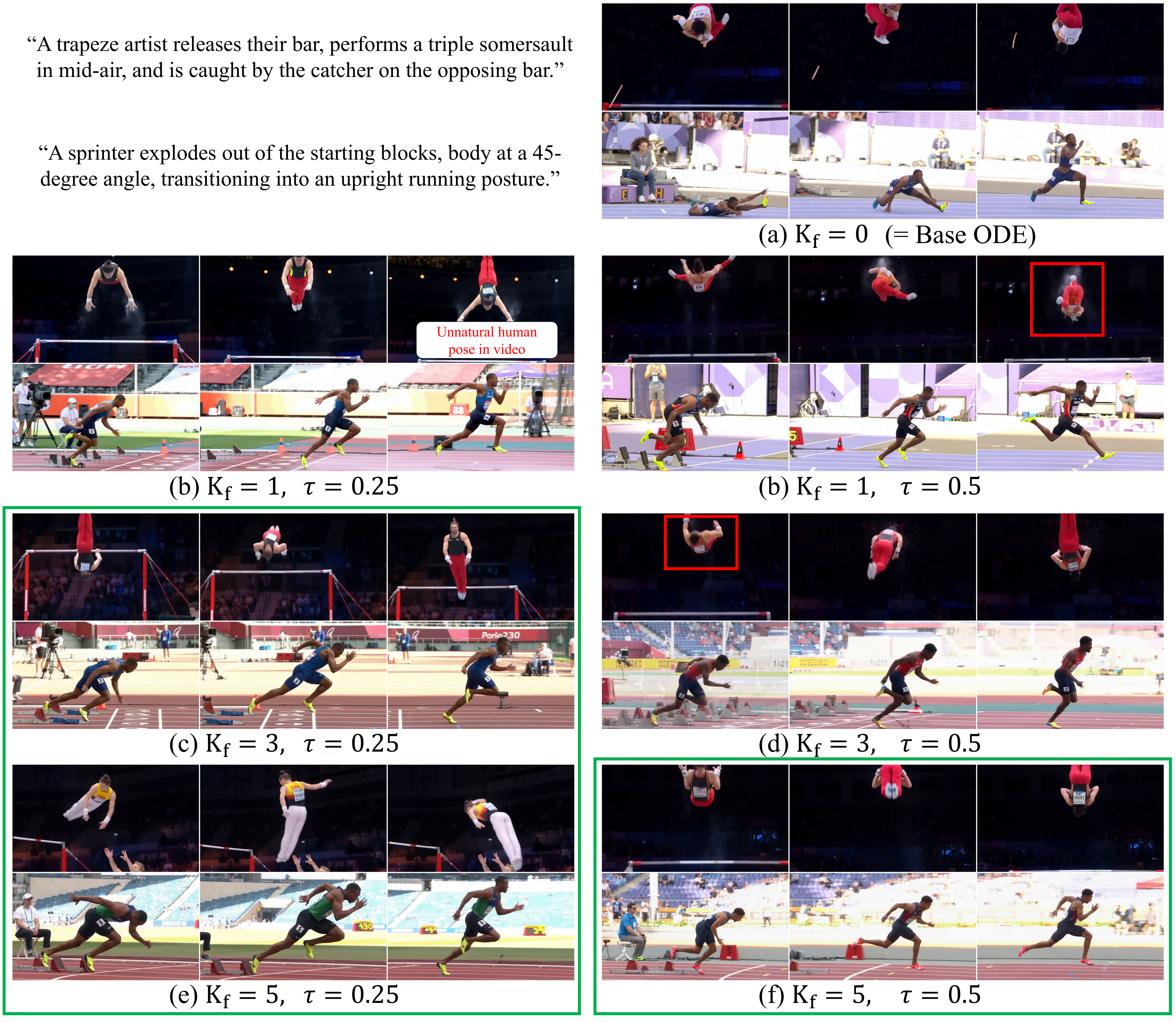}
    \vspace{5mm}
    \vspace{-0.3in}
    \caption{Ablation studies on the hyperparameters $K_f$ and $\tau$.}\label{fig:fig_app_hyper_ablation}
    \vspace{2mm}
        \includegraphics[width=0.84\linewidth]{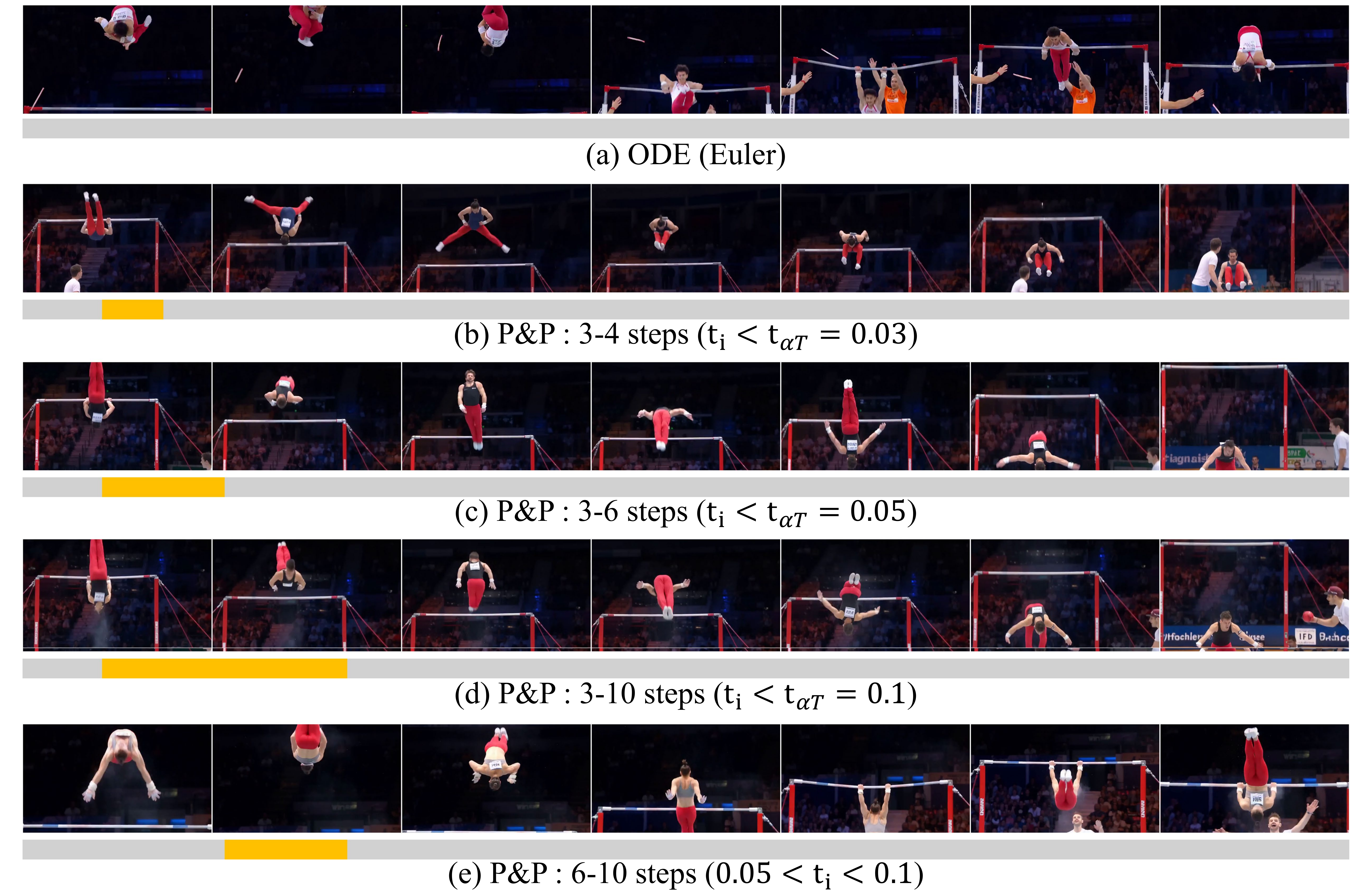}
    \vspace{-0.09in}
    \caption{Ablation studies on the hyperparameter $\alpha$. Gray blocks indicate Euler method and orange blocks indicate P\&P. P\&P significantly improves motion coherence when applied in earlier steps (b-c), while providing only marginal gains at later steps (d-e).}\label{fig:fig_app_hyper_ablation2}
    \vspace{-0.05in}
\end{figure*}

We provide ablation studies on the key hyperparameters of our method, the number of P\&P iterations $K_f$ and the confidence threshold $\tau$, in \fref{fig:fig_app_hyper_ablation}. Increasing $K_f$ strengthens the refinement effect of P\&P, but also incurs additional NFEs and results in larger deviations from the base ODE samples. In contrast, a small value such as $K_f\!=\!1$ is insufficient to adequately refine large motion in the generated videos.

The confidence threshold $\tau$ primarily controls how well background appearance and overall color tone from the base ODE sampling are preserved. As $\tau$ increases, refinement becomes more conservative, slightly reducing refinement strength while better preserving the original background structure and color tone. As discussed in Sec.~\ref{sec:uncertainty}, when $K_f$ becomes large, saturation artifacts may still appear even with uncertainty-aware strategy, such as an overall brightening of the video. In such cases, jointly increasing $\tau$ effectively mitigates these artifacts by restricting refinement to more uncertain regions. Based on these observations, we use $K_f\!=\!3$ and $\tau\!=\!0.25$ by default, which provides a favorable trade-off between sample quality and computational cost.

As shown in \fref{fig:fig_app_hyper_ablation2}, we conduct an ablation study on $\alpha$, which determines the temporal extent of the motion stage where P\&P refinement is applied. When $\alpha$ exceeds a certain threshold, all configurations in \fref{fig:fig_app_hyper_ablation2}(b–d) produce stable and coherent motion. However, comparing (c) and (d) reveals that later inference steps contribute less to motion dynamics. From an efficiency perspective, reducing $K_f$ or disabling P\&P at later steps is more effective. Similarly, applying P\&P only at late stages, as in \fref{fig:fig_app_hyper_ablation2}(e), reduces visual artifacts since the motion has already been largely determined, but fails to fully correct motion errors due to strong cross-frame consistency.

% \section{Additional Generated Examples}\label{app:additional_examples}
% Compare with commercial models, Veo3, Sora?

%%%%%%%%%%%%%%%%%%%%%%%%%%%%%%%%%%%%%%%%%%%%%%%%%%%%%%%%%%%%%%%%%%%%%%%
\begin{figure}[t!]
    \centering
    \includegraphics[width=1\linewidth]{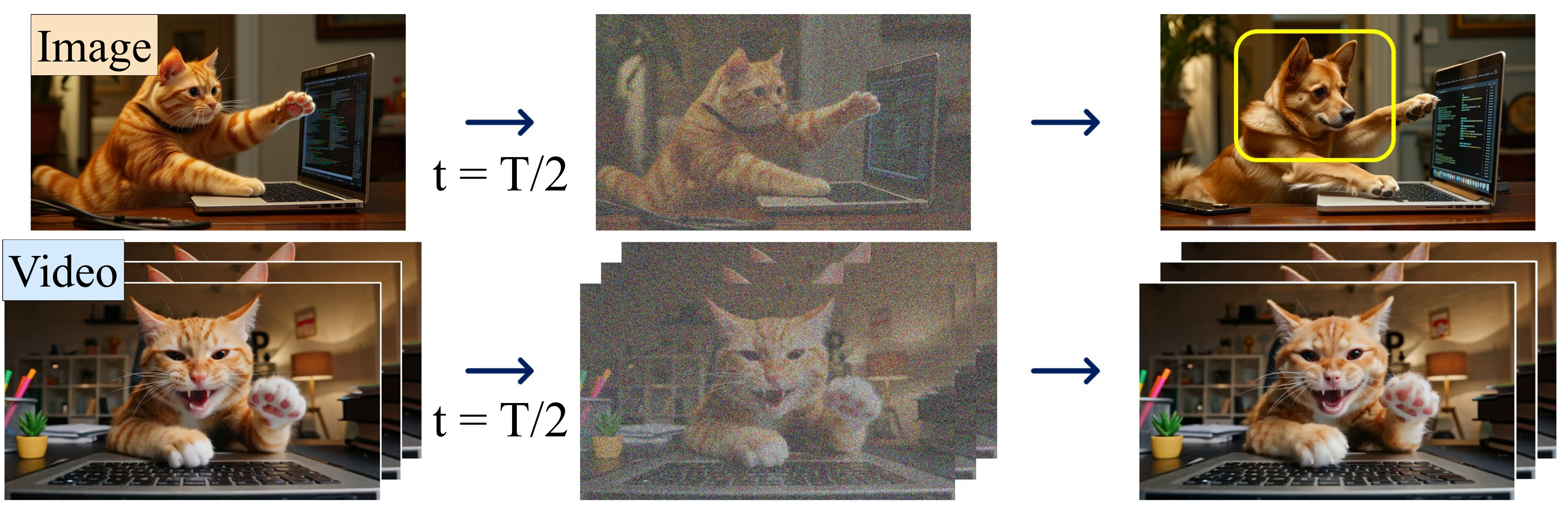}
    \vspace{-0.25in}
    \caption*{(a) Comparison of SDEdit results on an image and a video while changing the prompt from \texttt{orange\;cat} to \texttt{brown\;dog}.}
    \vspace{0.05in}
    \includegraphics[width=1\linewidth]{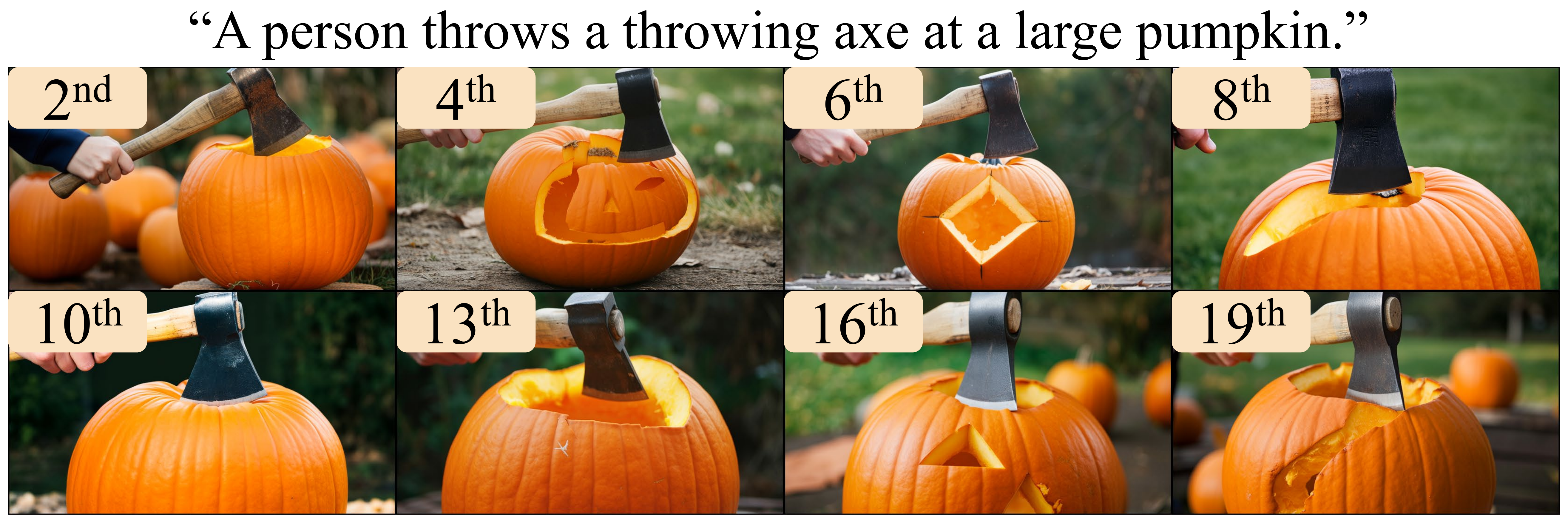}
    \vspace{-0.25in}
    \caption*{\textbf{Image} generation with P\&P}
    \includegraphics[width=1\linewidth]{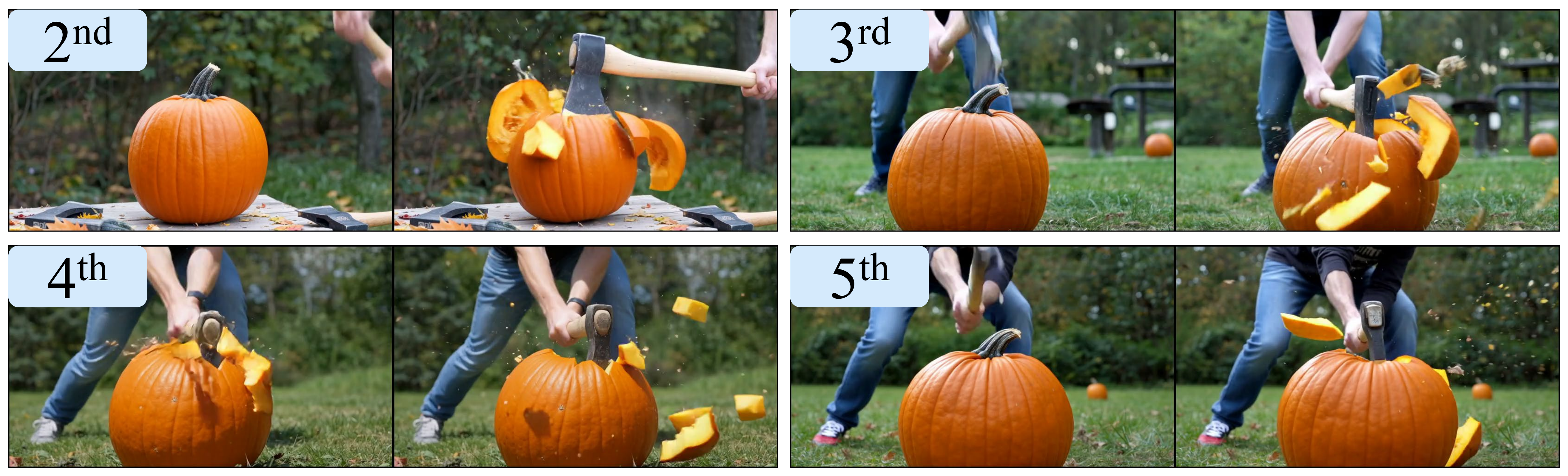}
    \vspace{-0.25in}
    \caption*{\textbf{Video} generation with P\&P}
    \vspace{-0.05in}
    \caption*{(b) Comparison of applying P\&P on an image and a video. 
    % 3 P\&P updates are applied at a single inference step, as indicated by the top-left number of each sample.}
    We apply 3 P\&P iterations at the inference step indicated by the top-left number of each sample.}
    % \vspace{0.05in}
    % \begin{minipage}[t]{0.49\linewidth}
    %     \includegraphics[width=1\linewidth]{figure/image_early.pdf}
    % \end{minipage}
    % \hfill
    % \begin{minipage}[t]{0.49\linewidth}
    %     \includegraphics[width=1\linewidth]{figure/video_early.pdf}
    % \end{minipage}
    % \vspace{-0.10in}
    % \caption*{(c) L2 distance to the final refined endpoint during P\&P iterations at a fixed inference step (second step, $t = 0.009T$). Results are obtained using the Wan2.2-A14B T2V~\citep{wan2025}.}
    \vspace{-0.05in}
    % \caption{(a) Visualization of cross-frame consistency in videos. (b) Convergence of samples during P\&P iterations.}
    \caption{Cross-frame consistency in videos. Due to strong temporal correlations across frames, video is more robust to perturbation than image.}
    \label{fig:frame_consistency}
    \vspace{-0.15in}
\end{figure}
%%%%%%%%%%%%%%%%%%%%%%%%%%%%%%%%%%%%%%%%%%%%%%%%%%%%%%%%%%%%%%%%%%%%%%

% \vspace{0.05in}
% \begin{minipage}[t]{0.49\linewidth}
%     \includegraphics[width=1\linewidth]{figure/image_early.pdf}
% \end{minipage}
% \hfill
% \begin{minipage}[t]{0.49\linewidth}
%     \includegraphics[width=1\linewidth]{figure/app_cross_frame_consistency_image.pdf}
% \end{minipage}
% \vspace{-0.10in}
% \caption*{(b) L2 distance to the final refined endpoint during P\&P iterations at a fixed inference step (second step, $t = 0.009T$). Results are obtained using the Wan2.2-A14B T2V~\citep{wan2025}.}

\section{More Discussions} \label{app:more_discussion}
\subsection{Cross-Frame Consistency of Video} \label{app:cross-frame}
As discussed in Sec.~\ref{sec:cross-frame}, strong cross-frame consistency makes videos robust to perturbations. As shown in \fref{fig:frame_consistency}, videos are less responsive to both SDEdit and multiple P\&P iterations than images, making late-stage one-shot perturbations less effective. This indicates that effective video refinement, especially for motion, requires larger early-stage perturbations and iterative refinement.

\begin{figure}[t!]
    \centering
    \begin{minipage}[t]{0.49\linewidth}
        \includegraphics[width=1\linewidth]{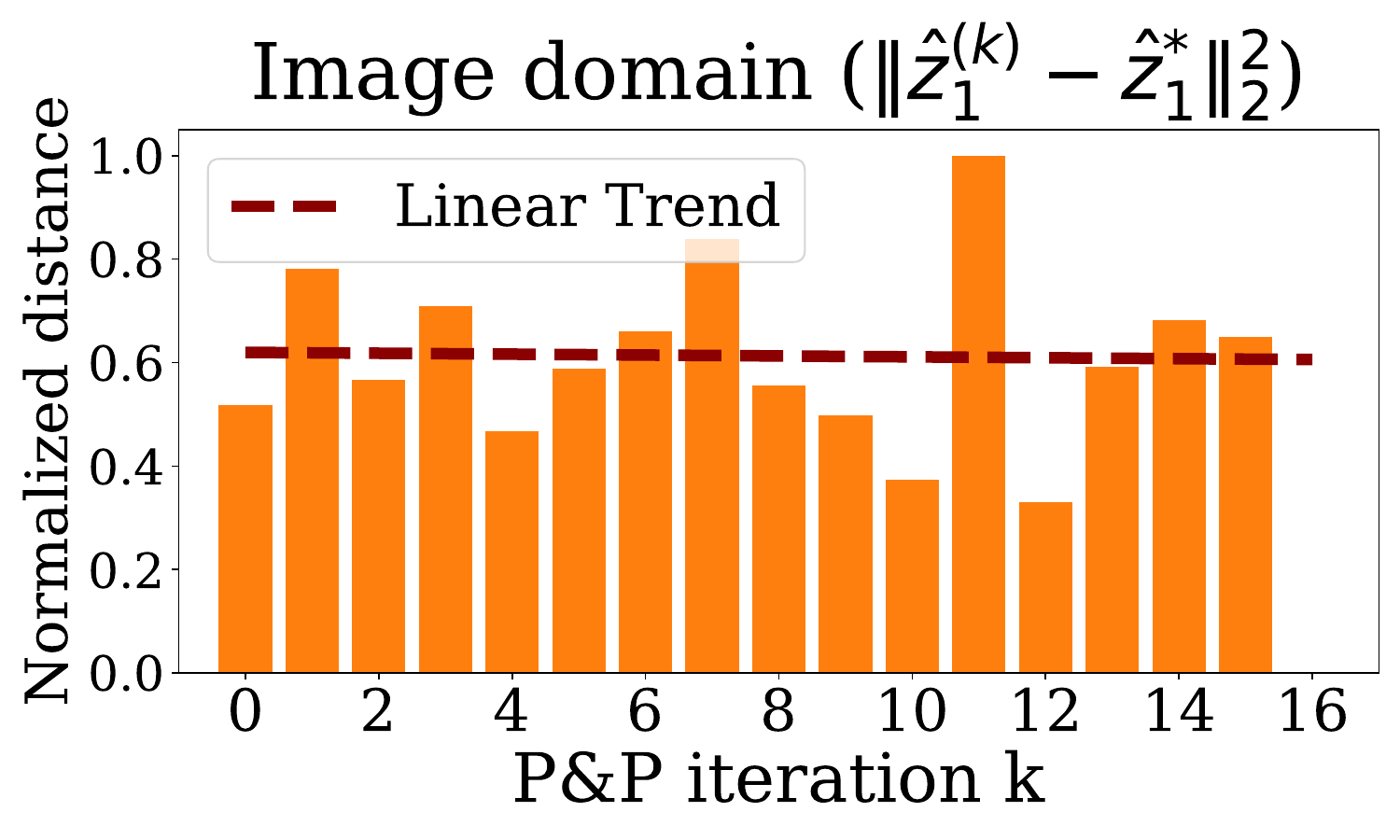}
    \end{minipage}
    \hfill
    \begin{minipage}[t]{0.49\linewidth}
        \includegraphics[width=1\linewidth]{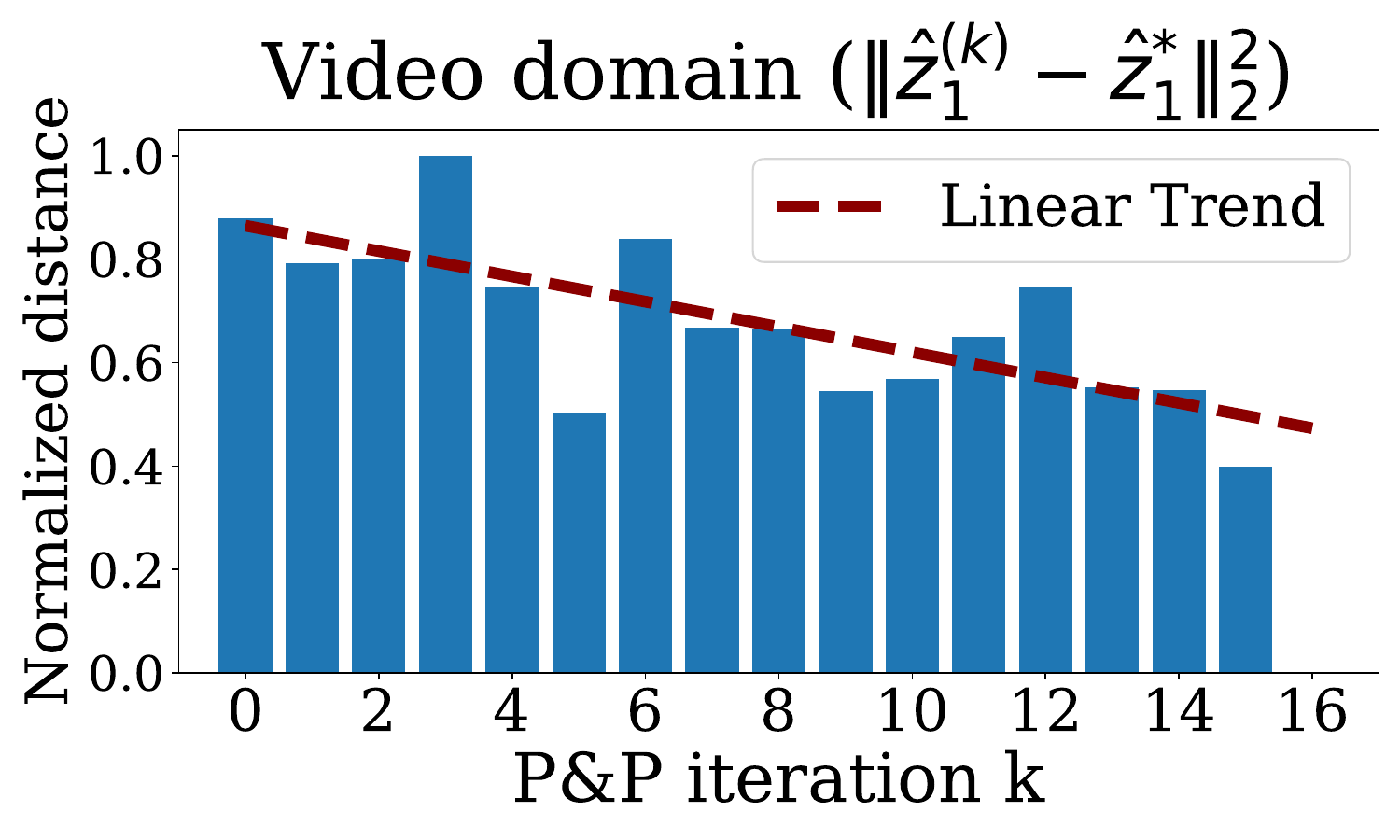}
    \end{minipage}
    \vspace{-0.10in}
    \caption{Accumulated effect of iterative P\&P at an early inference step. 
    We plot the L2 distance between the intermediate refined latent $\hat{z}_1^{(k)}$ and the final refined latent $\hat{z}_{1}^{*}$ at a fixed inference step $t=0.009T$. Results are obtained using Wan2.2-A14B T2V.}
    % We visualize L2 distance to the final refined endpoint across P\&P iterations at a fixed inference step (the second step, $t = 0.009T$). Results are obtained using Wan2.2-A14B T2V.}
    \label{fig:frame_consistency_decrease}
    \vspace{-0.15in}
\end{figure}

Despite this difficulty, such robustness allows refinement effects to accumulate stably across iterations. When we measure the L2 distance between the final refined prediction $\hat{z}_1^*$ and intermediate predictions $\hat{z}_1^{(k)}$ during P\&P iterations at an early timestep, we observe a near-linear decrease for videos, as shown in \fref{fig:frame_consistency_decrease}. In contrast, images exhibit oscillatory behavior, indicating inconsistent refinement directions. This behavior indicates that refinement effects accumulate consistently across iterations for videos, motivating an early-stage, iterative refinement strategy.

\subsection{Other domains} \label{app:other-domain}

\begin{figure}[t!]
    \centering
    \includegraphics[width=1\linewidth]{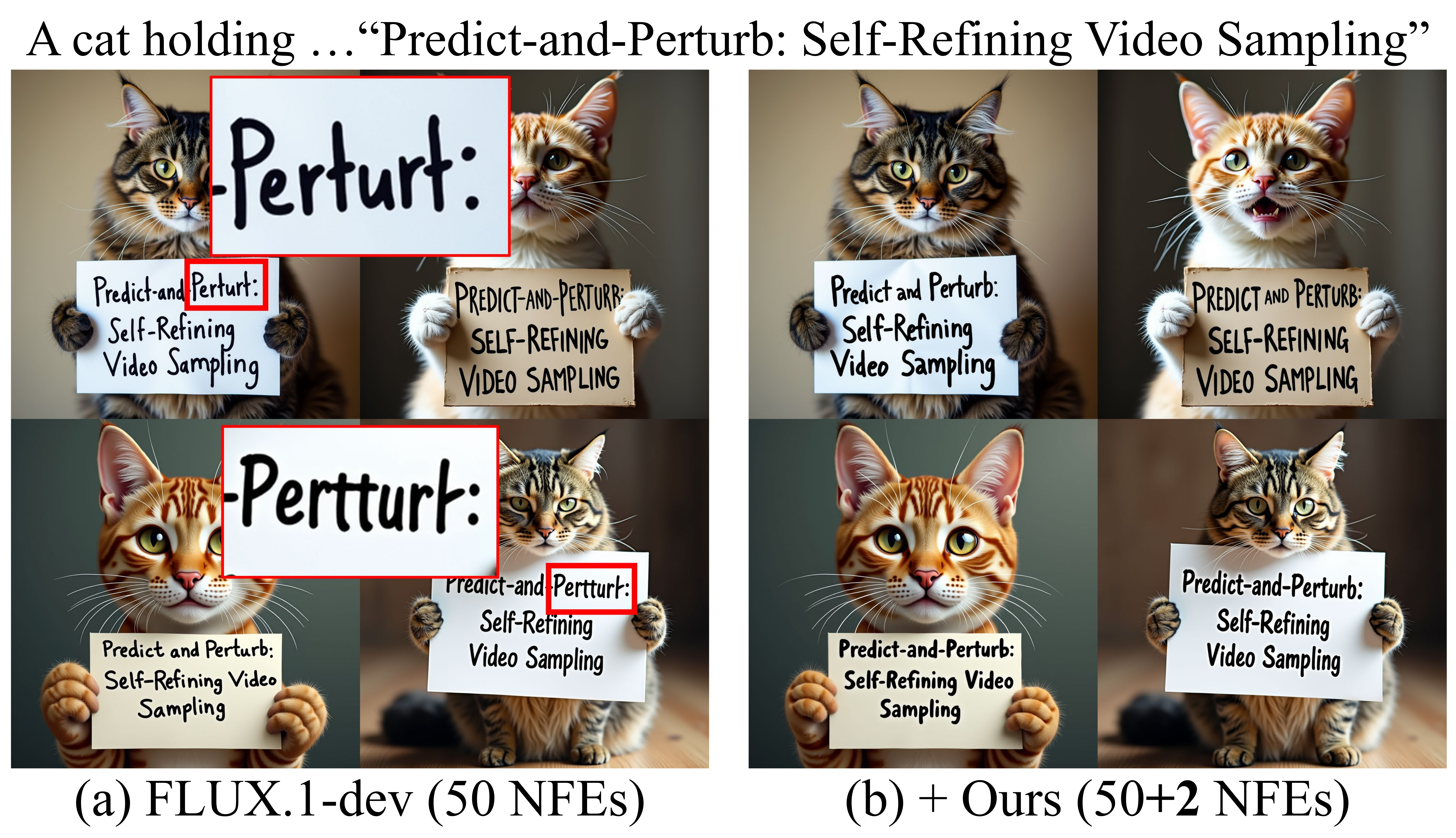}
    \vspace{-0.20in}
    \caption{
    Image generation with P\&P using FLUX.1-dev. With only two additional NFEs (\textbf{4\%}), our method effectively reduces text-related artifacts, resulting in clearer and more coherent text.
    }\label{fig:fig_app_flux}
    \vspace{-0.15in}
\end{figure}

Proposed P\&P is applicable to general flow matching generators. We further examine the effectiveness of this framework on the image generation using FLUX-1.dev~\citep{flux}. As shown in Fig.~\ref{fig:fig_app_flux}, P\&P can reduce text-related artifacts, leading to clearer and more coherent text rendering compared to the base ODE sampling.
We generate four samples with different random seeds using the prompt ``A cat holding a sign that says `Predict-and-Perturb: Self-Refining Video Sampling'.''

For image generation, P\&P is applied only twice at the 10th inference step out of 50 total steps in FLUX, resulting in only \textbf{a 4\% increase in NFEs}. Unlike the video domain, where cross-frame consistency makes refinement more robust but often requires multiple iterations, image generation typically benefits from only a few P\&P iterations at a fixed noise level to achieve noticeable improvements. Notably, the uncertainty estimation in the image domain is primarily concentrated on challenging regions such as text rendering, leading P\&P to selectively refine the text while leaving the rest of the image unchanged.

%%%%%%%%%%%%%%%%%%%%%%%%%%%%%%%%%%%%%%%%%%%%%%%%%%%%%%%%%%%%%%%%%%%%%%%
%%%%%%%%%%%%%%%%%%%%%%%%%%%%%%%%%%%%%%%%%%%%%%%%%%%%%%%%%%%%%%%%%%%%%%
%%%%%%%%%%%%%%%%%%%%%%%%%%%%%%%%%%%%%%%%%%%%%%%%%%%%%%%%%%%%%%%%%%%
%%%%%%%%%%%%%%%%%%%%%%%%%%%%%%%%%%%%%%%%%%%%%%%%%%%%%%%%%%%%%%%%%%%

%%%%%%%%%%%%%%%%%%%%%%%%%%%%%%%%%%%%%%%%%%%%%%%%%%%%%%%%%%%%%%%%%%%
%%%%%%%%%%%%%%%%%%%%%%%%%%%%%%%%%%%%%%%%%%%%%%%%%%%%%%%%%%%%%%%%%%%

%%%%%%%%%%%%%%%%%%%%%%%%%%%%%%%%%%%%%%%%%%%%%%%%%%%%%%%%%%%%%%%%%%%
% \begin{figure*}[t!]
%     \centering
%     \includegraphics[width=0.95\linewidth]{figure/figure_spatial_mine.pdf}
%     \vspace{-0.06in}
%     \caption{TBD.}\label{fig:fig_app_spatial_mine}
%     \vspace{-0.05in}
% \end{figure*}
%%%%%%%%%%%%%%%%%%%%%%%%%%%%%%%%%%%%%%%%%%%%%%%%%%%%%%%%%%%%%%%%%%%
%%%%%%%%%%%%%%%%%%%%%%%%%%%%%%%%%%%%%%%%%%%%%%%%%%%%%%%%%%%%%%%%%%%
\begin{figure*}[p]
    \centering
    \includegraphics[width=0.9\linewidth]{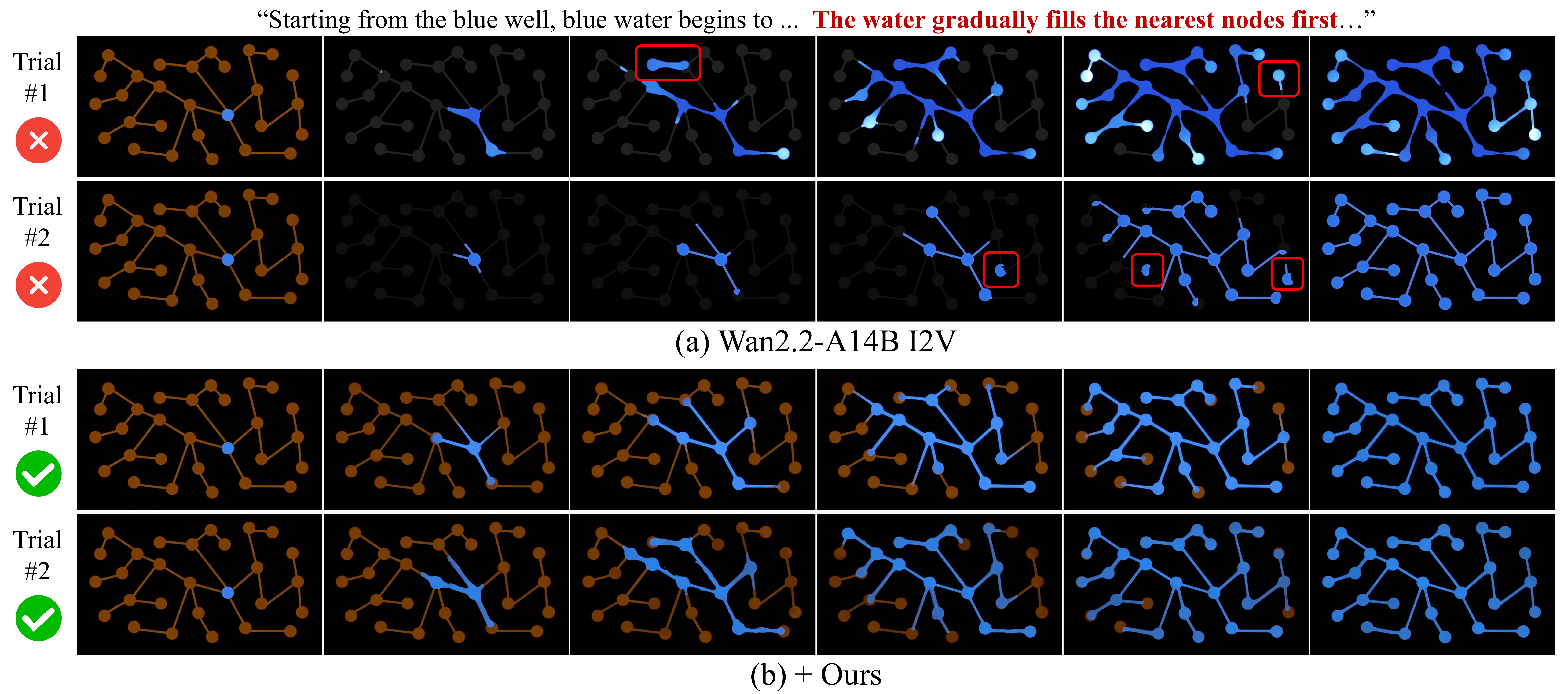}
    \vspace{-0.07in}
    \caption{Graph traversal task in \citet{wiedemer2025videomodelszeroshotlearners}. We use Wan2.2-A14B I2V with an upsampled prompt: ``Starting from the blue well, blue water begins to flow slowly through the connected channel system. The water gradually fills the nearest nodes first...''. The success rate increases from 0.1 to 0.8 with P\&P method.}
    \label{fig:app_reasoning_graph}
% \end{figure*}
%%%%%%%%%%%%%%%%%%%%%%%%%%%%%%%%%%%%%%%%%%%%%%%%%%%%%%%%%%%%%%%%%%%
%%%%%%%%%%%%%%%%%%%%%%%%%%%%%%%%%%%%%%%%%%%%%%%%%%%%%%%%%%%%%%%%%%%
% \begin{figure*}[t!]
    \centering
    \includegraphics[width=0.9\linewidth]{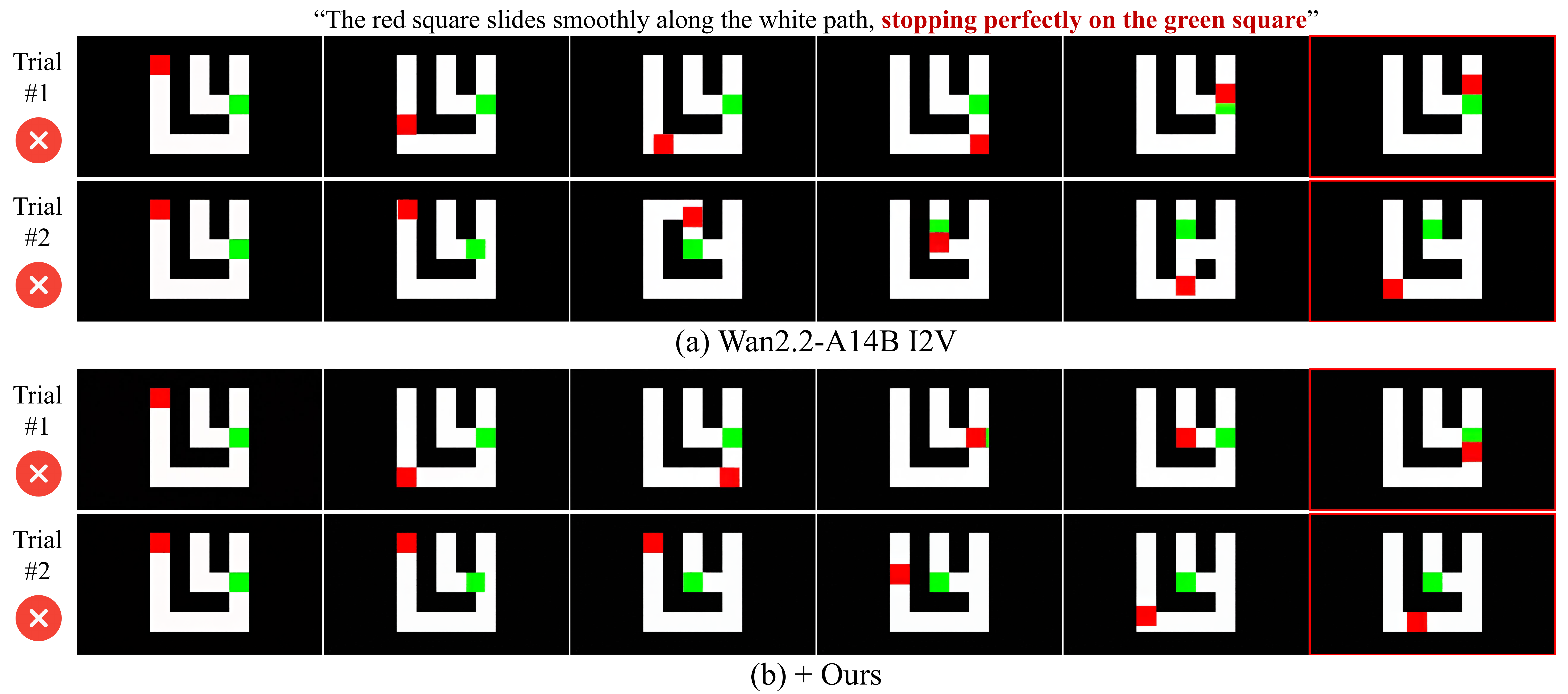}
    \vspace{-0.07in}
    \caption{Maze solving task in \citet{wiedemer2025videomodelszeroshotlearners}. We use Wan2.2-A14B I2V with a base prompt: ``The red square slides smoothly along the white path, stopping perfectly on the green square.'' Both the base model and P\&P method achieve near-zero success rates.}
    \label{fig:app_reasoning_maze}
    \vspace{0.08in}
% \end{figure*}
%%%%%%%%%%%%%%%%%%%%%%%%%%%%%%%%%%%%%%%%%%%%%%%%%%%%%%%%%%%%%%%%%%%
%%%%%%%%%%%%%%%%%%%%%%%%%%%%%%%%%%%%%%%%%%%%%%%%%%%%%%%%%%%%%%%%%%%
% \begin{figure*}[t!]
    \centering
    \includegraphics[width=1\linewidth]{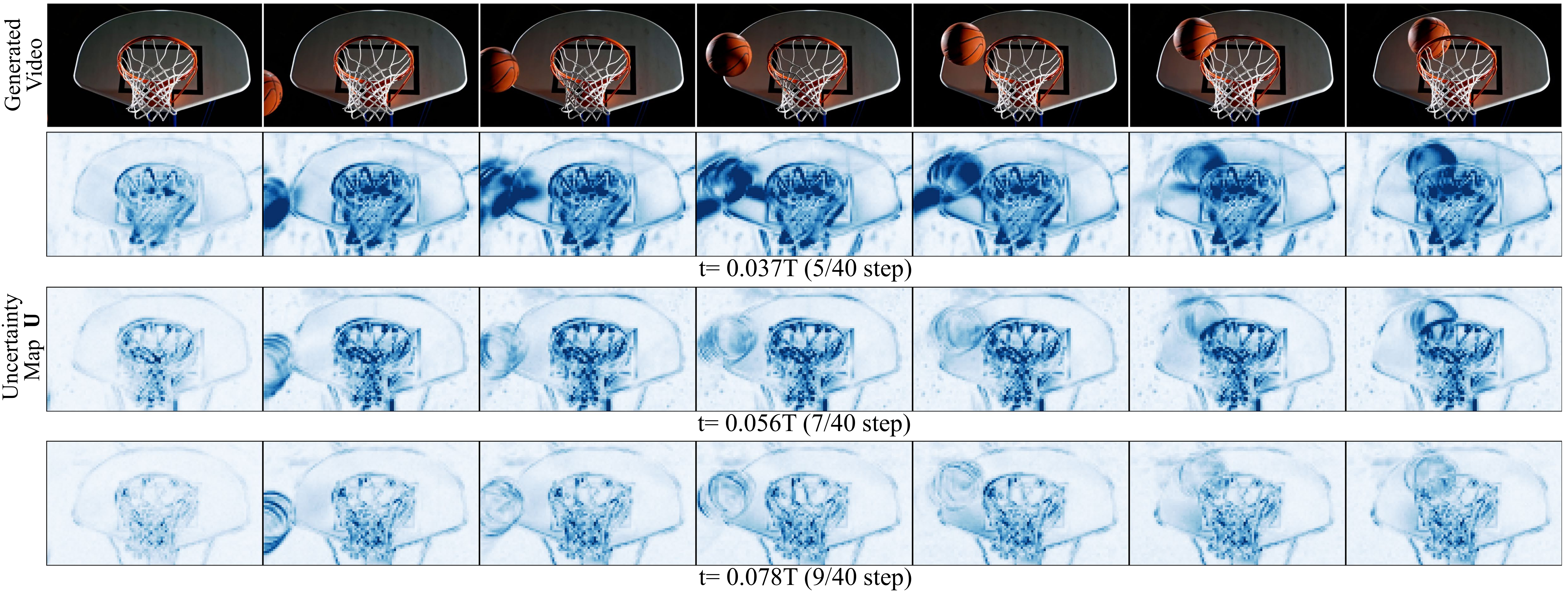}
    \vspace{-0.25in}
    \caption{Visualization of uncertainty maps across inference timesteps. Overall uncertainty gradually decreases as inference progresses. Even at an early timestep ($t=0.0037T$), higher uncertainty values are observed for objects exhibiting motion.}
    \label{fig:app_uncertainty_time}
\end{figure*}
%%%%%%%%%%%%%%%%%%%%%%%%%%%%%%%%%%%%%%%%%%%%%%%%%%%%%%%%%%%%%%%%%%%

%%%%%%%%%%%%%%%%%%%%%%%%%%%%%%%%%%%%%%%%%%%%%%%%%%%%%%%%%%%%%%%%%%%
\begin{figure*}[t!]
    \centering
    \includegraphics[width=0.8\linewidth]{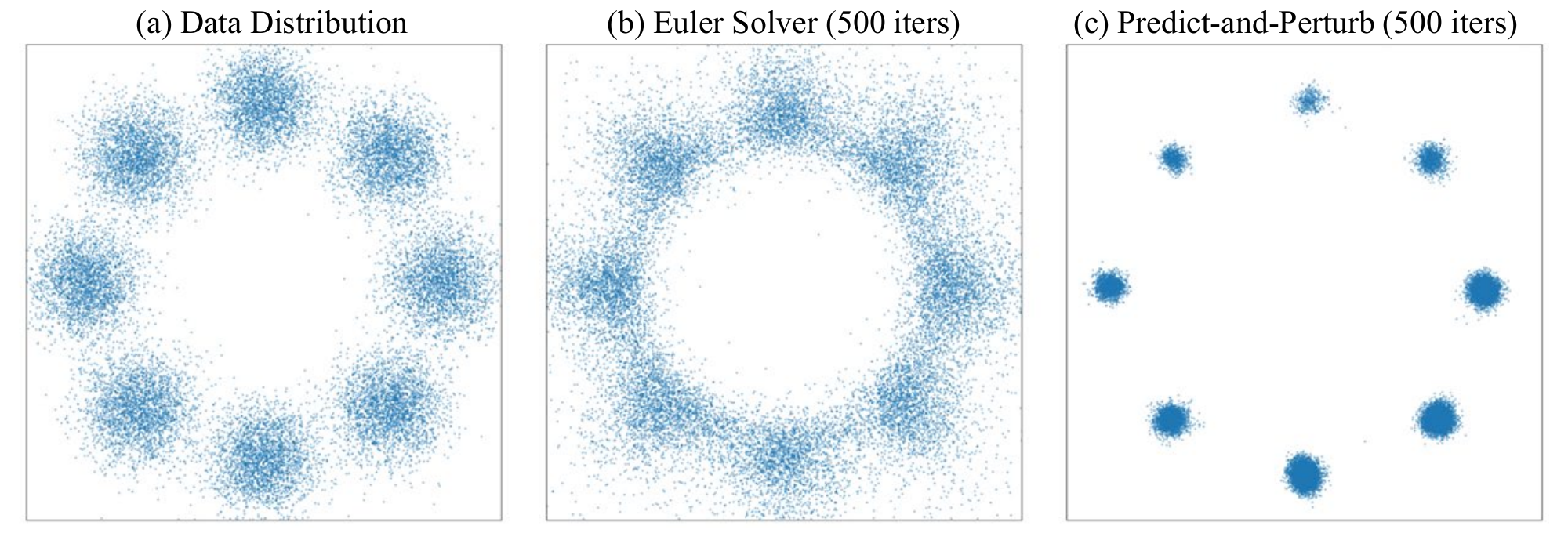}
    \vspace{-0.05in}
    \caption{Toy experiment on a 2D Gaussian mixture. Repeated P\&P iterations (i.e., $K_f\!=\!32$) yield samples concentrated in the modes. 
    % In the unconditional generation setting, sampling concentrates in high-density regions, similar to strong classifier-free guidance.
    }
    \label{fig:app_pnp_gaussian_mode}
    \vspace{-0.1in}
\end{figure*}
%%%%%%%%%%%%%%%%%%%%%%%%%%%%%%%%%%%%%%%%%%%%%%%%%%%%%%%%%%%%%%%%%%%
%%%%%%%%%%%%%%%%%%%%%%%%%%%%%%%%%%%%%%%%%%%%%%%%%%%%%%%%%%%%%%%%%%%
\begin{figure*}[t!]
    \centering
    \includegraphics[width=1\linewidth]{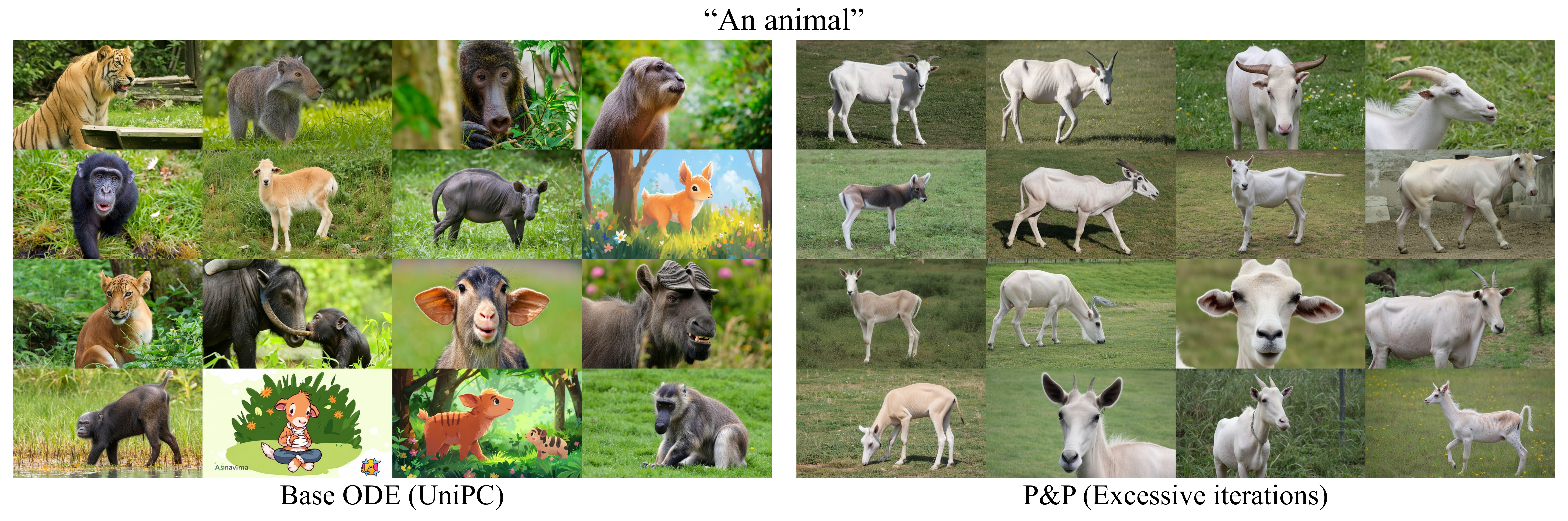}
    \vspace{-0.25in}
    \caption{Mode-seeking behavior induced by excessive P\&P iterations in image generation. We use Wan2.2-A14B T2V with a single frame and apply P\&P with $K_f\!=\!8, \tau\!=\!0$ at steps 16–20 of the 40 step flow matching inference.}
    \label{fig:app_pnp_image_mode}
    \vspace{-0.1in}
\end{figure*}
%%%%%%%%%%%%%%%%%%%%%%%%%%%%%%%%%%%%%%%%%%%%%%%%%%%%%%%%%%%%%%%%%%%

\subsection{Application to Visual Reasoning}\label{app:extending-reasoning-task}
As discussed in Sec.~\ref{sec:exp_reasoning}, we evaluate whether P\&P also improves visual reasoning capabilities~\citep{cai2025mmgr,wiedemer2025videomodelszeroshotlearners} using Wan2.2-A14B I2V. We first consider the graph traversal task introduced by \citet{wiedemer2025videomodelszeroshotlearners}.
In this task, a graph is visualized as connected nodes and edges, and the model is required to simulate a traversal process that progressively propagates from a designated source node to neighboring nodes over time. 
A qualitative example is provided in \fref{fig:app_reasoning_graph}. 
With frame-by-frame human verification over 10 runs with different random seeds, the base Wan2.2-A14B I2V model achieves a success rate of 0.1, whereas our method improves this to 0.8.

However, as shown in \fref{fig:app_reasoning_maze}, when evaluated on maze-solving tasks, the improvement remains limited.
In particular, the model frequently generates invalid paths that cross walls, and the moving block often fails to stop precisely at the target green cell.
This contrast highlights an inherent limitation of our approach: due to its local search nature, P\&P is effective for reasoning tasks that can be partially refined through motion correction or temporal consistency, such as graph traversal, where errors can be progressively corrected during sampling. In contrast, tasks whose success is determined by discrete or semantic correctness, such as maze solving, require global planning and path-level decisions, which are not easily corrected by local refinement. In such cases, incorporating external verifiers or global search mechanisms is likely necessary.

\subsection{Uncertainty Map}
Instead of variance-based methods~\citep{de2025diffusion, kou2024bayesdiff}, for example, those that estimate uncertainty by computing the variance of multiple score predictions from repeated stochastic forward passes (typically $N\!=\!5$ evaluations per step), we adopt a simpler and more efficient formulation. Our uncertainty estimate is obtained directly within the base P\&P iteration, introducing no additional sampling or computational overhead.

We provide more visual examples of uncertainty estimation in \fref{fig:app_uncertainty_time}. As inference progresses, the magnitude of perturbations decreases, leading to a gradual reduction in uncertainty. While this observation suggests that an adaptive threshold, such as a time-dependent $\tau_t$, could be considered instead of a fixed $\tau$, we leave the investigation of such adaptive schemes for future work.

%%%%%%%%%%%%%%%%%%%%%%%%%%%%%%%%%%%%%%%%%%%%%%%%%%%%%%%%%%%%%%%%%%%
\begin{figure*}[t!]
    \centering
    \includegraphics[width=1\linewidth]{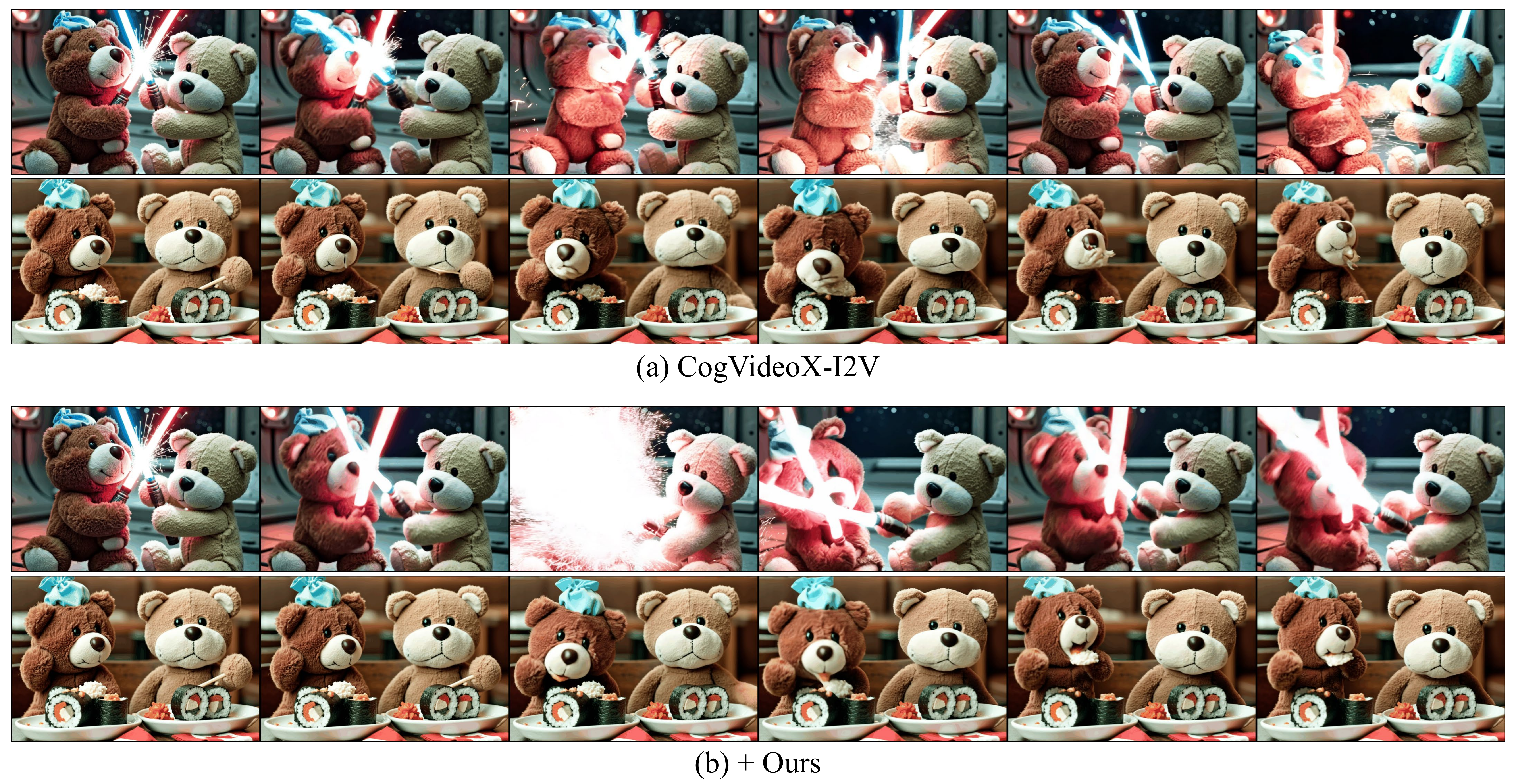}
    \vspace{-0.20in}
    \caption{P\&P is also applicable to diffusion-based video generation models (e.g., CogVideoX~\citep{yang2025cogvideox}), where it corrects video artifacts, such as a truncated lightsaber and distortions around the teddy bear's mouth. (Image credit: MuDI~\citep{jang2024identity})}
    \label{fig:app_cogx}
\end{figure*}
%%%%%%%%%%%%%%%%%%%%%%%%%%%%%%%%%%%%%%%%%%%%%%%%%%%%%%%%%%%%%%%%%%%

\subsection{Mode-Seeking Behavior of P\&P}
Toy experiments on a 2D Gaussian mixture show that excessive P\&P concentrates samples in high-density regions, exhibiting clear mode-seeking behavior, as illustrated in Fig.~\ref{fig:app_pnp_gaussian_mode}.
A similar effect is observed in image generation. As shown in \fref{fig:app_pnp_image_mode}, increasing the number of P\&P iterations ($K_f\!=\!8$) significantly reduces sample diversity, with the prompt “an animal” producing nearly identical white goats.

In the video domain, however, the effect is different. Due to cross-frame consistency and uncertainty-aware sampling, the method does not \emph{collapse} semantic diversity.
Instead, it primarily refines motion while preserving the original content, reducing undesired temporal variance such as motion artifacts or flickering.
From this perspective, our method can be viewed as an \emph{intended temporal mode-seeking} for improving output consistency.

\subsection{P\&P with Diffusion Models}
In this paper, we primarily focus on flow matching–based models, which are widely adopted in recent video and image generators. Our framework is also applicable to diffusion models, such as CogVideoX~\citep{yang2025cogvideox}, since diffusion models are trained with similar objectives, allowing our method to be applied in the same manner at inference time. 
As shown in \fref{fig:app_cogx}, P\&P corrects artifacts such as a truncated lightsaber and distortions around the teddy bear’s mouth.

\section{Limitations and Future Work} \label{app:limitations}
In this section, we discuss the limitations of our approach and outline directions for future research.

\textbf{Risk of Over-Refinement\pspace}
Hyperparameters such as the uncertainty threshold $\tau$ help prevent over-refinement and loss of semantic diversity. However, conservative settings can weaken refinement or require a larger number of P\&P iterations $K_f$. Finding a better balance between refinement strength and diversity remains future work.

\textbf{Local-Search Behavior\pspace}
Our method can be viewed as a local search process. For tasks such as maze solving, finding a good initial noise may be more effective than iterative refinement. Combining refinement with global search strategies or external verifiers is a possible direction for future research.

\textbf{Refinement Model Choice\pspace}
Although we use the same model for self-refinement, this is not a strict requirement. Future work may explore using different generative models or a fine-tuned model specialized for refinement.

\textbf{Stochasticity in Refinement\pspace}
More refinement iterations increase the chance of improvement but still rely on stochastic noise. Developing more effective ways to control or utilize this stochasticity is left for future work.

\textbf{Reducing Inference Time\pspace}
Our method introduces additional refinement steps, increasing inference time by approximately 40\% under the default setting. To reduce this overhead, we explored a simple variant that decreases both the refinement steps $K_f$ and the base ODE steps. 
Even when matching the total NFE to the baseline sampler, this variant remains more effective than standard ODE sampling. We leave further improving this trade-off between efficiency and performance for future work.

% Local search. For reasoning task, maze solving, it also requires verifier and global search algorithm.

% More iteration allows more chances for self-refinement, however, it also rely on stochasticity. Does not guarantee success.

% Future direction. 1) improved design, 2) refinement framework with external tool

%%%%%%%%%%%%%%%%%%%%%%%%%%%%%%%%%%%%%%%%%%%%%%%%%%%%%%%%%%%%%%%%%%%
\begin{figure*}[p]
    \centering
        \includegraphics[width=0.95\linewidth]{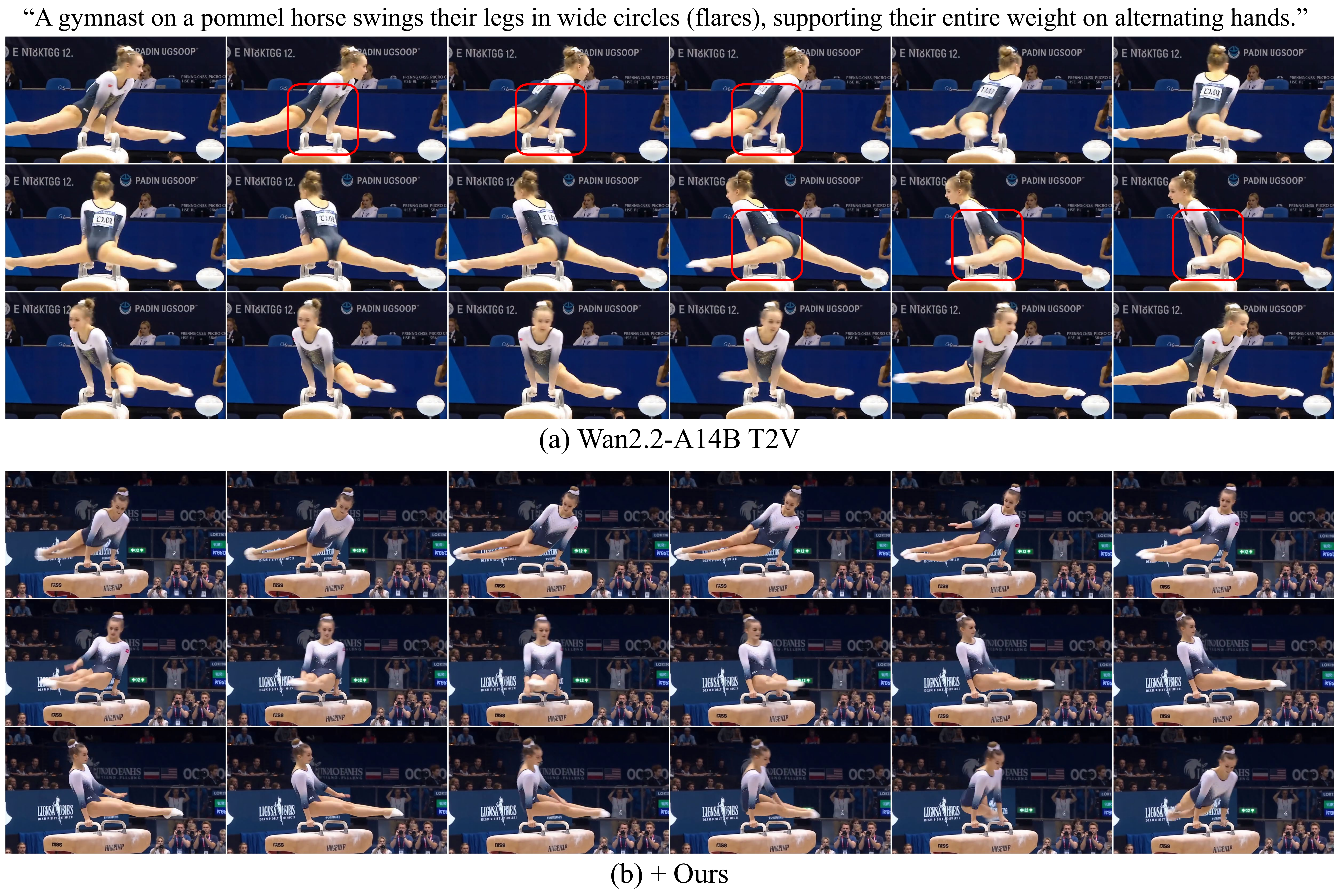}
    \vspace{-0.11in}
    % \caption{TBD.}\label{fig:fig_app_motion}
    \vspace{5mm}
        \includegraphics[width=0.95\linewidth]{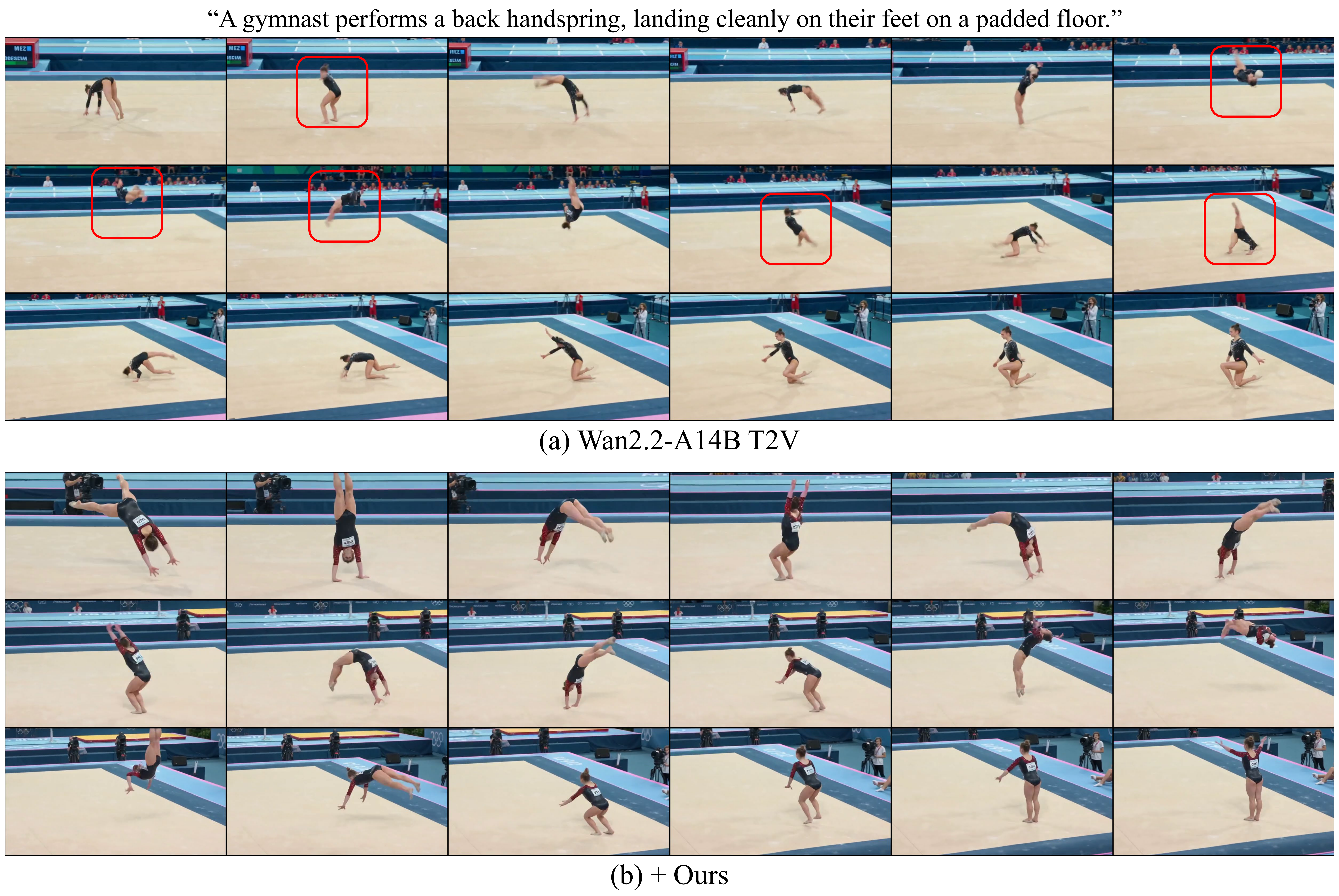}
    \vspace{-0.08in}
    \caption{Additional visual examples of complex motion generation using Wan2.2-A14B T2V.}\label{fig:fig_app_motion}
    % , visualized at full frame rate.}\label{fig:fig_app_motion}
    \vspace{-0.05in}
\end{figure*}
%%%%%%%%%%%%%%%%%%%%%%%%%%%%%%%%%%%%%%%%%%%%%%%%%%%%%%%%%%%%%%%%%%%
%%%%%%%%%%%%%%%%%%%%%%%%%%%%%%%%%%%%%%%%%%%%%%%%%%%%%%%%%%%%%%%%%%%
\begin{figure*}[t!]
    \centering
    \includegraphics[width=0.95\linewidth]{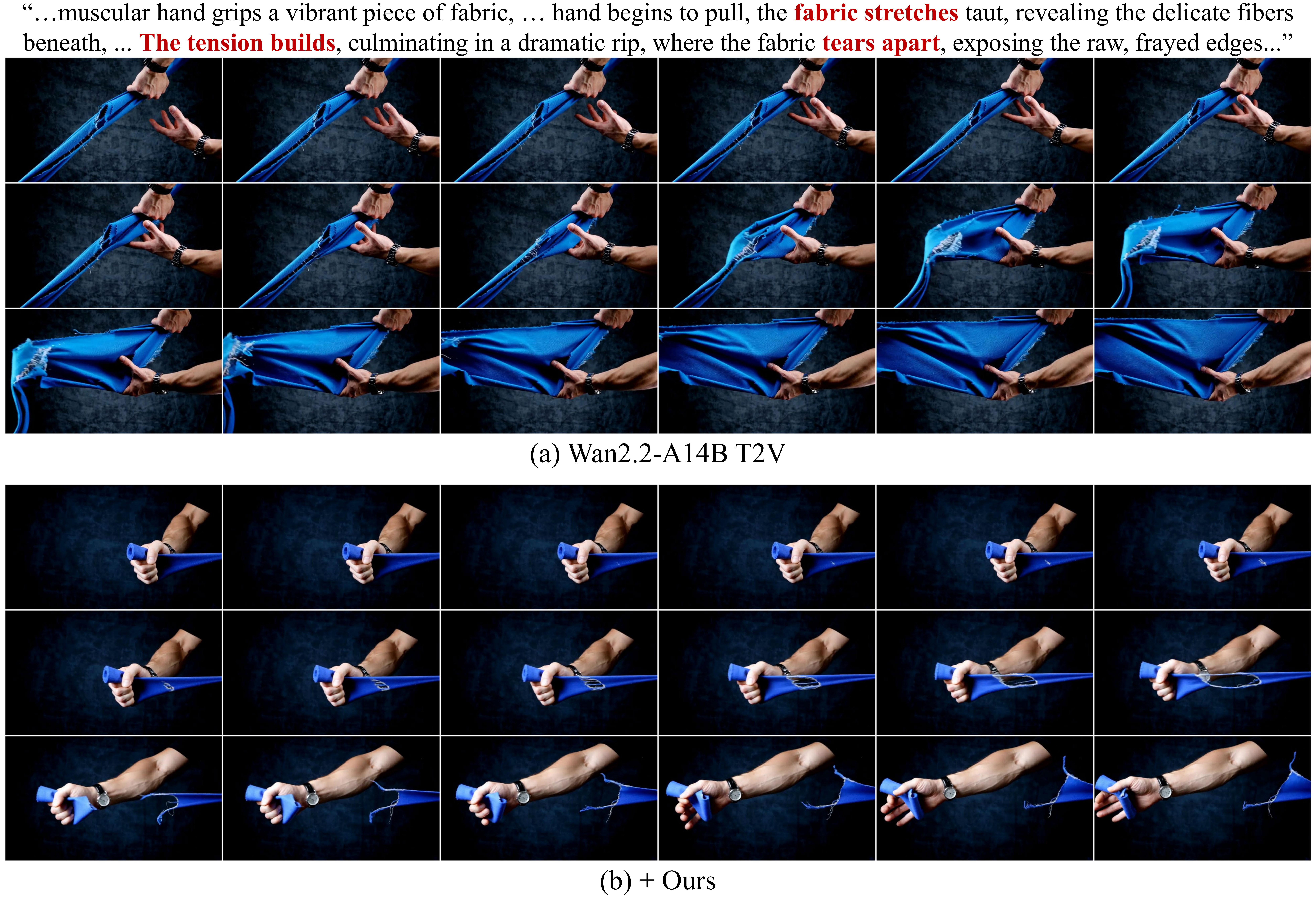}
    \vspace{-0.08in}
    \caption{Additional visual examples of physics-aligned generation using Wan2.2-A14B T2V. Our method also captures realistic physical interactions and fine-grained visual details.}\label{fig:fig_app_physics}
    \vspace{-0.05in}
\end{figure*}
%%%%%%%%%%%%%%%%%%%%%%%%%%%%%%%%%%%%%%%%%%%%%%%%%%%%%%%%%%%%%%%%%%%
%%%%%%%%%%%%%%%%%%%%%%%%%%%%%%%%%%%%%%%%%%%%%%%%%%%%%%%%%%%%%%%%%%%
%%%%%%%%%%%%%%%%%%%%%%%%%%%%%%%%%%%%%%%%%%%%%%%%%%%%%%%%%%%%%%%%%%%
%%%%%%%%%%%%%%%%%%%%%%%%%%%%%%%%%%%%%%%%%%%%%%%%%%%%%%%%%%%%%%%%%%%
% \begin{figure*}[t!]
%     \centering
%     \includegraphics[width=0.95\linewidth]{figure/app_commercial.pdf}
%     \vspace{-0.08in}
%     \caption{Qualitative comparison with commercial closed models, Veo 3.1~\citep{veo3} and Kling 2.6~\citep{kling}. While the commercial models produce more aesthetic visual quality, our method demonstrates competitive performance on complex motion scenarios. Prompt: ``A parkour athlete runs up a vertical wall, grabs the ledge, and muscles up to stand on the roof in one fluid motion.'' and ``A gymnast on a pommel horse swings their legs in wide circles (flares), supporting their entire weight on alternating hands.''}
%     \vspace{-0.05in}
% \end{figure*}
%%%%%%%%%%%%%%%%%%%%%%%%%%%%%%%%%%%%%%%%%%%%%%%%%%%%%%%%%%%%%%%%%%%
\clearpage
\onecolumn
\section{Dynamic Bench}\label{app:dynamic_bench}
1-40: Multi-object interactions, 41-80: Complex human motions, 81-120: Physics-driven dynamics.
% \begin{multicols}{1}

\begin{enumerate}
    \item \textit{A bowling ball rolls down a polished lane and strikes a perfect strike, sending all ten pins flying in different trajectories.}
    \item \textit{A chef tosses a pizza dough high into the air, catching it on their knuckles and spinning it to expand its size.}
    \item \textit{A playful Golden Retriever catches a frisbee in mid-air, causing the dog to twist its body and land on its hind legs.}
    \item \textit{A robot arm on an assembly line picks up a car door and precisely welds it onto a chassis, creating sparks upon contact.}
    \item \textit{A gust of wind blows a stack of papers off an outdoor table, causing a person to scramble and catch them before they fly away.}
    \item \textit{A sword fighter parries a heavy blow from an opponent’s axe, causing the axe to slide down the blade and spark against the crossguard.}
    \item \textit{A child builds a tower of wooden blocks, then pulls a bottom block out, causing the structure to wobble and collapse chaotically.}
    \item \textit{A pool player executes a jump shot; the cue ball hops over a blocking ball to sink the 8-ball in the corner pocket.}
    \item \textit{A sweeping broom pushes a pile of dust and small debris into a dustpan, with some dust particles escaping into the air.}
    \item \textit{A drone flies into a hanging wind chime, tangling its propellers in the strings and causing the chimes to swing violently.}
    \item \textit{A basketball hits the rim, bounces straight up, hits the backboard, and finally falls through the net.}
    \item \textit{A wrecking ball smashes through a brick wall, sending debris and dust clouding into the interior of the building.}
    \item \textit{A person pours hot milk into a cup of coffee, creating a swirling mixture of brown and white liquids.}
    \item \textit{Two bumper cars collide head-on at a carnival, causing both drivers to jolt forward while the cars recoil backward.}
    \item \textit{A tennis ball is served at high speed, deforming against the racket strings before launching across the net.}
    \item \textit{A bartender shakes a cocktail mixer vigorously, with ice cubes audibly clinking and condensation forming on the metal exterior.}
    \item \textit{A cat paws at a dangling yarn ball, causing it to swing in a pendulum motion while the cat tries to grab it again.}
    \item \textit{A heavy book falls from a shelf onto a beanbag chair, causing the chair to depress deeply and then slowly regain some shape.}
    \item \textit{A person opens a shaken soda can, causing foam to spray out and coat their hand and the table.}
    \item \textit{A skateboarder grinds along a metal rail, sparks flying from the trucks before they land on the concrete.}
    \item \textit{A knife slices through a ripe tomato, separating a slice that falls flat onto the cutting board while juice spreads.}
    \item \textit{A person types rapidly on a mechanical keyboard, with each key depressing and springing back up individually.}
    \item \textit{A wrecking crew uses a grapple to pull down a rusted metal tower, which twists and buckles before hitting the ground.}
    \item \textit{A soccer goalkeeper punches a high ball, changing its trajectory from toward the net to over the crossbar.}
    \item \textit{A magnet is brought close to a pile of iron filings, causing them to leap up and attach to the magnet in a spiky pattern.}
    \item \textit{A domino chain reaction begins, with the dominoes splitting into two separate paths that eventually trigger a small flag to raise.}
    \item \textit{A person struggles to close an overfilled suitcase, sitting on it to compress the clothes inside before zipping it shut.}
    \item \textit{A hammer strikes a nail, driving it partially into the wood, but the second strike bends the nail sideways.}
    \item \textit{A bird lands on a thin tree branch, causing the branch to bow significantly under the weight and bounce as the bird stabilizes.}
    \item \textit{A figure skater lifts their partner overhead, rotating while the partner holds a pose, their costumes flowing together.}
    \item \textit{A person uses a wrench to tighten a leaking pipe; as the nut turns, the water spray reduces to a drip.}
    \item \textit{A coin is spun on a table, wobbling faster and faster until it settles flat with a distinctive rattle.}
    \item \textit{A car drives through a large puddle, splashing water high onto the sidewalk and drenching a nearby fire hydrant.}
    \item \textit{A robotic vacuum bumps into a sleeping dog, causing the dog to lift its head and the vacuum to rotate and move away.}
    \item \textit{A majestic eagle swoops down to the water surface, snatching a fish with its talons and creating a splash pattern.}
    \item \textit{A person playing Jenga carefully pushes a block from the center, the tower swaying slightly but remaining upright.}
    \item \textit{A grandiose chandelier falls from the ceiling, crashing onto a banquet table and shattering plates and glasses.}
    \item \textit{A baker kneads heavy dough, pushing their palms into it, causing it to stretch and fold back over itself.}
    \item \textit{A bicyclist hits a curb, the front tire compressing and the rider jerking the handlebars to maintain balance.}
    \item \textit{A Newton’s Cradle is set in motion; one ball hits the stack, and the ball on the opposite end swings out, demonstrating momentum transfer.}
    \item \textit{A breakdancer performs a headspin, transitioning smoothly into a freeze pose with legs crossed in the air.}
    \item \textit{A parkour athlete runs up a vertical wall, grabs the ledge, and muscles up to stand on the roof in one fluid motion.}
    \item \textit{A ballerina performs a series of rapid fouetté turns en pointe, maintaining a fixed spotting point with her head.}
    \item \textit{A martial artist executes a flying spinning hook kick, landing in a crouched combat stance.}
    \item \textit{A gymnast on the uneven bars swings from the high bar, releases, performs a double backflip, and re-catches the bar.}
    \item \textit{A figure skater executes a triple axel, taking off forward and rotating three and a half times before landing backward on one foot.}
    \item \textit{A capoeira practitioner performs a ginga movement followed immediately by a low sweeping leg kick (meia lua de compasso).}
    \item \textit{A high jumper performs the Fosbury Flop, arching their back severely over the bar and kicking their legs up at the last second.}
    \item \textit{A yoga instructor flows from a downward dog into a scorpion handstand, balancing on their forearms with legs arched over their head.}
    \item \textit{A sprinter explodes out of the starting blocks, body at a 45-degree angle, transitioning into an upright running posture.}
    \item \textit{A rock climber performs a dynamic "dyno" move, leaping from one hold to a distant hold, catching it with one hand and swinging.}
    \item \textit{A rhythmic gymnast throws a hoop high into the air, performs a cartwheel, and catches the hoop with her foot.}
    \item \textit{A snowboarder rides up a halfpipe, performs a McTwist (inverted 540 degree spin), and lands cleanly on the transition.}
    \item \textit{A professional wrestler performs a suplex on a dummy, arching their back to throw the weight over their head.}
    \item \textit{A salsa dancer spins their partner rapidly, then dips them low to the ground, pausing for a beat before pulling them back up.}
    \item \textit{A pole vaulter plants the pole, the pole bends dramatically, launching the athlete feet-first over the bar.}
    \item \textit{A surfer performs a sharp cutback on a wave, twisting their torso and shifting weight to spray water off the tail of the board.}
    \item \textit{A contortionist slowly bends backward from a standing position until they grab their own ankles.}
    \item \textit{A hip-hop dancer performs "the worm," rippling their body along the floor from chest to feet.}
    \item \textit{A soccer player performs a bicycle kick, leaping back-first into the air and scissoring legs to strike the ball.}
    \item \textit{A diver performs a reverse 2.5 somersault from the 10-meter platform, entering the water with minimal splash.}
    \item \textit{A fencer lunges deeply with a foil, extending their arm fully while their back leg remains straight and grounded.}
    \item \textit{A heavy metal drummer plays a rapid blast beat, arms and legs moving in a blur of independent rhythms.}
    \item \textit{A traditional Indian dancer (Bharatanatyam) stomps rhythmically while performing complex mudras (hand gestures) and eye movements.}
    \item \textit{A cheerleader is thrown into the air, performs a twist, and is caught in a cradle position by her teammates.}
    \item \textit{A skateboarder performs a tre-flip (360 pop shove-it plus a kickflip) down a set of stairs.}
    \item \textit{A stunt performer is "shot," jerking backward violently and falling over a railing, flailing arms.}
    \item \textit{A tai chi master performs "Parting the Wild Horse's Mane," moving with extreme slowness and fluid weight transfer.}
    \item \textit{A basketball player performs a crossover dribble, fake-drives left, spins right, and performs a slam dunk.}
    \item \textit{A swimmer performs a tumble turn underwater, tucking tightly and pushing off the wall to glide in a streamline.}
    \item \textit{A trapeze artist releases their bar, performs a triple somersault in mid-air, and is caught by the catcher on the opposing bar.}
    \item \textit{A person slips on a banana peel (cartoon style), feet flying up above their head before they land flat on their back.}
    \item \textit{A cricket bowler runs up and delivers the ball with a straight-arm action, following through with their body momentum.}
    \item \textit{A baton twirler spins the baton around their body, under their legs, and over their neck without using their hands.}
    \item \textit{A synchronized swimming team emerges from the water in a pyramid formation, holding the pose before sinking back down.}
    \item \textit{A BMX rider performs a backflip tailwhip over a dirt jump, kicking the bike frame around while upside down.}
    \item \textit{A slackliner walks across a loose line, arms flailing to maintain balance as the line shakes violently.}
    \item \textit{An ice hockey goalie drops into a butterfly position to block a shot, then quickly scrambles back to a standing position.}
    \item \textit{A conductor leads an orchestra with vigorous arm movements, hair flying as they signal a crescendo.}
    \item \textit{A gymnast on a pommel horse swings their legs in wide circles (flares), supporting their entire weight on alternating hands.}
    \item \textit{A glass of red wine shatters on a marble floor, the liquid splashing outward in slow motion while shards glide across the surface.}
    \item \textit{Thick, golden honey is poured from a jar onto a stack of pancakes, folding over itself and slowly dripping down the sides.}
    \item \textit{A silk scarf blows in a violent gale storm, rippling rapidly and snapping in the wind without tearing.}
    \item \textit{A water balloon hits a person’s face in slow motion, the rubber expanding around their features before bursting and spraying water.}
    \item \textit{A large soap bubble floats through the air, wobbling and reflecting an iridescent rainbow before popping into tiny droplets.}
    \item \textit{A campfire crackles in the night, with sparks rising in a spiral pattern and smoke shifting direction with the breeze.}
    \item \textit{A car drives through thick fog, its headlights creating volumetric beams that illuminate the swirling mist particles.}
    \item \textit{A block of dry ice is dropped into warm water, instantly generating a thick, heavy white fog that spills over the container's edge.}
    \item \textit{A handful of glitter is thrown into the air, catching the light and twinkling as it drifts slowly to the ground.}
    \item \textit{A large wave crashes against a cliffside, the water atomizing into a fine mist and white foam running down the rocks.}
    \item \textit{A cannonball is fired into a sand dune, displacing a massive crater of sand that sprays outward and slides back into the hole.}
    \item \textit{A heavy velvet curtain is pulled back, bunching up in thick, heavy folds that sway heavily with the movement.}
    \item \textit{A distinct drop of ink falls into a glass of clear water, blooming into abstract, smoke-like tendrils as it diffuses.}
    \item \textit{A pristine snowbank collapses, triggering a small avalanche where clumps of snow break apart into powder as they slide.}
    \item \textit{A jellyfish swims in the deep ocean, its translucent bell pulsing rhythmically and its long tentacles trailing fluidly behind.}
    \item \textit{A person with long hair stands in front of a high-powered fan, the hair whipping chaotically and obscuring their face.}
    \item \textit{Molten lava flows slowly down a volcano, the surface cooling into black crust while red-hot magma breaks through the cracks.}
    \item \textit{A rubber ball bounces on a trampoline, depressing the surface deeply and launching higher with every bounce.}
    \item \textit{A stack of newspapers is left in the rain; the paper darkens, sags, and begins to disintegrate into pulp.}
    \item \textit{A tornado touches down in a field, pulling up dirt, grass, and debris into a rotating funnel cloud.}
    \item \textit{A high-speed bullet passes through an apple, causing the exit side to explode outward in a cone of pulp and juice.}
    \item \textit{A candle flame flickers in a drafty room, the wax melting and dripping down the side of the candle unevenly.}
    \item \textit{A bowl of Jell-O is nudged, wobbling vigorously with a gelatinous, elastic motion that slowly dampens.}
    \item \textit{A heavy metal chain is dropped onto a metal floor, coiling and uncoiling as the links settle with a metallic weight.}
    \item \textit{Dust motes dance in a shaft of sunlight in an old attic, moving with Brownian motion.}
    \item \textit{A wet dog shakes itself dry in slow motion, the loose skin rippling and water droplets forming a halo around the animal.}
    \item \textit{A porcelain vase is glued back together, but when filled with water, it slowly leaks from the cracks, forming beads on the surface.}
    \item \textit{A huge flag waves in slow motion, showcasing the heavy fabric rolling and snapping, creating shadows within the folds.}
    \item \textit{Oil and vinegar are shaken in a bottle, forming temporary emulsions of small bubbles that slowly separate back into layers.}
    \item \textit{A meteor enters the atmosphere, burning up with a fiery tail and shedding glowing debris before disintegrating.}
    \item \textit{A feather falls in a vacuum chamber (straight down) versus a feather falling in air (drifting side to side).}
    \item \textit{A mesmerizing ferrofluid spikes and dances in response to a moving magnetic field, the black liquid looking alien and sharp.}
    \item \textit{Raindrops hit a puddle, creating concentric ripples that interfere with one another in a complex geometric pattern.}
    \item \textit{A marshmallow is roasted over a fire, the outer skin bubbling, browning, and eventually catching a small blue flame.}
    \item \textit{A piece of paper burns from the center, the edges curling and turning to black ash that flakes away.}
    \item \textit{A slime toy is stretched between two hands, becoming thin and translucent before snapping back into a glob.}
    \item \textit{Heavy rain falls on a car windshield, the wipers pushing the water aside in sheets that immediately reform.}
    \item \textit{A wrecking ball hits a building made of glass, causing a cascade of shattering panes that reflect the sky as they fall.}
    \item \textit{Steam rises from a hot geyser, billowing rapidly and dissipating into the cold air above.}
    \item \textit{A hand touches a plasma globe, causing the purple arcs of electricity to concentrate and follow the fingers across the glass.}
\end{enumerate}
% \end{multicols}
\twocolumn

\end{document}